\documentclass[twoside]{article}

\usepackage[accepted]{aistats2023}

\usepackage[colorlinks,citecolor=blue]{hyperref}
\usepackage[utf8]{inputenc} 
\usepackage{url}            
\usepackage{booktabs}       
\usepackage{amsfonts}       
\usepackage{nicefrac}       
\usepackage{microtype}      

\usepackage{times}             

\usepackage[english]{babel} 

\usepackage{amsmath}
\usepackage{amssymb}
\usepackage{mathtools}
\usepackage{amsthm}

\usepackage{bbm}
\usepackage{verbatim,float,dsfont}
\usepackage{psfrag}
\usepackage{algorithm,algorithmic}
\usepackage{thmtools,thm-restate}
\usepackage{mathtools,enumitem}
\mathtoolsset{showonlyrefs}
\usepackage{tcolorbox}
\usepackage{breqn,lipsum}
\usepackage{stackengine}
\usepackage{array}
\usepackage{caption}
\usepackage{boldline}
\usepackage{amssymb}
\usepackage{multirow}

\usepackage{subfig}
\usepackage{graphicx}
\usepackage{wrapfig}
\usepackage[nocomma]{optidef}
\usepackage{mdframed}

\usepackage{tikz}
\usetikzlibrary{shapes,arrows}

\usetikzlibrary{positioning, quotes}

\newtheorem{theorem}{Theorem}
\newtheorem{lemma}[theorem]{Lemma}
\newtheorem{corollary}[theorem]{Corollary}
\newtheorem{proposition}[theorem]{Proposition}
\newtheorem{remark}[theorem]{Remark}
\newtheorem{definition}[theorem]{Definition}

\newtheorem{example}[theorem]{Example}

\makeatletter
\newcommand{\setword}[2]{%
  \phantomsection
  #1\def\@currentlabel{\unexpanded{#1}}\label{#2}%
}
\makeatother

\makeatletter
\def\set@curr@file#1{\def\@curr@file{#1}} 
\makeatother
\usepackage[load-configurations=version-1]{siunitx} 

\def\shownotes{1}  
\ifnum\shownotes=1
\newcommand{\authnote}[2]{$\ll$\textsf{\footnotesize #1 notes: #2}$\gg$}
\else
\newcommand{\authnote}[2]{}
\fi

\usepackage{accents}

\makeatletter
\newcommand*\rel@kern[1]{\kern#1\dimexpr\macc@kerna}
\newcommand*\widebar[1]{%
  \begingroup
  \def\mathaccent##1##2{%
    \rel@kern{0.8}%
    \overline{\rel@kern{-0.8}\macc@nucleus\rel@kern{0.2}}%
    \rel@kern{-0.2}%
  }%
  \macc@depth\@ne
  \let\math@bgroup\@empty \let\math@egroup\macc@set@skewchar
  \mathsurround\z@ \frozen@everymath{\mathgroup\macc@group\relax}%
  \macc@set@skewchar\relax
  \let\mathaccentV\macc@nested@a
  \macc@nested@a\relax111{#1}%
  \endgroup
}
\makeatother

\newcommand{\argmin}{\mathop{\mathrm{argmin}}}

\def\sign{\mathrm{sign}}

\def\cA{\mathcal{A}}

\def\cD{\mathcal{D}}

\def\cF{\mathcal{F}}
\def\cG{\mathcal{G}}
\def\cK{\mathcal{K}}
\def\cL{\mathcal{L}}

\def\cI{\mathcal{I}}

\def\cN{\mathcal{N}}
\def\cP{\mathcal{P}}
\def\cQ{\mathcal{Q}}
\def\cR{\mathcal{R}}

\def\cW{\mathcal{W}}

\def\TV{\mathrm{TV}}

\def\code#1{\texttt{#1}}

\def\bs{\ensuremath\boldsymbol}
\def\TV{\ensuremath\mathcal{TV}}

\newcommand{\grad}{\nabla}
\newcommand{\lamda}{\lambda}

\usepackage{times}

\DeclareMathSymbol{\mrq}{\mathord}{operators}{`'}

\usepackage[round]{natbib}

\begin{document}

%

%

\twocolumn[

\aistatstitle{Second Order Path Variationals in Non-Stationary Online Learning}

\aistatsauthor{Dheeraj Baby \And Yu-Xiang Wang}

\aistatsaddress{dheeraj@cs.ucsb.edu \And  yuxiangw@cs.ucsb.edu} 

\aistatsaddress{Department of Computer Science\\
University of California Santa Barbara }
]

\begin{abstract}
We consider the problem of \emph{universal} dynamic regret minimization under exp-concave and smooth losses. We show that appropriately designed Strongly Adaptive algorithms achieve a dynamic regret of $\tilde O(d^2 n^{1/5} [\TV_1(\bs w_{1:n})]^{2/5} \vee d^2)$, where $n$ is the time horizon and $\TV_1(\bs w_{1:n})$ a path variational based on \emph{second order differences} of the comparator sequence. Such a path variational naturally encodes comparator sequences that are piece-wise linear -- 
a powerful family that tracks a variety of non-stationarity patterns in practice \citep{l1tf}. The aforementioned dynamic regret is shown to be optimal modulo dimension dependencies and poly-logarithmic factors of $n$. To the best of our knowledge, this path variational has not been studied in the non-stochastic online learning literature before. Our proof techniques rely on analysing the KKT conditions of the offline oracle and requires several non-trivial generalizations of the ideas in \citet{Baby2021OptimalDR} where the latter work only implies an $\tilde{O}(n^{1/3})$ regret for the current problem.
\end{abstract}

 \section{INTRODUCTION} \label{sec:intro}
Online Convex Optimization (OCO) \citep{zinkevich2003online,hazan2016introduction} is a widely studied setup in machine learning that has witnessed a myriad of influential applications such as time series forecasting, building recommendation engines etc. In this setting, a learner plays an iterative game with an adversary that last for $n$ rounds. In each round $t \in [n]:=\{1,\ldots,n \}$, the learner makes a decision $\bs p_t$ that belongs to a \emph{decision space} $\cD \subset \mathbb{R}^d$. Then a convex loss \emph{loss function} $f_t:\mathbb{R}^d \rightarrow \mathbb{R}$ is revealed by the adversary. The learner suffers a cost of $f(\bs p_t)$ at round $t$ for making its decision. Now, given a \emph{benchmark space} of decisions $\cW \subseteq \cD$, we aim to study learners that can control its \emph{dynamic regret} against \emph{any} sequence of comparators from the benchmark: 
\begin{align}
    R_n(\bs w_{1:n})
    &:= \sum_{t=1}^n f_t(\bs p_t) - f_t(\bs w_t), \label{eq:dregret}
\end{align}
where we abbreviate the \emph{comparator sequence} $\bs w_{1:n} := \bs w_1,\ldots,\bs w_n$ where each $\bs w_t \in \cW$. This is known to be a good metric for characterizing the performance of a learner in non-stationary environments \citep{zinkevich2003online,zhang2018adaptive,Cutkosky2020ParameterfreeDA}. The quantity in Eq.\eqref{eq:dregret} is also sometimes referred as \emph{universal dynamic regret} \citep{zhang2018adaptive} because we do not impose any constraints on the comparator sequence $\bs w_{1:n}$ except that each sequence member must belong to the benchmark set $\cW$.
This is a different and more powerful way of tackling distribution-shifts than other methods that model the environment explicitly \citep[e.g.,][]{besbes2015non,baby2020higherTV}.

Let us illustrate the point in a weather forecasting application in which $f_t(\bs w_t) = \ell(y_t, x_t^T\bs w_t)$ where $x_t$ is a feature vector (e.g., humidity and temperature at Day $t$), $y_t$ is the actual precipitation of the next day and $\ell$ is a loss function. The underlying distribution of $y_t|x_t$ is determined by nature and could drift over time due to unobserved variables such as climate change. The approach of \citet{besbes2015non,baby2020higherTV} would be to assume a model, e.g., $y_t = x_t^Tw^*_t + \text{noise}$ and control the regret against $w^*_{1:n}$ in terms of the variation of the \emph{true} regression coefficients over time. In contrast, a \emph{universal dynamic regret} approach will not make any assumption about the world, but instead will compete with the best time-varying sequence of comparators that can be chosen \emph{in hindsight}. In the case when the model is correct, we can choose the comparators to be $w^*_{1:n}$; otherwise, we can compete with the best sequence of linear predictors that optimally balances the bias and variance.




A bound on $R_n(\bs w_{1:n})$ is usually expressed in terms of the time horizon $n$ and a \emph{path variation} that captures the smoothness of the comparator sequence $\bs w_{1:n}$. Some examples of such path variationals include $P(\bs w_{1:n}) = \sum_{t=1}^{n-1}\|\bs w_t - \bs w_{t+1} \|_2$ \citep{zinkevich2003online} and more recently $\mathcal{TV}(\bs w_{1:n}) = \sum_{t=1}^{n-1}\|\bs w_t - \bs w_{t+1} \|_1$ \citep{Baby2021OptimalDR}.

\paragraph{Comparator as a discretized function.} When we view the sequence of comparators as a function of time, it is natural to describe them as a discretization of (continuous-time) functions residing in some non-parametric function classes. We now proceed to expand upon this idea. For a function $f:[0,1] \rightarrow \mathbb{R}$ that is $k$ times (weakly) differentiable, define the Total Variation (TV) of its $k^{th}$ derivative $f^{(k)}$ to be:
\begin{align}
    TV(f^{(k)}) := \sup_{0=z_1<\ldots<z_{N+1}=1} \sum_{i=1}^N |f^{(k)}(z_{i+1}) - f^{(k)}(z_i)|. \label{eq:ctstv}
\end{align}
If the function has $k+1$ continuous derivatives, then $TV(f^{(k)})$ is equivalent to $\int_{0}^1 |f^{(k+1)}(x)|dx$. Given $n,C_n > 0$ one may define the function space:
\begin{align}
 \cF_k(C_n) := \{f:[0,1] \rightarrow \mathbb{R} |  TV(f^{(k)}) \le C_n \}.  \label{eq:fclass} 
\end{align}
This space is known to contain functions that have a piece-wise degree $k$ polynomial structure \citep{trendfilter}. We can generate interesting comparator sequence families by discretizing such function spaces. First we fix some notations. For a sequence of vectors $\bs v_{1:\ell} := \bs v_1, \ldots, \bs v_{\ell}$, define the first order discrete difference operation $D \bs v := \bs v_2 - \bs v_1, \ldots, \bs v_{\ell} - \bs v_{\ell-1}$. For any positive integer $k$, the $k^{th}$ order discrete difference -- $D^k$ -- of a sequence is obtained via applying the operation $D$ for $k$ times. For a sequence $\bs v_{1:\ell}$, we define $\|\bs v_{1:\ell} \|_1 = \sum_{j=1}^\ell \| \bs v_j\|_1 $

\paragraph{The higher order TV distance.} Next, we define a path length which is the discrete analogue of $TV(f^{(k)})$ in Eq.\eqref{eq:ctstv} as follows:
\begin{align}
    \mathcal{TV}_k(\bs w_{1:n}) := n^k \|D^{k+1} \bs w_{1:n}\|_1. \label{eq:tvk}
\end{align}

\begin{figure*}
  \centering
  \begin{minipage}{.33\textwidth}
  \centering
   \subfloat[\centering Two weakly differentiable functions ]{\includegraphics[width=0.8\textwidth,height=0.7\textheight,keepaspectratio]{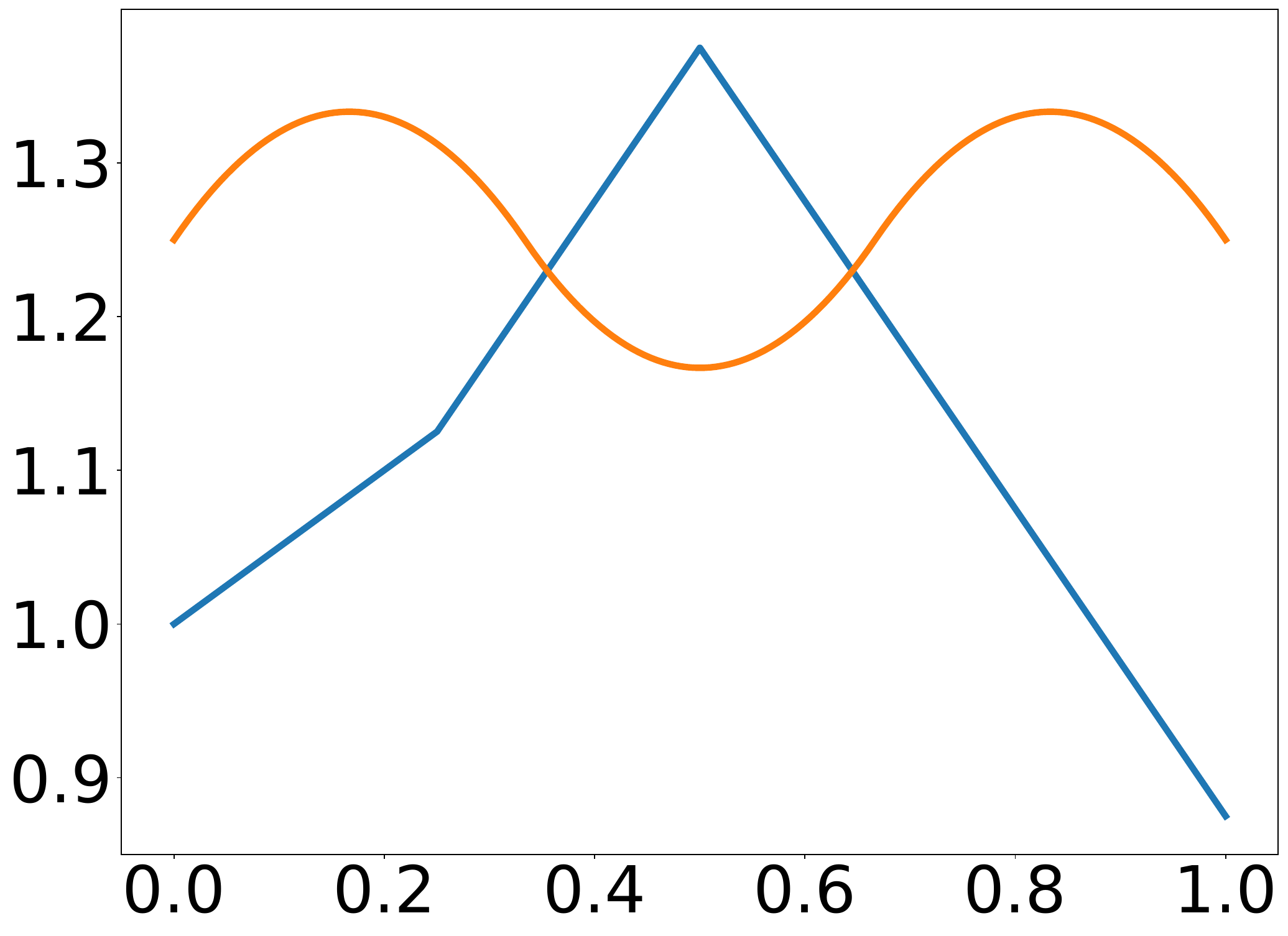}}%
  \end{minipage}%
  \begin{minipage}{.33\textwidth}
  \centering
  \subfloat[\centering discrete $TV^1$ distance ]{\includegraphics[width=0.9\textwidth,height=0.1\textheight]{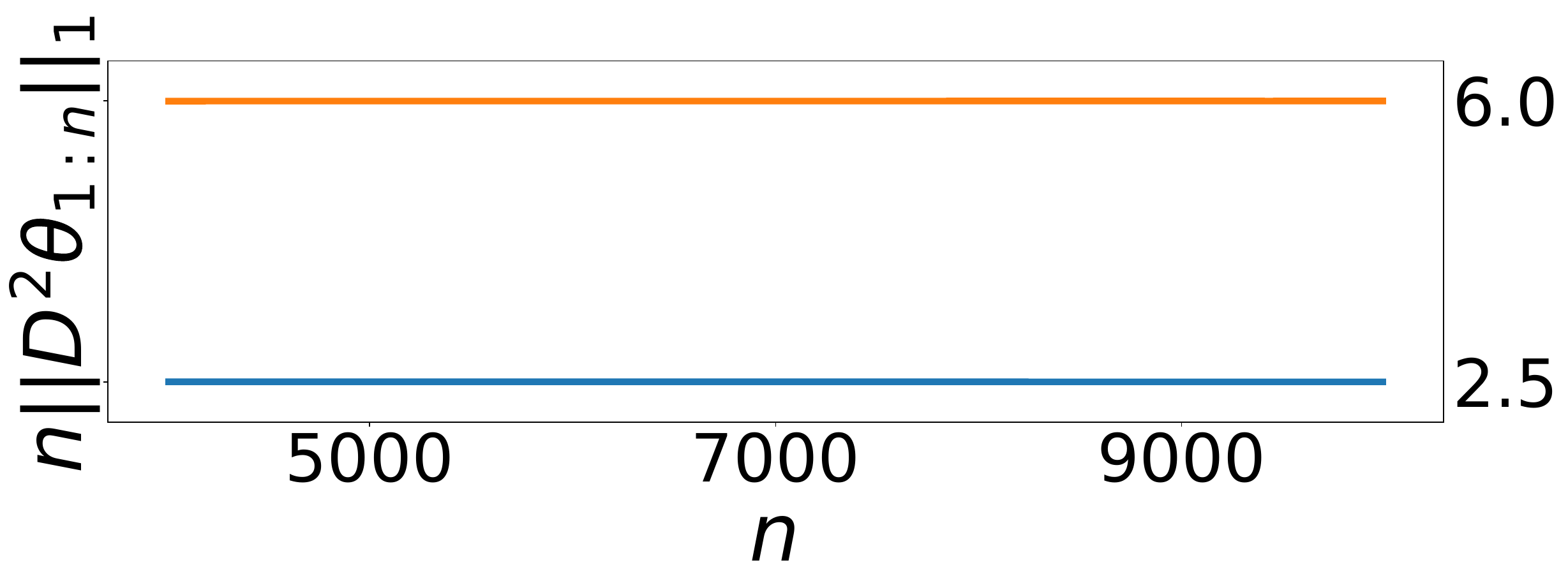}}%
\end{minipage}%
  \begin{minipage}{.33\textwidth}
  \centering
  \subfloat[\centering discrete $TV^0$ distance]{\includegraphics[width=0.9\textwidth,height=0.1\textheight]{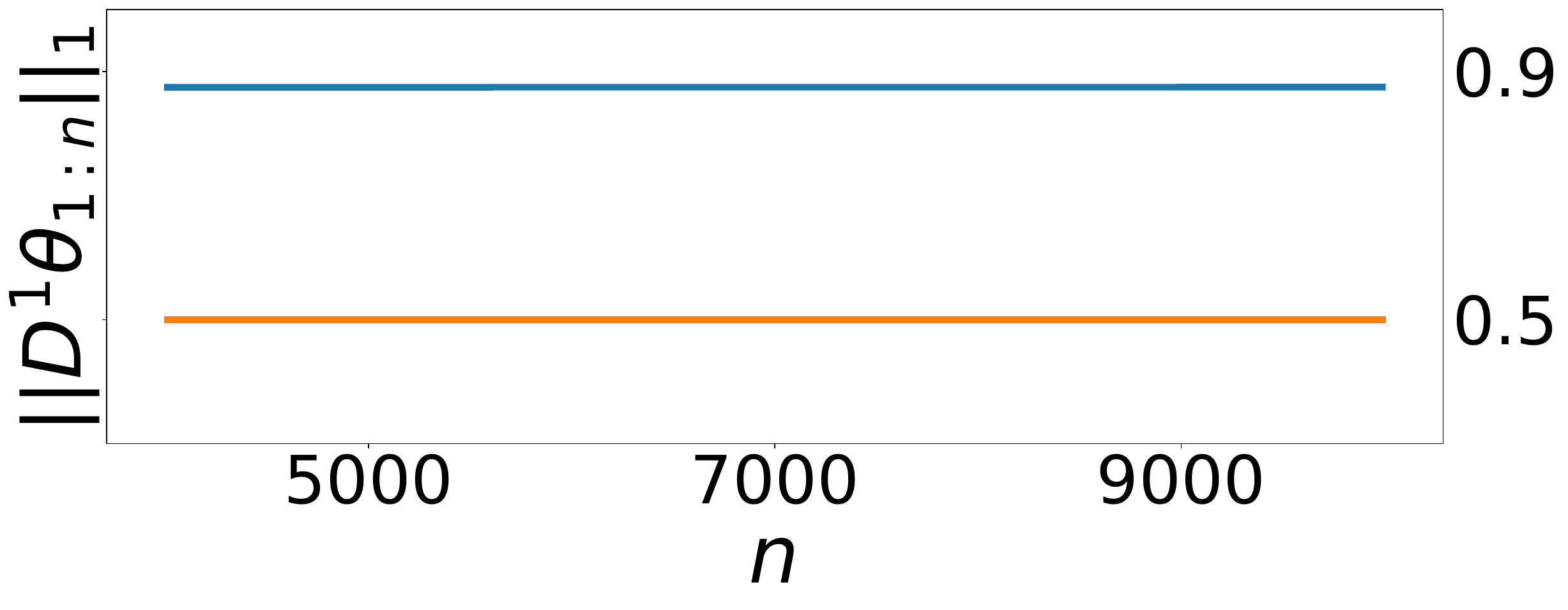}}%
\end{minipage}
  \caption{\emph{A $TV^1$ bounded comparator sequence $\bs w_{1:n}$ can be obtained by discretizing the weakly differentiable functions displayed in Fig(a) at points $i/n$, $i \in [n]$. In Fig(b), we plot the $TV^1$ distance (which is equal to $n \|D^2 \bs w_{1:n} \|_1$ by definition) of the generated sequence for various sequence lengths $n$. Blue (orange) curve in Fig(b) corresponds to the statistics of the discretization of the blue (orange) curve in Fig(a). As $n$ increases the discrete $TV^1$ distance converges to a constant value given by the continuous $TV^1$ distance of the functions in Fig(a). In Fig(c) we plot the $TV^0$ distance of the discretizations.  Thus in this example, we see that both $\|D^1\bs w_{1:n} \|_1$ and $n\|D^2 \bs w_{1:n} \|_1$ are  $O(1)$ as $n$ grows. Since $TV^1$ distance of the sequences is $O(1)$, the algorithm that we propose in Section \ref{sec:alg} is able to obtain the faster dynamic regret rate of $\tilde O(n^{1/5})$ as opposed to the rate of $\tilde O(n^{1/3})$ obtainable from \citet{Baby2021OptimalDR} for sequences with bounded $TV^0$ distance. Furthermore, the functions in Fig(a) are reminiscent to the real-life trends observed in Fig.\ref{fig:demo}.}}\label{fig:illustration}
\end{figure*}

It is a \textbf{common misconception} that constraining $\TV_k$ can be alarmingly restrictive due to the presence of the multiplicative factor of $n^k$ in its definition. To clarify that this is not the case, we observe that the multiplicative factor of $n^k$ arises naturally as a consequence of the Riemann approximation of the continuous Total Variation displayed in Eq.\eqref{eq:ctstv} at a resolution $1/n$. This observation leads to the following scheme of generating sequences with $\TV_1(\bs w_{1:n}) = O(C_n)$ for any given number $C_n$: Along any coordinate $j\in [d]$, generate the sequence $\bs w_1[j],\ldots,\bs w_n[j]$  via sampling a function $f_j(x) \in \cF_k(C_{n,j})$ at points $x = i/n$ for $i \in [n]$ with the property that $\sum_{j=1}^d C_{n,j} = C_n$. For example, considering the case of $k=1$ and $d=1$, if $TV(f^{(1)})$ is $O(n^{\alpha})$ for some $\alpha \ge 0$, then $\TV_1(\bs w_{1:n}) := n\|D^2 \bs w_{1:n} \|_1$ is also $O(n^{\alpha})$ despite the multiplicative factor of $n$ appearing in the quantity $\TV_1(\bs w_{1:n})$. A demonstration of this phenomenon for $\alpha=0$ is displayed in Fig.\ref{fig:illustration}.

\begin{figure}%
    \centering
    \subfloat[\centering S\&P500 stock prices ]{{\includegraphics[width=6cm]{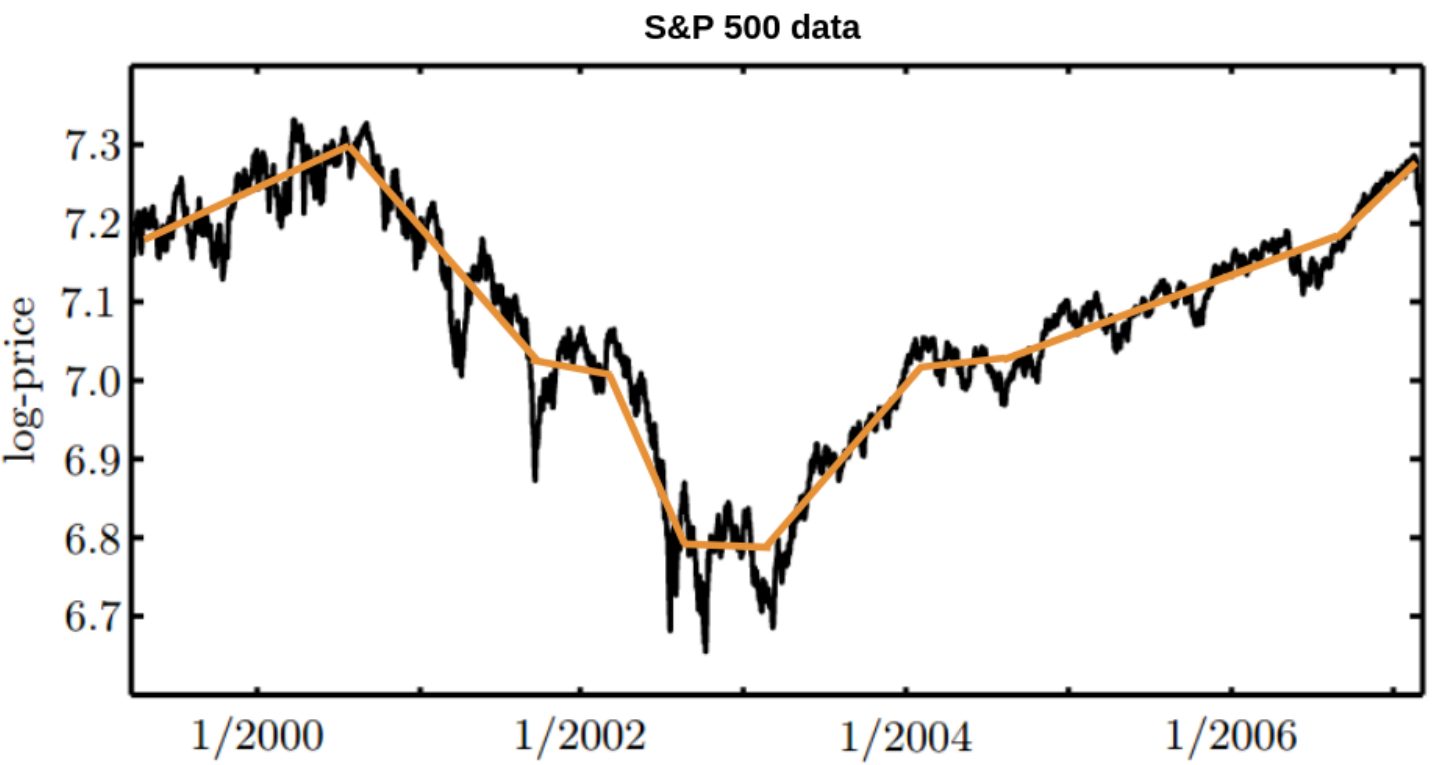} }}%
    \qquad
    \subfloat[\centering Daily COVID cases]{{\includegraphics[width=6cm]{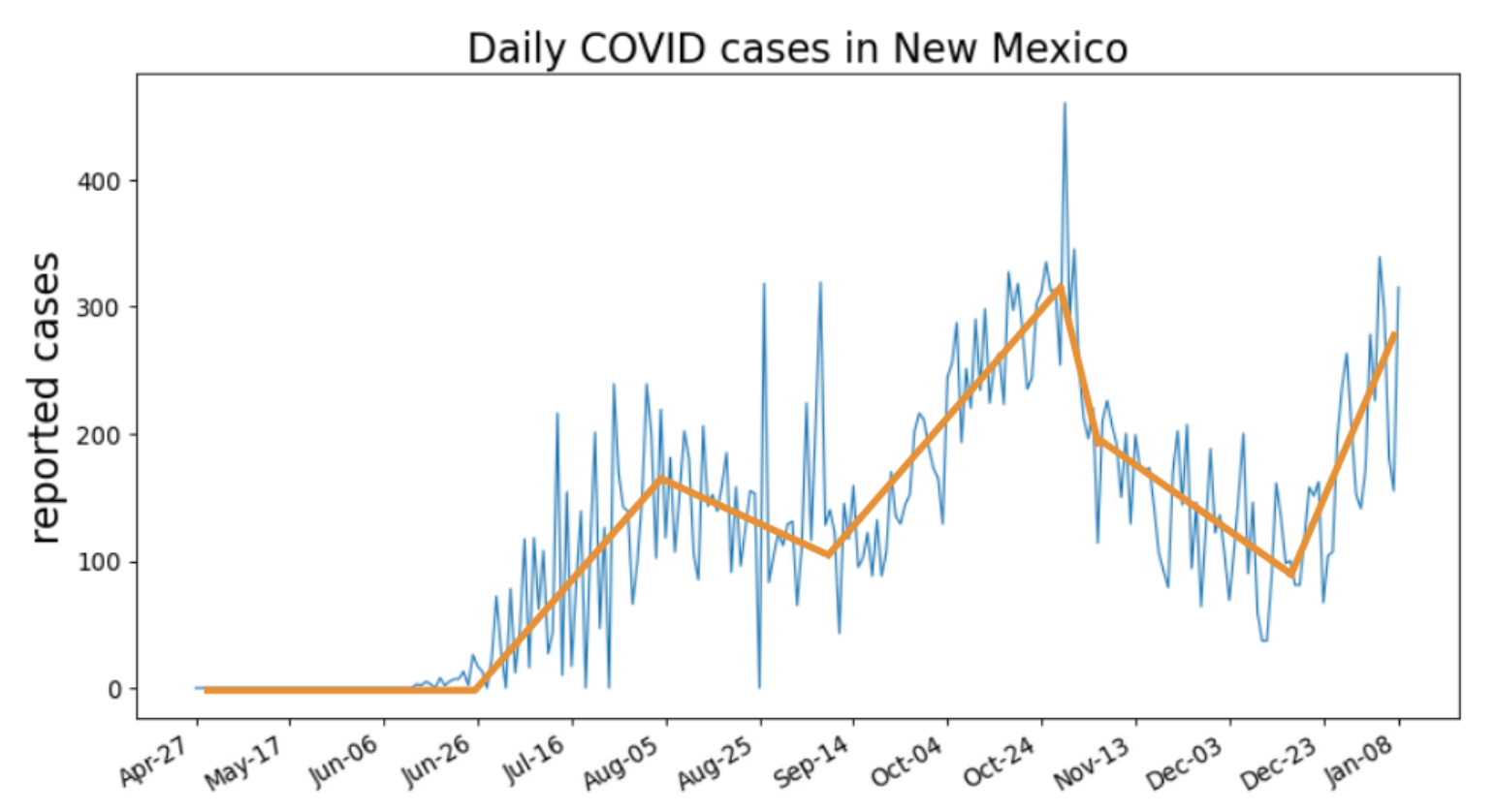} }}%
    \caption{\emph{Fig.(a) displays S\&P500 data and Fig.(b) displays Daily COVID cases reported in the state of New Mexico, USA. In both scenarios the underlying trend (obtained via an L1 Trend Filter \citep{l1tf}) exhibits a weakly differentiable piece-wise linear structure (orange).}} \label{fig:demo}
\end{figure}

\paragraph{Why is this useful?} In this paper, we focus on comparators with bounded $\TV_1$ distance (i.e Eq.\eqref{eq:tvk} with $k=1$). Our goal will be to bound the dynamic regret Eq.\eqref{eq:dregret} against $\bs w_{1:n}$ as a function of $n$ and $\TV_1(\bs w_{1:n})$. Due to the presence of second order differencing operation in the definition of $\TV_1$, this path length is ideal to capture the variation incurred by comparators with piece-wise linear structure across each coordinate (see Definition \ref{def:piecelinear}). The points where the sequence transition from one linear structure to other can be interpreted as abrupt changes or events in the underlying comparator dynamics. The value of $\TV_1(\bs w_{1:n})$ simultaneously captures the sparsity (due to the presence of $L_1$ norm in Eq.\eqref{eq:tvk}) and intensity of such changes. Many real world time series data are known to contain piece-wise linear trends. See for example Fig.\ref{fig:demo} or \citet{l1tf} for more examples. Hence controlling the dynamic regret in terms of $\TV_1(\bs w_{1:n})$ has \emph{significant practical value}.

\paragraph{Fast rate phenomenon.} Path lengths of the form $\TV_1$ (or more generally $\TV_k$) have gained significant attention and have been the subject of extensive study in the stochastic non-parametric regression community for over two decades \citep{vandegeer1990,donoho1998minimax,l1tf,trendfilter,graphtf}. These works aim to estimate an unknown scalar (i.e $d=1$) sequence  $w_{1:n}$ from $n$ noisy observations $y_t = w_t + \cN(0,\sigma^2)$ in an offline setting. They propose algorithms that produce estimates $\hat {\bs w}_{1:n}$ such that the expected total squared error $\sum_{t=1}^n E[(\hat w_t - w_t)^2]$ is controlled. In particular, an estimation rate of $\tilde O(n^{1/5} [\TV_1(\bs w_{1:n})]^{2/5})$ is shown to be attainable for the squared loss ($\tilde O(\cdot)$ hides poly-logarithmic factors of $n$). On the other hand, squared error losses are also exp-concave in a compact domain. \citet{Baby2021OptimalDR} proposes algorithms for controlling dynamic regret under exp-concave losses. Applying their algorithm will lead to an estimation error of $\tilde O(n^{1/3}[\TV_0(\bs w_{1:n})]^{2/3})$. However, there can be scenarios where the rate of $\tilde O(n^{1/5} [\TV_1(\bs w_{1:n})]^{2/5})$ can be faster than the rate of $\tilde O(n^{1/3}[\TV_0(\bs w_{1:n})]^{2/3})$. For instance, consider the canonical (and practically relevant) example in Fig.\ref{fig:illustration}. Let $\bs w_{1:n}$ be generated by discretizing the function in the left panel at a resolution $1/n$.  We can see that $\TV_0(\bs w_{1:n}) \le 1 = O(1)$ and $\TV_1(\bs w_{1:n}) \approx 2.5 = O(1)$ (for $n \ge 10)$. Here, the aforementioned results from stochastic non-parametric regression can yield a rate of $\tilde O(n^{1/5})$ while existing state-of-the-art results from adversarial online learning can only lead to $O(n^{1/3})$ rate of estimation. We refer the reader to Remark \ref{rmk:fast} for a discussion about more such examples.

\paragraph{Central question and summary of results.} A natural question that we ask here is:

\begin{center}
\quad
\textsf{Can we attain a universal dynamic regret (Eq.\eqref{eq:dregret}) of $\tilde O^*(n^{1/5}[\TV_1(\bs w_{1:n})]^{2/5})$  when the losses are exp-concave without imposing any stochastic assumptions?}
\end{center}

Here $O^*$ hides dimension dependencies. We remark that the rate of $\tilde O(n^{1/5} [\TV_1(\bs w_{1:n})]^{2/5})$ is faster than $\tilde O(n^{1/3}[\TV_0(\bs w_{1:n})]^{2/3})$ iff $\TV_1(\bs w_{1:n}) = O\left( n^{1/3} [\TV_0(\bs w_{1:n})]^{5/3} \right)$. In what follows,  we refer to this regime as the \textbf{low} $\mathcal{\bs{TV}}_1$ \textbf{regime}. We emphasize that this regime is not too restrictive as many different examples can be encompassed by it (see for eg. Fig.\ref{fig:illustration} and Remark \ref{rmk:fast}). A starting point in answering our central question is to observe that a sequence will have low $\TV_1$ distance if it exhibits a piece-wise linear structure across each coordinate and the number of linear sections (or kinks) is sparse. A sequence that is linear across each coordinate within some interval can be perfectly described using a \emph{fixed} vector $\bs u \in \mathbb{R}^{2d}$ where $\bs u[2k-1:2k] \in \mathbb{R}^2$ specifies the slope and intercept along coordinate $k \in [d]$. We will call such $\bs u$ to be a \emph{linear predictor}. If an algorithm guarantees that its \emph{static regret} against fixed linear predictors within \emph{any} interval is controlled, one can hope to perform nearly as well as the comparator sequence with low enough $\TV_1$. This is precisely an application of Strongly Adaptive algorithms \citep{hazan2007adaptive,daniely2015strongly,koolen2016specialist,Cutkosky2020ParameterfreeDA} which aim to control their static regret in any interval and hence we can use them off-the-shelf to achieve our goal. We refer the reader to Section \ref{sec:alg} for more details. Below, we briefly summarize our contributions:

\begin{itemize}
    \item We show that by using appropriate Strongly Adaptive algorithms, one can attain the (near) \emph{optimal} universal dynamic regret rate of $\tilde O^* \left ( \min\{n^{1/5} [\TV_1(\bs w_{1:n})]^{2/5} , n^{1/3} [\TV_0(\bs w_{1:n})]^{2/3}\}\right)$ (Theorem \ref{thm:main-d} and Proposition \ref{prop:low-order}; $a \vee b = \max\{ a,b\}$) whenever the comparators $\bs w_{1:n} \in \mathcal{TV}^{(1)}(C_n)$ and the losses are exp-concave and gradient smooth (see Section \ref{sec:main} for the list of Assumptions and associated definitions). Further this rate is attained \emph{without prior knowledge} of the path lengths $\TV_1(\bs w_{1:n})$ and $\TV_0(\bs w_{1:n})$.
    
    \item To the best of our knowledge, we are the \emph{first} to introduce path variationals based on second order differences to the setting of \emph{adversarial} online learning. We show how to import the \emph{fast rate} phenomenon observed in stochastic non-parametric regression problem under squared loss into the problem of controlling \emph{universal} dynamic regret under general exp-concave losses with no stochastic assumptions.
\end{itemize}

Even though in our proofs, we analyse the KKT conditions of an offline optimization problem akin to the spirit of \citet{Baby2021OptimalDR}, this similarity is only superficial. The offline optimization problem analysed in this work is different from what is considered in \citet{Baby2021OptimalDR}. So the KKT conditions, regret decomposition and the proof strategies we use are also different. Further, we introduce several new non-trivial ideas and generalizations (see Section \ref{sec:1d} and Appendix \ref{app:over}) while exploiting the smoothness of sequences with low $\TV_1$ distance to attain the challenging goal of deriving \emph{faster} (in comparison to \citet{Baby2021OptimalDR}) \emph{universal} dynamic regret rates.

Before we end this section, we briefly describe how the present work provide a new direction in the research thread of dynamic regret minimization.

\textbf{Notes on general outlook and potential impact.} Any meaningful dynamic regret bound has to be parameterized by the particular comparator sequence to avoid a trivial linear regret, i.e., 
$$
\sum_{t=1}^n f_t(\bs p_t) - f_t(\bs w_t) \le \mathrm{DynamicRegretBound}(\bs w_1,\dots,\bs w_n).
$$

Almost all existing dynamic regret bounds are parameterized by the movement costs --- a functional of consecutive differences of the comparator sequence. In particular, we can define $V_{p,q} :=\sum_{t=2}^n\|\bs w_{t}- \bs w_{t-1}\|_q^p$. This includes almost all existing variations, e.g., pathlength ($p=1,q=2$), square pathlength ($p=2,q=2$), total variation ($p=1,q=1$), number of changes ($p>0,q=0$) and so on. The optimal universal dynamic regret for each functional under different loss classes are now well-known. 

While it appears to be a complete story if we roll back to the general problem, there should be many other ways we can parameterize the $\mathrm{DynamicRegretBound}(\bs w_1,..., \bs w_n)$ by exploiting other structures of the comparator sequence than via the movement costs vector. This work can be thought of as the first one to resort to this idea. We reveal that one can attain faster rates by exploiting the smoothness / regularity (in terms of piece-wise polynomial structures) in the comparator sequence. 
As an example, when the first derivative (aka the $\mathcal{TV}_1$ distance) of the sequence has bounded variation, we show that the dynamic regret improves to $O(n^{1/5})$. This idea traces back to the nonparametric regression literature, where the higher order smoothness of functions are often used. It is our hope that this work can inspire further collaborations between researchers in these two (mostly disparate) communities of online learning and offline non-parametric regression.

We remark that our work only touches the surface of the idea of characterizing dynamic regret via interesting smoothness metrics of the comparator sequence. Indeed, there are other interesting functionals of the comparator sequence that we can exploit, e.g., periodicity, smoothness in an appropriately transformed domain and so on. We believe this is an interesting future direction for researchers working in dynamic regret minimisation. The present work takes only the first steps towards realising this bigger goal.

\section{RELATED WORK} \label{sec:lit}
Here we recall the most relevant works. The work of \citet{baby2020higherTV} aims at controlling Eq.\eqref{eq:dregret} under squared error when noisy realizations of a $\TV_1$ bounded sequence is revealed sequentially. For this setting, they propose an algorithm namely AdaVAW that combines Vovk-Azoury-Warmuth forecaster with wavelet denoising which relies strongly on the iid noise assumption and losses being squared error. The absence of such stochastic assumptions and handling general exp-concave losses in our setting poses a significant challenge in controlling the dynamic regret. Overall, we can conclude that results in the current paper dominates that of AdaVAW for TV order $k = 1$. As mentioned in Section \ref{sec:intro}, the work of \citet{Baby2021OptimalDR} fails to attain optimal regret rate for the current problem. We refer the readers to Appendix \ref{app:compare} for a detailed description on why the analysis of \citet{Baby2021OptimalDR} fails to attain optimal regret rate in our setting where the comparators has low $\TV_1$ distance. \citet{baby2021TVDenoise} reported experiments where they use a Strongly Adaptive algorithm for competing against best linear predictor in each time window for the task of forecasting COVID-19 cases. This method was shown to empirically out-perform state-of-the-art trend forecasting strategies. However, they didn't provide analysis for this strategy while our work supplements it with necessary theoretical grounding albeit with a slightly different Strongly Adaptive algorithm. Apart from these works, there is a rich body of literature on dynamic regret minimization such as \citep{jadbabaie2015online,yang2016tracking,Mokhtari2016OnlineOI,chen2018non,zhang2018adaptive,zhang2018dynamic,yuan2019dynamic,goel2019OBD,arrows2019,Zhao2020DynamicRO,Zhao2021StronglyConvex,Zhao2021Memory,baby2022optimal,jacobsen2022free,baby2022optimalLQR,zhang2023unconstrainedDR}. However, to the best of our knowledge none of these works are known to attain the optimal dynamic regret rate for our setting. An  elaborate literature survey is deferred to Appendix \ref{app:lit}.


\section{THE ALGORITHM} \label{sec:alg}

\begin{figure}[h!]
	\centering
	\fbox{
		\begin{minipage}{8 cm}
		FLH-SIONS: inputs: exp-concavity factor $\sigma$ and $n$ SIONS base learners $E^1,\ldots,E^n$ initialized with parameters $\epsilon = 2$, $\eta = \sigma$ and $C=20$. (see Fig. \ref{fig:sions})
            \begin{enumerate}
                \item For each $t$, $v_t = (v_t^{(1)},\ldots,v_t^{(t)})$ is a probability vector in $\mathbb{R}^t$. Initialize $v_1^{(1)} = 1$.
                \item For any SIONS expert $E_j$ with $j \le t$, define $\bs x_j^{(t)} = [1,t-j+1]^T$ to be given to $E_j$ at time $t$ before making its prediction $E_j(t) \in \mathbb{R}^d$.
                \item In round $t$, set $\forall j \le t$, $\bs y_t^j \leftarrow E_j(t)$ (the prediction of the $j^{th}$ base learner at time $t$). Play $\bs p_t =  \sum_{j=1}^t v_t^{(j)}\bs y_t^{(j)}$.
                \item After receiving $f_t$, set $\hat v_{t+1}^{(t+1)} = 0$ and perform update for $1 \le i \le t$:
                \begin{align}
                    \hat v_{t+1}^{(i)}
                    &= \frac{v_t^{(i)}e^{-\sigma f_t(\bs x_t^{(i)})}}{\sum_{j=1}^t v_t^{(j)}e^{-\sigma f_t(\bs x_t^{(j)})}}
                \end{align}
                \item Addition step - Set $v_{t+1}^{(t+1)}$ to $1/(t+1)$ and for $i \neq t+1$:
                \begin{align}
                    v_{t+1}^{(i)}
                    &= (1-(t+1)^{-1}) \hat v_{t+1}^{(i)}
                \end{align}
            \end{enumerate}
		\end{minipage}
	}
	\caption{FLH algorithm of \citet{hazan2007adaptive} with SIONS (see Fig.\ref{fig:sions}) base experts}
	\label{fig:flh}
\end{figure}

\begin{figure}[h!]
	\centering
	\fbox{
		\begin{minipage}{8 cm}
		SIONS: inputs: exp-concavity factor $\eta$, $\epsilon > 0$ and $C>0$.
            \begin{enumerate}
            
                \item For any round $t$, we define $\tilde f_t(\bs v) = f_j(\bs x_t^T \bs v[1:2], \bs x_t^T \bs v[3:4],\ldots,\bs x_t^T \bs v[2k-1:2k])$ for any vector $\bs v \in \mathbb{R}^{2d}$.
                
                \item At round $t+1$:
                \begin{enumerate}
                    \item Receive co-variate $\bs x_{t+1} \in \mathbb{R}^2$.
                    \item Let $\cK_{t+1} = \{\bs w \in \mathbb{R}^{2d}: |\bs x_{t+1}^T \bs w[2k-1:2k]| \le C \text{ for all } k \in [d]  \}$.                
                    \item Let $\bs A_t =\epsilon \bs I_{2d} + \eta \sum_{j=1}^t \grad \tilde f_j(\bs v_j) \grad \tilde f_j(\bs v_j)^T$.
                    \item Let $\bs u_{t+1} = \bs v_t - \bs A_t^{-1} \grad \tilde f_t(\bs v_t)$.
                    \item Let $\bs v_{t+1} = \argmin_{\bs w \in \cK_{t+1}} \|\bs w - \bs u_{t+1} \|_{\bs A_t}$.
                    \item Play $\bs w_{t+1} \in \mathbb{R}^{d}$ such that $\bs w_{t+1}[k] = \bs x_{t+1}^T\bs v_{t+1}[2k-1:2k]$ for all $k \in [d]$.
                \end{enumerate}
            \end{enumerate}
		\end{minipage}
	}
	\caption{An instance of SIONS algorithm from \citet{Luo2016EfficientSO}.}
	\label{fig:sions}
\end{figure}

In this section, we formally describe the main algorithm FLH-SIONS (Follow the Leading History-Scale Invariant Online Newton Step) in Fig.\ref{fig:flh} and provide intuition on why it can favorably control the dynamic regret against $\TV_1$ bounded comparators.
For the sake of simplicity, we capture the intuition in a uni-variate setting where the comparators $w_t \in \cW \subset \mathbb{R}$ for all $t \in [n]$.

\begin{definition} \label{def:piecelinear}
Within an interval $[a,b]$, we say that the comparator $w_{a:b}$ is a \emph{linear signal} or assumes a \emph{linear structure} if the slope $w_{t+1} - w_t$ is constant for all $t \in [a,b-1]$.
\end{definition}

As described in Section \ref{sec:intro}, we are interested in competing against comparator sequences $w_{1:n}$ that have a piece-wise linear structure (across each coordinate in multi-dimensions). The durations / intervals of $[n]$ where the comparator is a fixed linear signal is unknown to the learner. Suppose that an ideal oracle provides us with the exact locations of these intervals of $[n]$. Consider an interval $[a,b]$ provided by the oracle where the comparator has a fixed linear structure given by $w_t = \bs \mu^T \bs x_a^{(t)}$ for the \emph{co-variates} $\bs x_a^{(t)} := [1,t-a+1]^T$ and $\bs \mu$ such that $|w_t|$ is $O(1)$ bounded for all $t \in [a,b]$. An effective strategy for the learner is to deploy an online algorithm $E_a$ that starts from time $a$ such that within the interval $[a,b]$ its regret:
\begin{align}
    R_{[a,b]}(\bs \mu)
    &:= \sum_{t=a}^b f_t\left(E_a(t) \right) - f_t(\bs \mu^T \bs x_a^{(t)})
\end{align}
is controlled. Here $E_a(t)$ is the predictions of the algorithm $E_a$ at time $t$. Under exp-concave losses, an $O(\log n)$ bound on the above regret can be achieved by the SIONS algorithm (Fig.\ref{fig:sions})  from (\citet{Luo2016EfficientSO}, Theorem 2) run with co-variates $\bs x_a^{(t)}$.

In practice, the locations of such ideal intervals are unknown to us. So we maintain a pool of $n$ base SIONS experts in Fig.\ref{fig:flh} where the expert $E_\tau$ starts at time $\tau$ with the monomial co-variate $\bs x_\tau^{(t)} = [1,t-\tau+1]^T$ for all $t \ge \tau$. The adaptive regret guarantee of FLH with exp-concave losses (due to \citet{hazan2007adaptive}, Theorem 3.2) keeps the regret wrt \emph{any} base expert to be small. In particular, FLH-SIONS satisfies that
\begin{align}
    \sum_{t=\tau}^j f_t(p_t) - f_t\left(E_\tau(t)\right) = O(\log n),
\end{align}
where $p_t$ are the predictions of FLH-SIONS and $j \ge \tau$ for \emph{any} $\tau \in [n]$. Hence for the interval $[a,b]$ given by the ideal oracle, it follows that
\begin{align}
    \sum_{t=a}^b f_t(p_t) - f_t(\bs \mu^T \bs x_a^{(t)})
    &\le \sum_{t=a}^b f_t\left(E_a(t) \right) - f_t(\bs \mu^T \bs x_a^{(t)})\\
    &+ O(\log n)\\
    &= R_{[a,b]}(\bs \mu) + O(\log n)\\
    &= O(\log n),\label{eq:sa}
\end{align}
where in the last equation, we appealed to the logarithmic static regret of SIONS from  \citet{Luo2016EfficientSO}. As a minor technical remark, we note that the original results of \citet{Luo2016EfficientSO} assume that the losses are of the form $\tilde f_j(\bs w) = f_j(\bs x_j^T \bs w)$ for a uni-variate function $f_j$. However, we show in Lemma \ref{lem:SON-stat} (in Appendix) that their regret bounds can be straightforwardly extended to handle multivariate losses $f_j$ as in Line 1 of Fig.\ref{fig:sions} which is useful in our multi-dimensional setup.

Thus ultimately, the regret of the FLH-SIONS procedure is well controlled within each interval provided by the ideal oracle, thus allowing us to be competent against the piece-wise linear comparator. We remark that while both FLH and SIONS are well-known existing algorithms, our use of them with monomial co-variates is new. Our dynamic regret analysis is new too, which uncovers previously unknown properties of a particular combination of these existing algorithmic components using novel proof techniques.

\section{MAIN RESULTS} \label{sec:main}
In this section, we explain the assumptions used and the main results of this paper. Then we provide a brief proof sketch for Theorem \ref{thm:main-d} in a uni-variate setting highlighting the technical challenges overcome along the way. The case of multiple dimensions is handled by constructing suitable reductions that will allow us to re-use much of the analytical machinery developed for the case of uni-variate setting. For the sake of clarity, we present a detailed overview of our proof strategy in Appendix \ref{app:over}. The following are the assumptions made.

\begin{description}
\item[A1.] For all $t \in [n]$, the comparators $\bs w_t$ belongs to a given benchmark space $\cW \subset \mathbb{R}^d$. Further we have $\cW \subseteq [-1,1]^d$.
\item[A2.]The loss function $f_t:\mathbb{R}^d \rightarrow \mathbb{R}$ revealed at time $t$ is $1$-Lipschitz in $\| \cdot \|_2$ norm over the interval $[-20,20]^d$.
\item[A3.]The losses $f_t$ are $1$-gradient Lipschitz over the interval $[-20,20]^d$. This implies that $f_t(\bs y) \le f_t(\bs x) + \grad f_t(\bs x )^T (\bs y- \bs x) + \frac{1}{2} \|(\bs y- \bs x)\|_2^2$ for all $\bs x,\bs y \in [-20,20]^d$.
\item[A4.] The losses $f_t$ are $\sigma$ exp-concave over $[-20, 20]^d$. This implies that $f_t(\bs y) \ge f_t(\bs x) + \grad f_t(\bs x)^T (\bs y-\bs x) + \frac{\sigma}{2} \left( \grad f_t(\bs x)^T (\bs y-\bs x) \right)^2$ for all $\bs x,\bs y \in [-20, 20]^d$.
\end{description}

Assumptions A3 and A4 ensure the smoothness and curvature of the losses which we crucially rely to derive fast regret rates. Assumptions about Lipschitzness as in A2 are usually standard in online learning. In assumption A1 we consider comparators that belong to an interval that is smaller than the intervals in other assumptions. This is due to the fact that we allow our algorithms to be improper in the sense that the decisions of the algorithm may lie outside the benchmark space $\cW$.

We start with a lower bound on the dynamic regret (Eq.\eqref{eq:dregret}) which is obtained by adapting the arguments in \citet{donoho1998minimax} to the case of bounded sequences as in Assumption A1. See Appendix \ref{app:lb} for a proof.
\begin{restatable}{proposition}{proplb}\label{prop:lb}
Under Assumptions A1-A4, any online algorithm necessarily suffers $\sup_{\bs w_{1:n} \text{ with } \TV_1(\bs w_{1:n}) \le C_n }R_n(\bs w_{1:n}) = \Omega(d^{3/5} n^{1/5}C_n^{2/5} \vee d)$.
\end{restatable}
We have the following guarantee for FLH-SIONS.
\begin{theorem} \label{thm:main-d}
Let $\bs p_t$ be the predictions of FLH-SIONS algorithm with parameters $\epsilon = 2$, $C = 20$ and exp-concavity factor $\sigma$. Under Assumptions A1-A4, we have that,
$$
    \sum_{t=1}^n f_t(\bs p_t) - f_t(\bs w_t)
    = \tilde O (d^2 n^{1/5} [\TV_1(\bs w_{1:n})]^{2/5}  \vee d^2 ),\nonumber
$$
where $\tilde O$ hides poly-logarithmic factors of $n$ and $a \vee b = \max\{ a,b\}$.
\end{theorem}
\begin{remark}
Compared with the lower bound in Proposition \ref{prop:lb}, we conclude that the regret rate of the above theorem is optimal modulo factors of $d$ and $\log n$.
\end{remark}

\begin{proposition} \label{prop:low-order}
It can be shown that the same algorithm FLH-SIONS under the setting of Theorem \ref{thm:main-d} enjoys a regret rate of $\tilde O(n^{1/3} [\TV_0(\bs w_{1:n})]^{2/3})$ as well. This result is a straight-forward consequence of the arguments in \citet{Baby2021OptimalDR} and summarized in Appendix \ref{app:tv0}. When combined with Theorem \ref{thm:main-d} we conclude that under Assumptions A1-A4, FLH-SIONS attains an adaptive guarantee of
\begin{align}
    \sum_{t=1}^n f_t(\bs p_t) - f_t(\bs w_t)
    &= \tilde O \Bigg ( \Bigg.d^2 \Bigg( \Bigg. \left(n^{1/5} [\TV_1(\bs w_{1:n})]^{2/5} \right)\\
    &\wedge \left(n^{1/3} [\TV_0(\bs w_{1:n})]^{2/3}\right) \Bigg. \Bigg)  \vee d^2 \Bigg. \Bigg),\nonumber    
\end{align}

for any comparator sequence $\bs w_{1:n}$. Here $\tilde O$ hides poly-logarithmic factors of $n$, $a \vee b = \max\{ a,b\}$ and $a \wedge b = \min\{ a,b\}$.
\end{proposition}

From the above Proposition, we see that FLH-SIONS has the nice property of safe-guarding the regret in case the $\TV_1(\bs w_{1:n})$ distance of the comparator doesn't fall in the low $\TV_1$ regime defined in Section \ref{sec:intro}.

\begin{remark} \label{rmk:inst}
We note that the upper bound in Proposition \ref{prop:low-order} does not contradict the lower bound in Proposition \ref{prop:lb}. The lower bound holds in a worst case sense while the upper bound in Proposition \ref{prop:low-order} is instance-dependant and can sometimes be faster than the worst case rate in Proposition \ref{prop:lb}. Indeed, for the hard comparator sequence $\bs w_{1:n}$ we construct in Appendix \ref{app:lb}, the rates $n^{1/5}[\TV_1(\bs w_{1:n})]^{2/5}$ and $n^{1/3} [\TV_0(\bs w_{1:n})]^{2/3}$ are of the same order.
\end{remark}

\begin{remark} \label{rmk:fast}
We can construct many examples of sequences $\bs w_{1:n}$ that fall in the low $\TV_1$ regime besides the one in Fig.\ref{fig:illustration}, such that the rate of $n^{1/5} [\TV_1(\bs w_{1:n})]^{2/5} \vee 1$ is faster than the rate $n^{1/3} [\TV_0(\bs w_{1:n})]^{2/3} \vee 1$. We defer a non-exhaustive list of such examples to Appendix \ref{app:example}.
\end{remark}

\begin{remark}\label{rmk:linsmooth}
One may ask if a simpler algorithm such as carefully tuned online gradient descend (OGD) can enjoy these fast rates too. However, Proposition 2 of \citet{baby2020higherTV} implies that properly tuned OGD algorithm which is optimal against set of all comparators $\bs w_{1:n}$ with $\TV_0(\bs w_{1:n}) \le 1$ under convex losses, necessarily suffers a slower dynamic regret of $\Omega(n^{1/4})$ against set of all comparators with $\TV_1(\bs w_{1:n}) \le 1$ under exp-concave losses (see Lemma \ref{lem:linsmooth} in Appendix \ref{app:lit}).
\end{remark}

\begin{remark}[\textbf{optimized implementation}]
In the presented form of FLH-SIONS, we need to hedge over $O(n)$ SIONS base learners per round to make predictions. However, this can be reduced to hedging over $O(\log n)$ base learners per round by using AFLH from \citet{hazan2007adaptive} or PAE from \citet{Zhang2021DualAA} as the aggregation algorithm. This route will help to achieve a near-linear overall complexity of $O(n \log n)$ calls to the base learners. The cost of doing this is that it enlarges the dynamic regret bound by a factor of $O(\log n)$. Moreover, the run-time of the SIONS base learners can be further ameliorated at the expense of slightly increasing the regret bound by using randomized sketchings as described in \citet{Luo2016EfficientSO}. It is also possible to specify custom base learners in the case we know the form of losses ahead of time. If the base learners incur logarithmic static regret under the specified loss, then we can enjoy the regret rate specified in Proposition \ref{prop:low-order}. For example, if the losses are linear regression type losses, one can use Vovk-Azoury-Warmuth forecaster \citep{BianchiBook2006} as base learners. If the losses are logistic regression losses, one can use the algorithm in \citet{jezequel2020logistic} as base learners. Such custom base learners can have much lower run-time than SIONS base learners which are designed to support the fully general exp-concave losses. We also note that losses such as linear and logistic regression losses are commonly used in practice as well.
\end{remark}

\subsection{Proof Summary of Theorem \ref{thm:main-d} for one dimension} \label{sec:1d}

In what follows, we present several useful lemmas and provide a running sketch on how to chain them to arrive at Theorem \ref{thm:main-d} in a uni-variate setting (i.e $d=1$). Detailed proofs are deferred to Appendix \ref{app:1d}.

Suppose that we need to compete against comparators whose $\TV_1$ distance (i.e $n \|D^2 w_{1:n} \|_1$) is bounded by some number $C_n$. This quantity could be unknown to the algorithm. Consider the \emph{offline oracle} who has access to the entire sequence of loss functions $f_1,\ldots,f_n$ and the $\TV_1$ bound $C_n$. It may then solve for the strongest possible comparator respecting the $\TV_1$ bound through the following convex optimization problem.

\begin{mini!}|s|[2]                   
    {\tilde u_1,\ldots,\tilde u_n}                               
    {\sum_{t=1}^n f_t(\tilde { u}_t)}   
    {\label{eq:obj}}             
    {}                                
    \addConstraint{\|D^2 \tilde u_{1:n} \|_1}{\le C_n/n, \label{eq:constr-ec-1}}  
    \addConstraint{-1}{\le \tilde u_t \: \forall t \in [n],\label{eq:constr-ec-2}}
    \addConstraint{\tilde u_t}{\le 1 \: \forall t \in [n],\label{eq:constr-ec-3}}
\end{mini!}
Let $u_1,\ldots,u_n$ be the optimal solution of the above problem. This sequence will be referred as \textbf{offline optimal} hence-forth. Clearly we have that the regret against any comparator sequence $\bs w_{1:n}$ with $\TV_1(\bs w_{1:n}) \le C_n$ obeys
\begin{align}
 \sum_{t=1}^n f_t(p_t) - f_t(w_t) 
 &\le \sum_{t=1}^n f_t(p_t) - f_t(u_t),
\end{align}
and hence it suffices to bound the right side of the above inequality.

Next, we provide a partition of the horizon with certain useful properties.
\begin{restatable}{lemma}{lemkeypart}(\textbf{key partition}) \label{lem:keypart}
For some interval $[a,b] \in [n]$, define $\ell_{a \rightarrow b} := b-a+1$. There exists a partitioning of the time horizon $\cP := \{[1_s,1_t], \ldots, [i_s,i_t], \ldots [M_s,M_t] \}$  where $M = |\cP|$ such that for any bin $[i_s,i_t] \in \cP$ we have: 1) $\|D^2 u_{i_s:i_t}\|_1 \le 1/\ell_{i_s \rightarrow i_t}^{3/2}$; 2) $\|D^2 u_{i_s:i_t+1}\|_1 > 1/\ell_{i_s \rightarrow i_t+1}^{3/2}$ and 3) $M = O \left( n^{1/5}C_n^{2/5} \vee 1\right)$.
\end{restatable}

Going forward, the idea is to bound the dynamic regret within each bin  in $\cP$ by an $\tilde O(1)$ quantity. Then we can add them up across all bins to arrive at the guarantee of Theorem \ref{thm:main-d} (with $d=1$). We pause to remark that eventhough this high-level idea resembles to that of \citep{Baby2021OptimalDR}, the underlying details of our analysis to materialize this idea requires highly non-trivial deviations from the path followed by \citep{Baby2021OptimalDR}.

First, we need some definitions. Consider a bin $[i_s,i_t] \in \cP$ with length at-least 2. Let's define a co-variate $\bs x_j := [1,j-i_s+1]^T$. Let $\bs X^T := [\bs x_{i_s},\ldots,\bs x_{i_t}]$ be the matrix of co-variates and $u_{i_s:i_t} := [u_{i_s},\ldots,u_{i_t}]^T$. Let $\bs \beta = \left(\bs X^T \bs X \right)^{-1}\bs X^T u_{i_s:i_t}$ be the least square fit coefficient computed with co-variates $\bs x_j$ and labels $u_j$. Define a second moment matrix $\bs A = \sum_{j=i_s}^{i_t} \bs x_j \bs x_j^T$. Let $\bs \alpha := \bs \beta - \bs A^{-1} \sum_{j=i_s}^{i_t} \grad f_j(\bs \beta^T \bs x_j) \bs x_j$. ($\bs A^{-1}$ is guaranteed to exist when length of the bin is at-least 2). We remind the reader that $\grad f_j(\bs \beta^T \bs x_j)$ is a scalar as we consider uni-variate $f_j$ in this section.

We connect these quantities via a \emph{key regret decomposition} as follows:
\begin{align}
    &\sum_{j=i_s}^{i_t} f_j(p_j) - f_j(u_j)
    =
    \underbrace{\sum_{j=i_s}^{i_t} f_j(p_j) - f_j(\bs \alpha^T \bs x_j)}_{T_1}+\\
    &\underbrace{\sum_{j=i_s}^{i_t} f_j(\bs \alpha^T \bs x_j) - f_j(\bs \beta^T \bs x_j)}_{T_2}+\underbrace{\sum_{j=i_s}^{i_t} f_j(\bs \beta^T \bs x_j) - f_j(u_j)}_{T_3}\\ \label{eq:main-decomp}
\end{align}

It can be shown that $| \bs \alpha^T \bs x_j| \le 20 = O(1)$. Hence the term $T_1$ can be controlled by an $O(\log n)$ bound due to Strong Adaptivity of FLH-SIONS as described in Section \ref{sec:alg}, Eq.\eqref{eq:sa}. The quantity $\bs \alpha$ is obtained via moving in a direction reminiscent to that of Newton method. This is in sharp contrast to the one step gradient descent update used in \citet{Baby2021OptimalDR}.  More precisely, consider the function $F(\bs \beta) = \sum_{j=i_s}^{i_t} f_j(\bs \beta^T \bs x_j)$. Then $\bs \alpha = \bs \beta - \bs A^{-1} \grad F(\bs \beta)$. By exploiting gradient Lipschitzness of $f_j$, the correction matrix $\bs A$ can be shown to satisfy the Hessian dominance $\grad^2 F(\bs \beta) \preccurlyeq \bs A$. This Newton style update is shown to keep the term $T_2$ to be negative through the following generalized descent lemma:
\begin{restatable}{lemma}{lemND} \label{lem:nd}
We have that $T_2
\le -\frac{1}{2} \left \| \grad F(\bs \beta) \right  \|_{\bs A^{-1}}^2$.
\end{restatable}

The negative descent term displayed in the above Lemma is similar to the standard (squared) Newton decrement \citep{Nesterov2004IntroductoryLO} in the sense that it is also influenced by the local geometry through the norm induced by the inverse correction matrix $\bs A^{-1}$.

We then proceed to show that the negative $T_2$ can diminish the effect of $T_3$ by keeping $T_2+ T_3$ to be an $O(1)$ quantity. Thus the dynamic regret within the bin $[i_s,i_t] \in \cP$ is controlled to $\tilde O(1)$. Adding the bound across all bins in $\cP$ from Lemma \ref{lem:keypart} yields Theorem \ref{thm:main-d} in one dimension. 

A major challenge in the analysis is to prove that the term $T_2 + T_3 = O(1)$ without imposing restrictive assumptions such as Self-Concordance or Hessian Lipschitnzess as in the classical analysis of Newton method (see for eg.\citep{Nesterov2004IntroductoryLO}). In the rest of this section, we outline the arguments leading to this result.

\begin{restatable}{lemma}{lemmain} \label{lem:main}
We have that $T_2 + T_3 = O(1)$ where $T_2$ and $T_3$ are as defined in Eq.\eqref{eq:main-decomp}
\end{restatable}

Here the main idea is to exploit the KKT conditions of the offline optimization problem in Eq.\eqref{eq:obj} to show $T_2 + T_3 = O(1)$ even if $|T_2|$ and $|T_3|$ can be very large individually. Though this is similar to the observation in \citet{Baby2021OptimalDR}, our regret decomposition in Eq.\eqref{eq:main-decomp} and the KKT conditions (see Lemma \ref{lem:kkt} in Appendix \ref{app:over}) are different. So showing this result requires non-trivial deviations from the proof of \citet{Baby2021OptimalDR}.  Importantly, it was not apriori clear that $T_2 + T_3$ can be possibly bound by $O(1)$ for the current problem. The key novelty is that we bound $T_2+T_3$ by introducing an auxiliary function that is concave in its arguments which allows us to systematically explore the properties of its maximizers. We refer the reader to Appendix \ref{app:over} for a thorough overview on the construction and use of such auxiliary functions in proving the lemma.

As discussed before, the case of multiple dimensions is handled by constructing suitable reductions that will allow us to re-use much of the analytical machinery developed for the case of uni-variate setting. We refer the reader to Appendix \ref{app:over} for an overview of the details of such reductions. However, we emphasize that this reduction happens only in the analysis, and we \emph{do not} run $d$ uni-variate FLH-SIONS algorithms for handling multi-dimensions (see Theorem \ref{thm:main-d}).

We conclude this section by noting that our proof techniques also lead to a dynamic regret bound that \emph{simultaneously} hold for any sub-interval of $[n]$.

\begin{remark}
Assume the notations used in Lemma \ref{lem:keypart} and Proposition \ref{prop:low-order}. Consider an interval $[a,b] \subseteq [n]$. By partitioning this sub-interval as per Lemma \ref{lem:keypart} and applying the regret decomposition of Eq.\eqref{eq:main-decomp}, one can show the following dynamic regret bound over the interval $[a,b]$:
\begin{align}
    \sum_{t=a}^b f_t(\bs p_t) - f_t(\bs w_t)
    &= \tilde O \Bigg ( \Bigg.d^2 \Bigg( \Bigg. \left(\ell^{1/5} \left(\ell \|D^2 \bs w_{a:b}\|_1\right)^{2/5} \right)\\
    &\wedge \left(\ell^{1/3} \|D^1\bs w_{a:b} \|_1^{2/3}\right) \Bigg. \Bigg)  \vee d^2 \Bigg. \Bigg),\nonumber
\end{align}
where $\ell:=\ell_{a \rightarrow b}$.

\end{remark}



\section{CONCLUSION}
In this work, we derived universal dynamic regret rate parametrized by a \emph{novel} second-order path variational of the comparators. Such a path variational naturally captures the piecewise linear structures of the comparators and can be used to flexibly model many practical non-stationarities in the environment. Our results for the exp-concave losses achieved an adaptive universal dynamic regret of $\tilde O^* \left ( \min\{n^{1/5} [\TV_1(\bs w_{1:n})]^{2/5}, n^{1/3} [\TV_0(\bs w_{1:n})]^{2/3}\}  \vee 1 \right)$ which matches the minimax lower bound up to a factor that depends on $d$ and $\log n$. This is the first result of such kind in the adversarial setting and the first that works with general exp-concave family of losses. We conjecture that a similar algorithm as in Fig.\ref{fig:flh} based on degree $k$ monomial co-variates $[1,t,\ldots,t^k]$ can lead to optimal dynamic regret in terms of $\TV_k$.

\section*{Acknowledgements}
The authors are grateful to the anonymous reviewers whose constructive comments helped to enhance the paper. The work is partially supported by NSF Awards \#2007117 and a gift from Amazon.

\bibliography{tf,yx}
\bibliographystyle{plainnat}

\newpage

\onecolumn

\appendix

\section{More on Related Work} \label{app:lit}

In this section, we elaborate upon the related works mentioned in Section \ref{sec:lit}. We inherit all the notations and terminologies introduced in Section \ref{sec:intro}.

\paragraph{Dynamic regret against $\mathcal{TV}^{(1)}(C_n)$ in stochastic setting.} Perhaps, the most relevant to our work is that of \citet{baby2020higherTV}. They consider an online protocol where at each round, the learner makes a prediction $\hat \theta_t \in \mathbb{R}$. Then a label $y_t = \theta_t + \cN(0,1)$ is revealed. They assume that the ground truth sequence $\theta_{1:n} \in \mathcal{TV}^{(k)}(C_n)$ (see Eq.\eqref{eq:tvk}). The goal of the learner is control the expected cumulative squared error of the learner namely $\sum_{t=1}^n (\hat \theta_t - \theta_t)^2$. In this setting, they propose policies that can attain a near optimal estimation error of $\tilde O(n^{\frac{1}{2k+3}} C_n^{\frac{2}{2k+3}})$ for any $k > 0$. In retrospect, in this work we consider the case where comparators belong to a $\mathcal{TV}^{(1)}(C_n)$ class (i.e, with k=1). Further, the absence of stochastic assumptions in our setting poses a significant challenge in controlling the universal dynamic regret (Eq.\eqref{eq:dregret}).

\paragraph{Restricted dynamic regret minimization.} In this line of work, we consider a similar learning setting as mentioned in Section \ref{sec:intro}. However, the goal of the learner is to control the dynamic regret against point-wise minimizers:
\begin{align}
    R_{n,\text{restrict}}
    &= \sum_{t=1}^n f_t(\bs p_t) - f_t(\bs w_t^*),
\end{align}
where $\bs w_t^* \in \argmin{x \in \cW} f_t(\bs x)$. When the losses are strongly convex and gradient smooth, \citet{Mokhtari2016OnlineOI} proposes algorithms that can attain a restricted dynamic regret of $O(1+C^*_n)$, where $C_n^* = \sum_{t=1}^{n-1} \|\bs w^*_{t} - \bs w^*_{t+1}\|_2$. However, as noted in \citet{zhang2018adaptive}, such a guarantee can be sometimes overly pessimistic. For example, in the context of statistical learning where the losses are sampled iid from a distribution, the point-wise minimizers can incur a path length $C_n^* = O(n)$ due to random perturbations introduced by sampling.

\paragraph{Universal dynamic regret minimization.} This is the same framework as considered in the introduction. Obtaining universal dynamic regret guarantees is challenging since we need to bound the dynamic regret for \emph{any} comparator sequence from the bench mark set $\cW$ while automatically adapting to their path length. When the losses are convex \citet{zhang2018adaptive,Cutkosky2020ParameterfreeDA} provides an optimal universal dynamic regret of $O(\sqrt{n(1+P_n)})$, where
\begin{align}
 P_n = \sum_{t=1}^{n-1} \|\bs w_{t} - \bs w_{t+1} \|_2. \label{eq:pn}
\end{align}
For $C_n = \Omega(1)$, the embedding $\mathcal{TV}^{(1)}(C_n) \subseteq \mathcal{TV}^{(0)}(\kappa C_n)$ (see Proposition \ref{prop:embed}) where $\kappa$ is a constant doesn't depend on $n$ or $C_n$ implies that work of \citet{zhang2018adaptive} yields a dynamic regret rate of $O^*(\sqrt{nC_n})$ for competing against comparator sequences in $\mathcal{TV}^{(1)}(C_n)$ class. ($O^*$ hides dimension dependencies.) This is a sub-optimal rate when applied to our setting as expected, since they don't assume the losses are exp-concave. 

When the losses are in-fact exp-concave, one can apply the result of \citet{Baby2021OptimalDR} to produce a dynamic regret rate of $\tilde O^*(n^{1/3}C_n^{2/3} \vee 1)$. However, as noted in Section \ref{sec:intro}, this rate is sub-optimal. In Appendix \ref{app:compare}, we give an elaborate description on why their analysis lead to sub-optimal rate in our setting of competing against comparators from $\mathcal{TV}^{(1)}(C_n)$ class.

\paragraph{Dynamic regret based on functional variations.} It is also common to measure the non-stationarity of the environment in terms of the variation incurred by the loss function sequence. Define 
\begin{align}
    D_n
    &= \sum_{t=2}^n \max_{\bs w \in \cW} |f_t(\bs w) - f_{t-1}(\bs w)|.
\end{align}

There are works such as \citep{besbes2015non,yang2016tracking, chen2018non} that aims in controlling the dynamic(restricted / universal) in terms of $D_n$. \citet{jadbabaie2015online} proposes algorithms that can control dynamic regret simultaneously in the terms of $D_n$ and $P_n$.

\paragraph{Static regret minimization.} In classical OCO, a well known metric is to control the static regret of an algorithm namely, $\sum_{t=1}^n f_t(\bs p_t) - f_t(\bs w)$. Algorithms such as Online Gradient Descent \citep{zinkevich2003online} can attain an optimal $O(\sqrt{n})$ static regret when losses are convex. When the losses are exp-concave or strongly convex it is possible to attain logarithmic static regret \citep{hazan2007logregret}. However, static regret is not a good metric for measuring the performance of a leraner in non-stationary environments.

\paragraph{Strongly Adaptive (SA) regret minimization.} This notion of regret is introduced by \citet{daniely2015strongly}. In this framework, the learner aims in controlling its static regret in any local time window as a function of window-length (modulo factors of $\log n$). The algorithms in \cite{daniely2015strongly,Cutkosky2020ParameterfreeDA} provides a static regret of $\tilde O(\sqrt{|I|})$ across any local interval $I$ whenever the losses are convex. When the losses are strongly convex or exp-concave, the algorithms in \citep{hazan2007adaptive,koolen2016specialist,Zhang2021DualAA} yields logarithmic static regret in any local time window when the base learners are chosen appropriately. 

\citet{zhang2018dynamic} shows that SA algorithms can be used to control the dynamic regret in terms of the functional variation $D_n$.

\paragraph{Online non-parametric regression.} In section \ref{sec:intro}, we modelled the dynamics of the comparator sequence as a member of a non-parametric function class. \citet{rakhlin2014online,rakhlin2015online} studies the minimax rate of learning against a non-parametric function class. They establish the right minimax rate (in terms of dependencies wrt $n$) using arguments based on sequential Rademacher complexity in a non-constructive manner. In-fact, their results on Besov spaces imply that the minimax dynamic regret rate for our problem is indeed $O(n^{1/5})$ since the $\mathcal{TV}^{(1)}$ class is sandwiched between two Besov spaces having the same minimax rate (see for eg. \citep{DeVore1993ConstructiveA} and \citep{donoho1998minimax}). There are other line of works that study the non-parametric regression problem against other function classes of interest such as \citet{gaillard2015chaining} (for Holder class of functions), \citet{kotlowski2016} (for isotonic functions) and \citet{koolen2015minimax} (for Sobolev functions). These function classes doesn't capture the $\mathcal{TV}^{(1)}$ class that we study in this work. For instance, the (discretized) higher order Holder and Sobolev classes features sequences that are more regular than that of $\mathcal{TV}^{(1)}$ class (see for example \citet{arrows2019}).

\paragraph{Further explanations on Remark \ref{rmk:linsmooth}.} In rest of this section, we focus on proving Remark \ref{rmk:linsmooth}.

It is sufficient to focus on OGD algorithms with step-size $\eta < 1$. Otherwise if $\eta \ge 1$, one can come up with sequence of losses that can enforce linear regret. An example of this scenario is described as follows:

The loss at time $t$ is given by $f_t(x) = \frac{(x-y_t)^2}{2}$ with $y_t = -1$ at odd rounds and $y_t = 1$ at even rounds. The decision set is given by $\cD=[-1,1]$. We focus on OGD algorithms which plays $0$ at the first round, though similar arguments can be given for any valid initialization point. Suppose the step size is $\eta \ge 1$. The iterate at step $t+1$, denoted by $w_{t+1}$, is maintained as
\begin{align}
    w_{t+1} = \text{clip}_{[-1,1]} \left(w_t - \eta (w_t - y_t)\right),
\end{align}
where clip function clips the argument to $[-1,1]$.

Recall that $\eta \ge 1$. So we have $w_1=0$, $w_2 = \text{clip}_{[-1,1]}(-\eta) = -1$, $w_3 = \text{clip}_{[-1,1]}(-1(1-\eta) + \eta) = 1$, $w_4 = \text{clip}_{[-1,1]}(1(1-\eta) -\eta) = -1$ and so on.

Thus the iterates oscillates between $-1$ and $1$. However, the best fixed comparator for the sequence of losses is given by $0$. Hence we have that

\begin{align}
    \sum_{t=1}^n f_t(w_t) - f_t(0)
    &= 1/2 + 2(n - 1) - n/2\\
    &\ge 3n/4,
\end{align}
for all $n \ge 2$.

Thus choosing step-size $\eta \ge 1$ can be exploited by the adversary to enforce a linear regret even for the case of static comparators.

So it suffices to consider OGD algorithms with step size $\eta < 1$ as what is done in the following lemma.

First, let's define the comparator class:
\begin{align}
    \TV^{(1)}(1)
    &:= \{\theta_{1:n} : \TV_1(\theta_{1:n}) \le 1 \}.
\end{align}

\begin{lemma} \label{lem:linsmooth}
There exist a choice of loss functions, comparator sequence $\theta_{1:n} \in  \TV^{(1)}(1)$ and decision set such that OGD with steps size $\eta < 1$ necessarily suffers a dynamic regret of $\Omega(n^{1/4})$ for all $n \ge 35$.
\end{lemma}
\begin{proof}
Consider a setup where the decision set $\cD = [-1/(2\sqrt{2.2}),1/(2\sqrt{2.2})]$. Let $y_t = \theta_t + \epsilon_t$ where $|\theta_t| \le 1/(8\sqrt{2.2})$, and $\epsilon_t$ are iid uniformly chosen from $[-1/(8\sqrt{2.2}),1/(8\sqrt{2.2})]$.  Thus $y_t \in [-1/(4\sqrt{2.2}),1/(4\sqrt{2.2})]$. Further $\theta_{1:n} \in \mathcal{TV}^{(1)}(1)$.

The loss at time $t$ is $f_t(x) = \frac{2\sqrt{2.2}}{3}(y_t - x)^2$. So the Lispchitzness coefficient of these losses is bounded by $|\grad f_t(x)| \le \frac{4\sqrt{2.2}}{3}(|y_t| + |x|) \le 1 := G$ as $|y_t| \le 1/(4\sqrt{2.2})$ and $|x| \le 1/(2\sqrt{2.2})$ for all $x \in \cD$.

Let D be the diameter of $\cD$. So $D = 1/\sqrt{2.2}$.

Under this setup, the projected online gradient descent  (OGD) with learning rates $\eta < 1$ doesn't need to do any projection. This can be seen as follows. Assume that OGD till step $t$ doesn't project. Let $\Pi$ denote the projection to set $\cD$. Then the iterate at time $t+1$ (denoted by $x_{t+1}$) is given by $x_{t+1} = \Pi(z_{t+1})$ where
\begin{align}
    z_{t+1}
    &= \sum_{k=1}^{t} (1-\eta)^{t-k} \eta y_k \label{eq:linfit}.
\end{align}

We have that
\begin{align}
    |z_{t+1}| \le  1/(4\sqrt{2.2}),
\end{align}
where we applied triangle inequality, summed up the infinite series and used the fact that $|y_t| \le  1/(4\sqrt{2.2})$. So $z_{t+1} \in \cD$ and therefore $x_{t+1} = z_{t+1}$. Hence by induction, we have that OGD with learning rate $\eta < 1$ doesn't need to project.

Looking at Eq.\eqref{eq:linfit} we see that the OGD output at any time is a fixed linear function of the revealed labels $y_t$. \citet{baby2020higherTV} calls such forecasters to be linear forecasters. They provide the following proposition (para-phrased here for clarity) about such forecasters:

\begin{proposition}(Proposition 2 in \citet{baby2020higherTV} for $k=1$) For any online estimator producing estimates $\hat \theta_t$ which is a fixed linear function of past labels $y_{1:t-1}$, $t \in [n]$ we have
\begin{align}
    \sup_{\theta_{1:n} \in \in \mathcal{TV}^{(1)}(1)}\sum_{t=1}^n E[(\hat \theta_t - \theta_t)^2] = \Omega(n^{1/4}). \label{eq:linlb}
\end{align}

\end{proposition}

Thus we have
\begin{align}
    \sup_{\theta_{1:n} \in \in \mathcal{TV}^{(1)}(1)} \sum_{t=1}^n E[(y_t - \hat \theta_t)^2] - E[(y_t - \theta_t)^2]
    &= \sup_{\theta_{1:n} \in \in \mathcal{TV}^{(1)}(1)} \sum_{t=1}^n E[(\hat \theta_t-\theta_t)^2] - 2 E[\epsilon_t (\hat \theta_t-\theta_t)] + E[\epsilon_t^2] - E[\epsilon_t^2]\\
    &=_{(a)} \sup_{\theta_{1:n} \in \in \mathcal{TV}^{(1)}(1)} \sum_{t=1}^n E[(\hat \theta_t-\theta_t)^2] - 2 E[\epsilon_t] E[(\hat \theta_t-\theta_t)]\\
    &= \sup_{\theta_{1:n} \in \in \mathcal{TV}^{(1)}(1)} \sum_{t=1}^n E[(\hat \theta_t-\theta_t)^2]\\
    &=_{(b)} \Omega (n^{1/4}),
\end{align}
where line (a) is due to the fact that $\hat \theta_t$ and $\epsilon_t$ are mutually independent (because of online nature of algorithm) and line (b) is due to Eq.\eqref{eq:linlb}.

Thus we conclude that
\begin{align}
   \sup_{\theta_{1:n} \in \in \mathcal{TV}^{(1)}(1)} \sum_{t=1}^n E[f_t(\hat \theta_t) - f_t(\theta_t)]
    &= \Omega(n^{1/4}),
\end{align}
where the losses $f_t$ are as defined in the beginning of the proof. This concludes the lemma.
\end{proof}

\section{Overview of proof strategy} \label{app:over}

\begin{remark}(\textbf{reason behind faster rates}).
We remark that in the low TV1 regime, the sequence assumes a piecewise linear structure with gradually changing slopes. This regularity of the comparator sequence is what enables to derive fast regret rates in low TV1 regimes.
\end{remark}

For the sake of clarity, we give an elaborate overview of our proof scheme before presenting the analysis in Appendix \ref{app:analysis}. We adopt the notations introduced in Section \ref{sec:main}.

We start with the KKT conditions of the offline optimization problem (Eq.\eqref{eq:obj}) defined in Section \ref{sec:1d}.
\begin{restatable}{lemma}{lemkkt}(\textbf{KKT conditions}) \label{lem:kkt}
Let $u_1,\ldots,u_n$ be the optimal primal variables and let $\lambda \ge 0$ be the optimal dual variable corresponding to the constraint \eqref{eq:constr-ec-1}. Further, let $\gamma_t^- \ge 0, \gamma_t^+ \ge 0$ be the optimal dual variables that correspond to constraints \eqref{eq:constr-ec-2} and \eqref{eq:constr-ec-3} respectively for all $t \in [n]$. By the KKT conditions, we have

\begin{itemize}
    \item \textbf{stationarity: } $\grad f_t({ u}_t) = \lambda \left ( (s_{t-1} - s_{t}) - (s_{t-2} - s_{t-1}) \right) +  \gamma^-_t -  \gamma^+_t$, where $s_t=\sign((u_{t+2} - u_{t+1})-(u_{t+1} - u_t))$. Here $\sign(x) = x/|x|$ if $|x| > 0$ and any value in $[-1,1]$ otherwise. For convenience of notations, we also define 
    $s_{-1} = s_0 = s_{n-1} = s_n = 0$.
    \item \textbf{complementary slackness: } (a) $\lamda \left(\|D^2 u_{1:n} \|_1 - C_n/n \right) = 0$; (b)  $ \gamma^-_t ( u_t + 1) = 0$ and $ \gamma^+_t ( u_t - 1) = 0$ for all $t \in [n]$
\end{itemize}

\end{restatable}

As mentioned in Section \ref{sec:main}, a key step is proving Lemma \ref{lem:main} which we restate for convenience.

\lemmain*
\begin{proof}[Proof Sketch]
For the sake of explaining ideas, we consider a case where the offline optimal within a bin $[i_s,i_t] \in \cP$ doesn't touch the boundary 1 but may touch boundary $-1$ at multiple time points. (In the full proof, we show that the partition $\cP$ can be slightly modified so that in non-trivial cases, the offline optimal can only touch one of the boundaries due to the TV1 constraint within the bins described in Lemma \ref{lem:keypart}.) Then by complementary slackness of Lemma \ref{lem:kkt} we conclude that $\gamma_j^+ = 0$ for all $j \in [i_s,i_t]$. Our analysis starts by considering a scenario where the offline optimal touches boundary $-1$ at precisely two points $r,w \in[i_s,i_t]$ with $r < w$ (see Fig.\ref{fig:lin}). Again via complementary slackness, only $\gamma_r^-$ and $\gamma_w^-$ can be potentially non-zero in this case. Through certain careful bounding steps, we show that:
\begin{align}
    T_2 + T_3
    &\le - B(\lamda,\gamma_r^-, \gamma_w^-;r,w), \label{eq:two-bound}
\end{align}
where $B$ is a function jointly convex in its arguments $\lamda,\gamma_r^-, \gamma_w^-$. We treat $r$ and $w$ to be fixed parameters.  The exact form of the function $B$ is present at Eq.\eqref{eq:bobj} in Appendix. 
Then we consider the following convex optimization procedure:
\begin{mini!}|s|[2]                   
    {\lamda, \gamma_r^-, \gamma_w^-}                               
    {B(\lamda,\gamma_r^-,\gamma_w^-;r,w)}   
    {}{}
    \label{eq:main-opt}             
    \addConstraint{\lamda}{\ge 0}
\end{mini!}

\begin{figure}[tb]
  \centering
  \stackunder{\hspace*{-0.5cm}\includegraphics[width=0.60\textwidth]{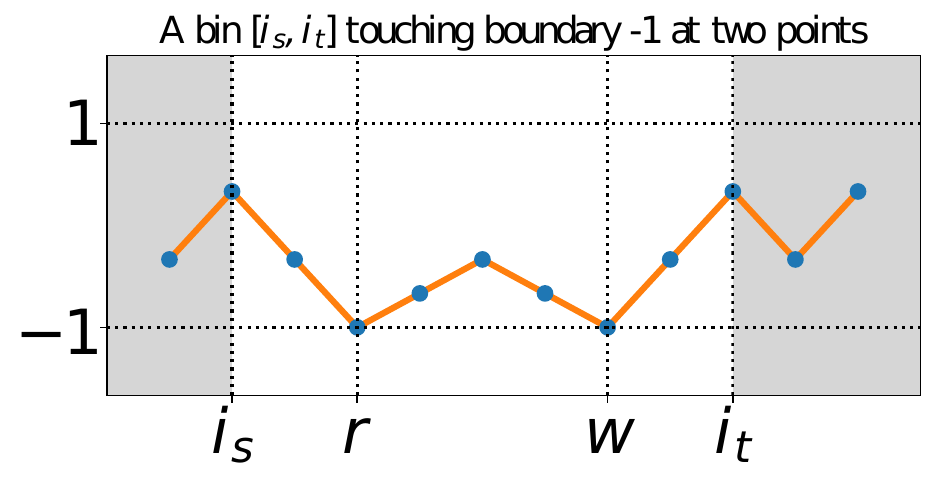}}{}
  \caption{\emph{A configuration referred in the proof sketch of Lemma \ref{lem:main}. The blue dots represent the offline optimal sequence.}} \label{fig:lin}  
\end{figure}

First, we perform a partial minimization wrt $\gamma_r^-$ and $\gamma_w^-$ keeping $\lamda$ fixed. Note that even-though $\gamma_r^- \ge 0$ and $\gamma_w^- \ge 0$ via Lemma \ref{lem:kkt}, we choose to perform an \emph{unconstrained} minimization wrt these variables as doing so can only increase the bound on $T_2+T_3$.

Let the optimal solutions of the partial minimization procedure be denoted by  $\hat \gamma_r^-$ and $\hat \gamma_w^-$. We find that:
\begin{align}
    B(\lamda,\hat \gamma_r^-,\hat \gamma_w^-;r,w) = \cL(\lamda), \label{eq:part-opt}
\end{align}
where $\cL(\lamda)$ is a linear function of $\lamda$ that \emph{doesn't depend} on $r$ or $w$ (Eq.\eqref{eq:linlamda} in Appendix). The constrained minimum of this linear function is then found to be attained at $\lamda = 0$ and we show that
\begin{align}
 -B(0,\hat \gamma_r^-,\hat \gamma_w^-;r,w) = O(1)   
\end{align}

This leaves us with an important question on how to handle more than two boundary touches at $-1$ where many of $\gamma_j^-$, $j \in [i_s,i_t]$ can potentially be non-zero. One could perform a similar unconstrained optimization as earlier wrt all $\gamma_j^-$. However, deriving the closed form expressions for the  optimal $\hat \gamma_j^-$ becomes very cumbersome as it involves solving for a complex system of linear equations. In the following, we argue that this general case can be handled via a reduction to the previous setting where only two dual variables $\gamma_r^-$ and $\gamma_w^-$ can be potentially non-zero. Specifically we show that the same auxiliary function $B$ as in Eq.\eqref{eq:two-bound} can be used to obtain
\begin{align}
    T_2 + T_3
    &\le - B(\lamda,\tilde \gamma_r^-, \tilde \gamma_w^-;\tilde r, \tilde w),
\end{align}
where $\tilde r, \tilde w, \tilde \gamma_r^-$ and $\gamma_w^-$ can be computed from the sequence of dual variables $\gamma_{i_s:i_t}^-$. Now we can proceed to optimize similarly as in Eq.\eqref{eq:main-opt} with the  optimization variables being $\lamda, \tilde \gamma_r^-, \tilde \gamma_w^-$ and use the same arguments as earlier to bound $T_2+T_3 = O(1)$. We remark that while doing so, it is an extremely fortunate fact that the partially minimized objective in Eq.\eqref{eq:part-opt} does not depend on the parameter values $r$ and $w$. This fact in hindsight is what permitted us to fully eliminate the dependence of all $\gamma_j^-$ where $j \in [i_s,i_t]$ on the bound via the method of reduction to the case of two non-zero dual variables considered earlier.
\end{proof}

\textbf{Proof summary for Theorem \ref{thm:main-d} in multi-dimensions}. \label{sec:hd}
In rest of this section, we focus on outlining the analysis ideas that facilitated the main result Theorem \ref{thm:main-d}. The high-level idea is to construct a reduction that helps us to re-use much of the machinery developed in Section \ref{sec:1d}. We emphasize that this reduction happens only in the analysis, and we \emph{do not} run $d$ uni-variate FLH-SIONS algorithms for handling multi-dimensions. Following Lemma serves a key role in materializing the desired reduction.

\begin{lemma} \label{lem:diagonal-main}
Let $\bs X_j \in \mathbb{R}^{d \times 2d}$ be as defined as:
\begin{gather}
    \bs X_j^T
    =
    \begin{bmatrix}
    \bs x_j[1:2] & \bs 0 & \dots & \bs 0\\
    \bs 0 & \bs x_j[3:4] & \dots & \bs 0\\
    \vdots & \ddots & & \vdots\\
    \bs 0 & \dots & & \bs x_j[2d-1:2d]
    \end{bmatrix}, \label{eq:block-cov-main}
\end{gather} 
where $\bs 0 = [0,0]^T$ and $\bs x_j \in \mathbb{R}^{2d}$. The entries $\bs x_j[2k-1:2k] \in \mathbb{R}^2$ for $k \in [d]$. Let $\tilde f_j(\bs v) = f_j(\bs X_j \bs v)$ for some $\bs v \in \mathbb{R}^{2d}$ and let $\bs \Sigma:= \bs X_j^T \bs X_j \in \mathbb{R}^{2d \times 2d}$ which is a block diagonal matrix. We have that 
\begin{align}
    \grad^2 \tilde f_j(\bs v) \preccurlyeq  \bs \Sigma.
\end{align}
\end{lemma}

In multi-dimensions also we form a partition $\cP$ of the offline optimal similar to Lemma \ref{lem:keypart}. Then we consider following regret decomposition for any bin $[i_s,i_t] \in \cP$.
\begin{align}
    &\sum_{j=i_s}^{i_t} f_j({\bs p}_j) - f_j(\bs u_j)
    = \underbrace{\sum_{j=i_s}^{i_t} f_j(\bs p_j) -  f_j(\bs X_j \bs \alpha_j)}_{T_1}
    + \underbrace{\sum_{j=i_s}^{i_t}  f_j(\bs X_j \bs \alpha_j) -  f_j(\bs X_j \bs \beta_j)}_{T_2}
    + \underbrace{\sum_{j=i_s}^{i_t}  f_j(\bs X_j \bs \beta_j) - f_j(\bs u_j)}_{T_3},\quad \label{eq:regdecomp-multi-main}
\end{align}
where we shall shortly describe how to construct the quantities $\bs X_j \in \mathbb{R}^{d \times 2d}, \bs \alpha_j \in \mathbb{R}^{2d}$ and $\bs \beta_j \in \mathbb{R}^{2d}$. For compactness of notations later, let's define  $\bs \alpha_{j,k} = \bs \alpha_j[2k-1:2k] \in \mathbb{R}^{2}$, $\bs \beta_{j,k} = \bs \beta_j[2k-1:2k] \in \mathbb{R}^{2}$ and $\bs y_{j,k} = \bs x_j[2k-1:2k] \in \mathbb{R}^{2}$ for some $\bs x_j \in \mathbb{R}^{2d}$ as in lemma \ref{lem:diagonal-main}. The Hessian dominance in Lemma \ref{lem:diagonal-main} leads to:
\begin{align}
    \tilde f_j(\bs \alpha_j) - \tilde f_j(\bs \beta_j)
    &\le \sum_{k=1}^d \langle \grad f_j(\bs X_j \bs \beta_j)[k] \bs y_{j,k}, \bs \alpha_{j,k} -  \bs \beta_{j,k} \rangle
    + \frac{1}{2} \sum_{k=1}^d \|\bs \alpha_{j,k} -  \bs \beta_{j,k}\|_{\bs y_{j,k} \bs y_{j,k}^T}^2\\
    &:= \sum_{k=1}^d t_{2,j,k}.
    \label{eq:t2diag-main}
\end{align}

Further, due to gradient Lipschitzness of $f_j$,
\begin{align}
    \tilde f_j(\bs \beta_j) - f_j(\bs u_j)
    &\le  \sum_{k=1}^d \grad f_j(\bs u_j)[k] \cdot \left( \bs \beta_{j,k}^T \bs y_{j,k} - \bs u_j[k] \right)
+ \sum_{k=1}^d \frac{1}{2} \| \bs \beta_{j,k}^T \bs y_{j,k} - \bs u_j[k] \|_2^2\\
    &:= \sum_{k=1}^d t_{3,j,k}
    \label{eq:t3diag-main}
\end{align}
Combining Eq.\eqref{eq:t2diag-main} and \eqref{eq:t3diag-main}, we see that  $T_2 + T_3$ in any bin $[i_s,i_t]$ can be bounded coordinate-wise:
\begin{align}
    T_2 + T_3
    &\le \sum_{k=1}^d \sum_{j=i_s}^{i_t} t_{2,j,k} + t_{3,j,k}. \nonumber
\end{align}
This form allows one to bound $\sum_{j=i_s}^{i_t} t_{2,j,k} + t_{3,j,k} = O(1)$ separately for each coordinate by constructing $\bs \alpha_{j,k}, \bs \beta_{j,k}$ and $\bs y_{j,k}$ similar to Section \ref{sec:1d}. We then sum across all coordinates to bound $T_2 + T_3 = O(d)$. We remark that the situation is a bit more subtle here because in-order to handle certain combinatorial structures imposed by the KKT conditions, we had to use a sequence of comparators $\bs \alpha_{i_s},\ldots,\bs \alpha_{i_t}$ (for linear predictors in Eq.\eqref{eq:regdecomp-multi-main}) that switches at-most $O(d)$ times . Finally by appealing to strong adaptivity of FLH-SIONS, we show that $T_1 = \tilde O(d^2)$ for each bin $[i_s,i_t] \in \cP$ and Theorem \ref{thm:main-d} then follows by adding the $\tilde O(d^2)$ regret across all $O(n^{1/5}C_n^{2/5} \vee 1)$ bins in $\cP$.

\section{Analysis} \label{app:analysis}
We start with the analysis in the uni-variate setting followed by the proof in multi-dimensions. The analysis requires very clumsy algebraic manipulations in certain places. We used Python's open-source simplification engine SymPy \citep{sympy} to assist with the algebraic manipulations. 

\paragraph{A remark.} The constants occurring in the proofs may be optimized further though we haven't aggressively focused on doing so. Throughout the analysis we compete with comparators whose TV1 distance is bounded by $C_n$. This quantity can be unknown to the algorithm. Hence the resulting regret rate of FLH-SIONS simultaneously holds for any value of $C_n$.

\subsection{One dimensional setting} \label{app:1d}

\lemkkt*
\begin{proof}

By introducing auxiliary variables, we can re-write the offline optimization problem as:
\begin{mini!}|s|[2]                   
    {\tilde u_1,\ldots,\tilde u_n}                               
    {\sum_{t=1}^n f_t(\tilde { u}_t)}   
    {}             
    {}                                
    \addConstraint{\tilde z_t}{ = \tilde u_{t+2} - 2\tilde u_{t+1} + \tilde u_t \: \forall t \in [n-2]}
    \addConstraint{\sum_{t=1}^{n-2}|\tilde z_t|}{\le C_n/n,}  
    \addConstraint{-1}{\le \tilde u_t \: \forall t \in [n],}
    \addConstraint{\tilde u_t}{\le 1 \: \forall t \in [n],}
\end{mini!}

The Lagrangian of the optimization problem can be written as
\begin{align}
    \cL(\tilde u_{1:n}, \tilde z_{1:n-2}, \tilde v_{n-2}, \tilde \gamma_{1:n}^-, \tilde \gamma_{1:n}^+ \tilde \lamda)
    &= \sum_{t=1}^n f_t(\tilde u_t) + \tilde \lamda \left( \sum_{t=1}^{n-2} |\tilde z_t| - C_n/n \right)\\
    &\quad + \sum_{t=2}^{n-2} \tilde v_t (\tilde u_{t+2} - 2 \tilde u_{t+1} + \tilde u_t - \tilde z_t) + \sum_{t=1}^n \gamma_t^{+} (\tilde u_t - 1) - \gamma_t^{-} (\tilde u_t + 1).
\end{align}

Due to stationary conditions wrt $u_t$, we have
\begin{align}
    \grad f_t(u_t) = 2v_{t-1} - v_t - v_{t-2} + \gamma_t^- - \gamma_t^{+},
\end{align}
where we define $v_{-1} = v_{0} = v_{n-1} = v_n = 0$ and, due to staionarity conditions wrt $v_t$ we have
\begin{align}
    v_t = \lamda \sign(z_t).
\end{align}

Combining the above two equations and the complementary slackness rule now yields the Lemma.

\end{proof}

\paragraph{Terminology.} In what follows, we refer to $u_{1:n}$ from the Lemma above to be the offline optimal sequence.

\lemkeypart*
\begin{proof}

Let the total number of bins formed be $M$. Consider the case where $M > 1$.
We have that
\begin{align}
   \|D^2 u_{1:n} \|_1
   &\ge \sum_{i=1}^{M-1} \|D^2 u_{i_s \rightarrow i_t+1} \|_1\\
   &\ge_{(a)} 1/\ell_{i_s \rightarrow i_t+1}^{3/2}\\
   &\ge_{(b)} \frac{(M-1)^{5/2}}{n^{3/2}},
\end{align}
where line (a) follows due to the construction of the partition and line (b) is due to Jensen's inequality applied to the convex fucntion $f(x) = 1/x^{3/2}$ for $x > 0$.

Rearranging and including the trivial case where $M=1$ yields the lemma.

\end{proof}

\begin{proposition} \label{prop:cts}
In the following analysis we will often use a useful represent offline optimal within a bin $[a,b]$  to be $m_a,m_a+m_{a+1},\ldots,\sum_{t=a}^b m_t$ WLOG. We can view this sequence to be samples obtained from a piece-wise linear signal that is continuous at every sampling point. 
\end{proposition}

\begin{lemma}\label{lem:residual} (\textbf{residual bound})
Consider a bin $[a,b]$. Let $\ell := b-a+1$. Define:
\begin{gather}
    X
    =
    \begin{bmatrix}
    1 & 1\\
    1 & 2\\
    \vdots \\
    1 & \ell
    \end{bmatrix} 
\end{gather}

Let $\bs \beta = (\bs X^T \bs X)^{-1}\bs X^T u_{a:b}$ be the least square fit coefficient computed with labels $u_t$ and co-variates $\bs x_t = [1,t-a+1]^T$ where $t \in [a,b]$. Then we have that the residuals satisfy
\begin{align}
    |\bs \beta^T \bs x_t - u_t| \le 20\ell\|D^2 u_{a:a+\ell-1}\|_1,
\end{align}
whenever $\ell \ge 6$.


\end{lemma}
\begin{proof}
We follow the notations of Proposition \ref{prop:cts} for representing the offline optimal $u_a,\ldots,u_b$. The residual at time $i \in [a,b]$ can be computed through straight forward algebra as:
\begin{align}
    u_i - \bs \beta^T \bs x_i
    &= \frac{1}{(\ell^2-1)} \sum_{j=2}^\ell \Bigg(\Bigg. 6(1+(1-2i)/\ell)(\ell-j+1/)(\ell+j)/2\\
    &\qquad +(6i+6i/\ell-4\ell-6-2/\ell)(\ell-j+1) + (\ell^2-1)\mathbb{I} \{ j \le i\} \Bigg. \Bigg) m_{a+j-1}, \label{eq:resid1}
\end{align}
where $\mathbb{I} \{ \cdot \}$ is the indicator function assuming value 1 if the argument evaluates true and 0 otherwise. Now we note that if all $m_k$ for $k \in [a+1,b]$ are same, then the residuals $u_i - \bs \beta^T \bs x_i$ must be zero for all $i$ as the least square fit exactly matches the labels in this case. In particular, this implies from Eq.\eqref{eq:resid1} that
\begin{align}
\frac{1}{(\ell^2-1)} \sum_{j=2}^\ell \Bigg(\Bigg. &6(1+(1-2i)/\ell)(\ell-j+1/)(\ell+j)/2\\
    &+(6i+6i/\ell-4\ell-6-2/\ell)(\ell-j+1) + (\ell^2-1)\mathbb{I} \{ j \le i\} \Bigg. \Bigg) m_{a+1}
    = 0. \label{eq:resid2}
\end{align}

Subtracting Eq.\eqref{eq:resid2} from \eqref{eq:resid1} we get,
\begin{align}
    u_i - \bs \beta^T \bs x_i
    &= \frac{1}{(\ell^2-1)} \sum_{j=2}^\ell \Bigg(\Bigg. 6(1+(1-2i)/\ell)(\ell-j+1/)(\ell+j)/2\\
    &\qquad +(6i+6i/\ell-4\ell-6-2/\ell)(\ell-j+1) + (\ell^2-1)\mathbb{I} \{ j \le i\} \Bigg. \Bigg) (m_{j+a-1}-m_{a+1})\\
    &\le \frac{1}{(\ell^2-1)} \max_{j \in [a+2,b]}|m_j-m_{a+1}| \sum_{j=3}^\ell \Bigg|\Bigg. 6(1+(1-2i)/\ell)(\ell-j+1/)(\ell+j)/2\\
    &\qquad +(6i+6i/\ell-4\ell-6-2/\ell)(\ell-j+1) + (\ell^2-1)\mathbb{I} \{ j \le i\} \Bigg. \Bigg|, 
\end{align}
where the last line is due to Holder's inequality. Further, we have $|m_j-m_{a+1}| \le \sum_{t=j}^{a+2}|m_j - m_{j-1} | \le \|D^2 u_{a:b} \|_1$ by the definition of the discrete difference operator $D^2$.

Now applying triangle inequality and the crude bounds $1+(1-2i)/\ell \le 3, (\ell-j+1) \le \ell, (\ell+j) \le 2\ell, i/\ell \le 1, 2\ell \ge 2/\ell$ and $-2/\ell \le 0$ we obtain

\begin{align}
    \Bigg|\Bigg. &6(1+(1-2i)/\ell)(\ell-j+1/)(\ell+j)/2\\
    &\qquad +(6i+6i/\ell-4\ell-6-2/\ell)(\ell-j+1) + (\ell^2-1)\mathbb{I} \{ j \le i\} \Bigg. \Bigg|
    &\le 19 \ell^2 + 2\ell.
\end{align}

So,

\begin{align}
    | u_i - \bs \beta^T \bs x_i|
    &\le \ell \cdot \frac{19 \ell^2 + 2\ell}{\ell^2-1}  \|D^2 u_{a:b} \|_1\\
    &\le 20 \ell \|D^2 u_{a:b} \|_1,
\end{align}
where the last line is due to $19 \ell^2 + 2\ell \le 20 \ell^2 - 20$ for all $\ell \ge 6$.

\end{proof}

\begin{lemma}(\textbf{bounding $T_3$}) \label{lem:t3}
Consider a bin $[a,b]$ with length $\ell = b-a+1$ obtained from the scheme in Lemma \ref{lem:keypart}. Assume the notations in Lemma \ref{lem:residual}. Let's represent the residual as $r_t: = \bs \beta^T \bs x_t - u_t = (t-a+1)M_{t-1}+C_{t-1}$ for $t > a$  and $r_1:= \bs \beta^T \bs x_a - u_a = M_a + C_a$ with $M_{b}:= M_{b-1} = M_{a+\ell-2}$ and $C_{b}:=C_{b-1} = C_{a+\ell-2}$. Suppose $\|D^2 u_{a:b}\|_1 \le \ell^{-3/2}$. We have,

\begin{align}
    \sum_{t=a}^{b} f_t(\bs \beta^T \bs x_t ) - f_t(u_t)
    &\le 200 + \lamda \Bigg ( \Bigg. (s_{a-1} - s_{a-2})(M_a+C_a) - (s_b - s_{b-1})(\ell M_b + C_b) \\
   & - s_{a-1} M_a + s_{b-1}M_{b-1} - \sum_{t=a+1}^{b} |M_t - M_{t-1}|  \Bigg. \Bigg )\\
   &+20\ell^{-1/2} \sum_{t=a}^{b} |\gamma_t^- - \gamma_t^+| \label{eq:t3lemma}
\end{align}
Further we have $|M_a| \le  \|D^2 u_{a:b}\|_1 $ and $|M_b| \le  \|D^2 u_{a:b}\|_1 $ whenever $\ell \ge 2$. 

Here the semantics is that each $M_t = r_{t+1} - r_t$ for all $t > a$ and $M_a = r_{a+1} - r_a$. Any two points $r_t$ and $r_{t+1}$ can be joined using a unique line segment which in turn defines $C_t$ appropriately.

\end{lemma}
\begin{proof}

By gradient Lipschitzness of $f$ we have

\begin{align}
    \sum_{t=a}^{b} f_t(\bs \beta^T \bs x_t ) - f_t(u_t)
    &\le \sum_{t=a}^{b} \langle \grad f_t(u_t), \bs \beta^T \bs x_t - u_t \rangle + \sum_{t=a}^{b} \frac{1}{2} (\bs \beta^T \bs x_t - u_t)^2.
\end{align}

Now will focus on bounding the last two terms above.

From the construction of bins in Lemma \ref{lem:keypart}, we know that $\ell\|D^2 u_{a:b}\|_1 \le 1/\sqrt{\ell}$. Hence we obtain using Lemma \ref{lem:residual} that
\begin{align}
    \sum_{t=a}^{b} \frac{1}{2} (\bs \beta^T \bs x_t - u_t)^2 \le 200.
\end{align}

Recall the representation of the residual $\bs \beta^T \bs x_t - u_t = tM_t+C_t$ mentioned in the lemma statement. Observe that in accordance with Proposition \ref{prop:cts} this residual can also be viewed as samples of a piece-wise linear signal that is continuous at every sampled point. In particular observe that for every $t \in [a,b]$ we have:
\begin{align}
    (t-a+1)M_{t-1} + C_{t-1} = (t-a+1)M_{t} + C_{t}
\end{align}

Consequently

\begin{align}
    C_{t} - C_{t-1} 
    &= (t-a+1)(M_{t-1} - M_{t}) \label{eq:incpt}
\end{align}

From KKT conditions of Lemma \ref{lem:kkt} we have
\begin{align}
 \sum_{t=a}^{b} \langle \grad f_t(u_t), \bs \beta^T \bs x_t - u_t \rangle &= \underbrace{\sum_{t=a}^{b} \lamda \left( \left((s_{t-1} - s_{t-2}) - (s_{t} - s_{t-1}) \right) ((t-a+1)M_t + C_t)\right)}_{X_1}\\
 &+ \underbrace{ \sum_{t=a}^{b} (\gamma_t^- - \gamma_t^+) (\bs \beta^T \bs x_t - u_t)}_{X_2}
\end{align}

\begin{align}
    \frac{X_1}{\lambda}
    &= (s_{a-1} - s_{a-2})(M_a+C_a) - (s_b - s_{b-1})(\ell M_b + C_b)\\
    &+ \sum_{t=a}^{b-1} (s_t - s_{t-1}) \left( (t-a+2)M_{t+1} + C_{t+1} - ((t-a+1)M_t + C_t) \right)\\
    &=_{(a)} (s_{a-1} - s_{a-2})(M_a+C_a) - (s_b - s_{b-1})(\ell M_b + C_b) + \sum_{t=a}^{b-1} (s_t - s_{t-1}) M_{t+1}\\
    &= (s_{a-1} - s_{a-2})(M_a+C_a) - (s_b - s_{b-1})(\ell M_b + C_b) + \sum_{t=a}^{b-1} (M_{t+1} - M_{t+2}) s_{t} - s_{a-1} M_2 + s_{b-1}M_{\ell}\\
    &=_{(b)} (s_{a-1} - s_{a-2})(M_a+C_a) - (s_b - s_{b-1})(\ell M_b + C_b)  - s_{a-1} M_a + s_{b-1}M_{b-1} - \sum_{t=a+1}^{b} |M_t - M_{t-1}|,
\end{align}
where in line (a) we used Eq.\eqref{eq:incpt} and in line (b) we used the fact that $s_t  = \sign{((u_{t+2} - u_{t+1}) - (u_{t+1} - u_t))} = \sign{(M_{t+2} - M_{t+1})}$ along with the fact that $M_a = M_{a+1}$ and $M_{b-1} = M_{b}$.

By Holder's inequality and Lemma \ref{lem:residual}, we have
\begin{align}
    X_2
    &\le 20\ell\|D^2 u_{a:b}\|_1 \sum_{t=a}^{b} |\gamma_t^- - \gamma_t^+|\\
    &\le 20\ell^{-1/2} \sum_{t=a}^{b} |\gamma_t^- - \gamma_t^+|,
\end{align}
where the last line is due to $\|D^2 u_{a:b}\|_1 \le \ell^{-3/2}$ as assumed in the lemma's statement. Putting everything together completes the proof.

Next, we proceed to give useful bounds on $|M_a|$ and $|M_{b-1}|$.

Since $M_a = M_{a+1}$ and $C_a = C_{a+1}$, we have $M_a = (u_{a+1} - \bs \beta^T \bs x_{a+1}) - (u_a - \bs \beta^T \bs x_a)$.So Eq.\eqref{eq:resid1} we have,
\begin{align}
    |M_a|
    &= \left|\sum_{j=2}^\ell \frac{6(\ell-j+1)(1-j)}{\ell^3-\ell} (m_{j+a-1}-m_{a+1}) \right|\\
    &\le \|D^2 u_{a:b}\|_1 \sum_{j=3}^\ell \frac{6(\ell-j+1)(j-1)}{\ell^3-\ell}\\
    &= \|D^2 u_{a:b}\|_1 \frac{\ell^2+\ell-6}{\ell(\ell+1)}\\
    &\le  \|D^2 u_{a:b}\|_1 , \label{eq:1}
\end{align}
where in the last line we used $\frac{\ell^2+\ell-6}{\ell(\ell+1)} \le 1$ for all $\ell \ge 2$.

Similarly $M_{b-1} = u_{b} - \bs \beta^T \bs x_{b} - ( u_{b-1} - \bs \beta^T \bs x_{b-1})$ by recalling that $M_b = M_{b-1}$ and $C_b = C_{b-1}$. Proceeding from Eq.\eqref{eq:resid1} we obtain,

\begin{align}
    |M_{b-1}|
    &= \left| \sum_{j=2}^{\ell-1} \frac{6(\ell-j+1)(1-j)}{\ell^3-\ell} (m_{j+a-1}-m_b) \right|\\
    &\le \|D^2 u_{a:b}\|_1 \sum_{j=2}^{\ell-1} \frac{6(\ell-j+1)(j-1)}{\ell^3-\ell}\\
    &= \|D^2 u_{a:b}\|_1 \frac{\ell^2+\ell-6}{\ell(\ell+1)}\\
    &\le  \|D^2 u_{a:b}\|_1. \label{eq:2}
\end{align}

\end{proof}

\begin{lemma} \label{lem:dual-bounded}
Consider a bin $[a,b] \in \cP$ of length $\ell$ from Lemma \ref{lem:keypart}. Suppose $|u_a| < 1$. Then either $\gamma_j^- = 0$ or $\gamma_j^+=0$ for all $j \in [a,b]$.
\end{lemma}
\begin{proof}
We will provide arguments for the case when the offline optimal first hits $-1$ before hitting $1$ for some point in $[a,b]$. The arguments for the alternate case where it hits $1$ first are similar.

If the offline optimal hits $-1$ at some point in $[a,b]$ it can only rise upto at-most $-1 + 1/\sqrt{l}$ afterwards. This is due to the constraint $\|D^2 u_{a:b}\|_1 \le 1/\ell^{3/2}$.

Since $-1 + 1/\sqrt{l} < 1$ as $\ell > 1/4$, we have that the offline optimal never touches $1$ within the bin $[a,b]$. Consequently $\gamma_j^+ = 0$ for all $j \in [a,b]$.

\end{proof}

\begin{definition}
The \textbf{slope} of the optimal solution at a time point $t$ is defined to be $u_{t+1}-u_t$ for all $t \in [n-1]$.
\end{definition}

\begin{proposition} \label{prop:resid-mono}
The bins in $\cP$ can be further refined in such a way that each bin either satisfy the condition in Lemma \ref{lem:dual-bounded} or has constant slope, meaning the L1 TV distance is zero. Further in doing so the size of partition $\cP$ only gets increased by at-most 2.
\end{proposition}
\begin{proof}
Suppose for a bin $[a,b] \in \cP$, if the offline optimal starts at $1$. Then we can split that bin into two bins $[a,c]$ and $[c+1,b]$ such that $u_c > -1$ and $\|D^2 u_{a:c} \|_1 = 0$. Similar splitting can also be done for bins that start from $-1$. Observe that this refinement only increases the partition size only by at-most 2.
\end{proof}

\begin{corollary} \label{cor:dualprops}
One powerful consequence of Lemma \ref{lem:dual-bounded} and Proposition \ref{prop:resid-mono} when combined with the fact that $\gamma_t^-$ and $\gamma_t^+$ are both non-negative (Lemma \ref{lem:kkt}) is that $\sum_{t=a}^b |\gamma_t^- - \gamma_t^+|$ is either equal to $\sum_{t=a}^b \gamma_t^-$ or $\sum_{t=a}^b \gamma_t^+$ for all bins $[a,b]$ in the refined partition of Proposition \ref{prop:resid-mono} whenever the $\|D^2 u_{a:b} \|_1 > 0$.
\end{corollary}

From here on WLOG we will assume that the bins $[a,b]$ in partition $\cP$ will satisfy the conditions:

\begin{itemize}
    \item $\|D^2 u_{a:b} \|_1 \le 1/l^{3/2}$, where $\ell = b-a+1$.
    
    \item It satisfies the conditions mentioned in Proposition \ref{prop:resid-mono} and consequently satisfying the condition in Corollary \ref{cor:dualprops}.
    
    \item $|\cP| = O(n^{1/5}C_n^{2/5})$.
    
\end{itemize}

\begin{lemma}(\textbf{bounding $T_2$}) \label{lem:t2}
Consider a bin $[a,b] \in \cP$ with length $\ell = b-a+1$ that doesn't touch boundary 1. Let $\Gamma = \sum_{j=a}^b \gamma_j^-$ and $\tilde \Gamma = \sum_{j=a}^b j'\gamma_j^-$ where $j' := j-a+1$ . Let $\bs \beta$ be as in Lemma \ref{lem:residual}.

Let $F(\bs \beta) := \sum_{j=a}^b f_j(\bs x_j^T \bs \beta)$. Define:
\begin{align}
    \bs A := \sum_{j=a}^b \bs x_j \bs x_j^T
\end{align}

Consider the following update:

\begin{align}
    \bs \alpha &= \bs \beta -  \bs A^{-1} \sum_{j=a}^b f_j'(\bs x_j^T \bs \beta) \bs x_j\\
    &= \bs \beta -  \bs A^{-1} \grad F(\bs \beta)
\end{align}

We have,

\begin{align}
    2L \left(F(\bs \alpha) - F(\bs \beta) \right)
    &\le -\| \bs g\|_{\bs A^{-1}}^2 - \| \bs h \|_{\bs A^{-1}}^2 - 2 < \bs g, \bs A^{-1} \bs h \rangle \\
    &+ 2 \langle \bs A^{-1} (\bs g + \bs h), \sum_{j=a}^b \bs x_j \left(  f_j'(\bs x_j^T \bs \beta) - f_j'(u_j) \right),
\end{align}

where $\bs g = \lamda [-s_{a-2} + s_{a-1} + s_{b-1} - s_b, -s_{a-2} + (\ell+1)s_{b-1} - \ell s_b]^T$ and $\bs h = [\Gamma, \tilde \Gamma]^T$ so that $\sum_{j=a}^b f_j'(u_j) \bs x_j = \bs g + \bs h$.

Further we have:
\begin{itemize}
    \item $\| \bs g\|_{\bs A^{-1}}^2$ as in Eq. \eqref{eq:norm}
    \item $\| \bs h\|_{\bs A^{-1}}^2$ as in Eq. \eqref{eq:a4}
    \item $ \langle \bs A^{-1} \bs g,  \bs h \rangle$ as in Eq. \eqref{eq:a5}
    \item $\langle \bs A^{-1} \bs g, \sum_{j=a}^b \bs x_j (f_j'(\bs x_j^T \bs \beta) - f_j'(u_j)) \rangle$ bounded above by Eq.\eqref{eq:a3}
    \item $\langle \bs A^{-1} \bs h, \sum_{j=a}^b \bs x_j (f_j'(\bs x_j^T \bs \beta) - f_j'(u_j)) \rangle$ bounded above by Eq.\eqref{eq:a6}
    
\end{itemize}

Similar expressions can be derived for bins $[a,b]$ that may touch boundary 1 instead of -1.

\end{lemma}
\begin{proof}
We note that due to gradient Lipschitnzess of $f$,
\begin{align}
    \grad^2 F(\bs \beta) = \sum_{j=a}^b f_j''(\bs x_j^T \bs \beta) \bs x_j \bs x_j^T \preccurlyeq \bs A
\end{align}

So by Taylor's theorem we have for some $\bs z = t \bs \alpha + (1-t) \bs \beta$
\begin{align}
    F(\bs \alpha) - F(\bs \beta)
    &= - \langle \grad F(\bs \beta), \bs A^{-1} \grad F(\bs \beta) \rangle + \frac{1}{2} \|  \bs A^{-1} \grad F(\bs \beta)\|_{\grad^2 F(\bs z)}^2\\
    &\le - \langle \grad F(\bs \beta), \bs A^{-1} \grad F(\bs \beta) \rangle +  \frac{1}{2} \|  \bs A^{-1} \grad F(\bs  \beta)\|_{\bs A}^2\\
    &= -\frac{1}{2} \|\grad F(\bs \beta)\|_{\bs A^{-1}}^2, \label{eq:nd1}
\ \end{align}
where 
\begin{gather}
    \bs A^{-1}
    =\frac{2}{(\ell-1) \ell}
    \begin{bmatrix}
    2\ell + 1 & -3\\
    -3 & \frac{6}{\ell+1}
    \end{bmatrix} 
\end{gather}

Next we turn to lower bounding the above RHS

\begin{align}
    \|\grad F(\bs \beta)\|_{\bs A^{-1}}^2
    &= \|\sum_{j=a}^b f_j'(u_j) \bs x_j + \sum_{j=a}^b (f_j'(\bs x_j^T \bs \beta) - f_j'(u_j)) \bs x_j \|_{\bs A^{-1}}^2\\
    &\ge \|\sum_{j=a}^b f_j'(u_j) \bs x_j\|_{\bs A^{-1}}^2 - 2 \langle \bs A^{-1} \sum_{j=a}^b f_j'(u_j) \bs x_j, \sum_{j=a}^b (f_j'(\bs x_j^T \bs \beta) - f_j'(u_j)) \bs x_j \rangle
\end{align}

From the KKT conditions in Lemma \ref{lem:kkt}, we have
\begin{align}
    \sum_{j=a}^b f_j'(u_j) \bs x_j
    &= [\lamda (-s_{a-2} + s_{a-1} + s_{b-1} - s_b) + \Gamma, \lamda (-s_{a-2} + (\ell+1)s_{b-1} - \ell s_b) + \tilde \Gamma]^T, \label{eq:st-kkt}
\end{align}
where $\Gamma$ and $\tilde \Gamma$ are as defined in the statement of the lemma.

For the sake of brevity let's denote $\bs g = \lamda [-s_{a-2} + s_{a-1} + s_{b-1} - s_b, -s_{a-2} + (\ell+1)s_{b-1} - \ell s_b]^T$ and $\bs h = [\Gamma, \tilde \Gamma]^T$ so that $\sum_{j=a}^b f_j'(u_j) \bs x_j = \bs g + \bs h$.

We have
\begin{align}
    \bs A^{-1} \bs g &= \frac{2\lamda}{(\ell-1)\ell} [(2-2\ell)s_{a-2} + (2\ell+1)s_{a-1} - (\ell+2)s_{b-1} + (\ell-1)s_b,\\
    &\quad \frac{3(\ell-1)}{\ell+1}s_{a-2} - 3s_{a-1} + 3 s_{b-1} - \frac{3(\ell-1)}{\ell+1}s_b]^T, \label{eq:a1}
\end{align}

and so

\begin{align}
    \| \bs g\|_{\bs A^{-1}}^2
    &= \frac{2\lamda^2}{(\ell-1)\ell(\ell+1)} \Bigg( \Bigg. (2\ell^2-3\ell+1)s_{a-2}^2 + (4-4\ell^2) s_{a-2}s_{a-1} - (2\ell^2-6\ell+4) s_{a-2}s_b +\\
    &\qquad (2\ell^2-2) s_{a-2} s_{b-1} + (2\ell^2+3\ell+1) s_{a-1}^2 + \\
    &\qquad (2\ell^2-2) s_{a-1} s_b - (2\ell^2+6\ell+4) s_{a-1} s_{b-1} + \\
    &\qquad (2\ell^2-3\ell+1) s_b^2 + (4-4\ell^2) s_{b-1} s_b + (2\ell^2+3\ell+1) s_{b-1}^2 \Bigg. \Bigg) \label{eq:norm}
\end{align}

Using Eq. \eqref{eq:a1} we get

\begin{align}
    \langle \bs A^{-1} \bs g, \sum_{j=a}^b \bs x_j (f_j'(\bs x_j^T \bs \beta) - f_j'(u_j)) \rangle
    &= \frac{2\lamda}{(\ell-1)\ell} \Bigg ( \Bigg. (2-2\ell)s_{a-2} + (2\ell+1)s_{a-1}\\
    &- (\ell+2)s_{b-1} + (\ell-1)s_b \Bigg. \Bigg) \sum_{j=a}^b (f_j'(\bs x_j^T \bs \beta) - f_j'(u_j))\\
    &+ \frac{2\lamda}{(\ell-1)\ell} \Bigg ( \Bigg. \frac{3(\ell-1)}{\ell+1}s_{a-2} - 3s_{a-1} \\
    &+ 3 s_{b-1} - \frac{3(\ell-1)}{\ell+1}s_b  \Bigg. \Bigg)  \sum_{j=a}^b j' (f_j'(\bs x_j^T \bs \beta) - f_j'(u_j)). \label{eq:a2}
\end{align}

Using gradient Lipschitzness, triangle inequality and Lemma \ref{lem:residual} we have
\begin{align}
    \sum_{j=a}^b  (f_j'(\bs x_j^T \bs \beta) - f_j'(u_j))
    &\le \sum_{j=a}^b  |\bs x_j^T \bs \beta - u_j|\\
    &\le 20\ell^2 \|D^2 u_{a:b} \|_1, \label{eq:a9}
\end{align}

and similarly

\begin{align}
    \sum_{j=a}^b  j'(f_j'(\bs x_j^T \bs \beta) - f_j'(u_j))
    &\le \sum_{j=a}^b L j'|\bs x_j^T \bs \beta - u_j|\\
    &\le 20\ell^3 \|D^2 u_{a:b} \|_1.\label{eq:a10}
\end{align}

So continuing from Eq. \eqref{eq:a2},

\begin{align}
    \langle \bs A^{-1} \bs g, \sum_{j=a}^b \bs x_j (f_j'(\bs x_j^T \bs \beta) - f_j'(u_j)) \rangle
    &\le \frac{40\lamda \ell \|D^2 u_{a:b} \|_1}{(\ell-1)} \Bigg | \Bigg. (2-2\ell)s_{a-2} + (2\ell+1)s_{a-1}
    - (\ell+2)s_{b-1} + (\ell-1)s_b \Bigg. \Bigg | \\
    &+ \frac{40\lamda\ell^2 \|D^2 u_{a:b} \|_1}{(\ell-1)} \Bigg | \Bigg. \frac{3(\ell-1)}{\ell+1}s_{a-2} - 3s_{a-1}
    + 3 s_{b-1} - \frac{3(\ell-1)}{\ell+1}s_b  \Bigg. \Bigg|\\
    &\le \frac{40\lamda \ell^{-1/2} }{(\ell-1)} \Bigg | \Bigg. (2-2\ell)s_{a-2} + (2\ell+1)s_{a-1}
    - (\ell+2)s_{b-1} + (\ell-1)s_b \Bigg. \Bigg | \\
    &+ \frac{40\lamda\ell^{1/2}}{(\ell-1)} \Bigg | \Bigg. \frac{3(\ell-1)}{\ell+1}s_{a-2} - 3s_{a-1}
    + 3 s_{b-1} - \frac{3(\ell-1)}{\ell+1}s_b  \Bigg. \Bigg|, \label{eq:a3}
\end{align}
where we used $\|D^2 u_{a:b} \|_1 \le \ell^{-3/2}$.

We have

\begin{align}
    \langle \bs A^{-1} \bs h, \sum_{j=a}^b \bs x_j \left( f_j'(\bs \beta^T \bs x_j) - f_j'(u_j) \right)
    &= \frac{2}{(\ell-1) \ell} \Bigg( \Bigg. ((2\ell+1) \Gamma - 3 \tilde \Gamma) \sum_{j=a}^b  f_j'(\bs \beta^T \bs x_j) - f_j'(u_j)\\
    &+ \left( \frac{6 \tilde \Gamma}{\ell+1} - 3\Gamma \right) \sum_{j=a}^b  (j-a+1) \left( f_j'(\bs \beta^T \bs x_j) - f_j'(u_j) \right) \Bigg. \Bigg)\\
    &\le \frac{40\ell \|D^2 u_{a:b}\|_1}{(\ell-1)} |(2\ell+1) \Gamma - 3 \tilde \Gamma| + \frac{40 \ell^2 \|D^2 u_{a:b}\|_1}{(\ell-1)} \left| \frac{6 \tilde \Gamma}{\ell+1} - 3\Gamma \right|\\
    &\le \frac{40\ell^{-1/2}}{(\ell-1)} |(2\ell+1) \Gamma - 3 \tilde \Gamma| + \frac{40 \ell^{1/2}}{(\ell-1)} \left| \frac{6 \tilde \Gamma}{\ell+1} - 3\Gamma \right|, \label{eq:a6}
\end{align}
where the last line is obtained by using similar arguments used for obtaining Eq.\eqref{eq:a3}.

By substituting the expression for $\bs A^{-1}$ and simplifying,

\begin{align}
    \|\bs h \|_{\bs A^{-1}}^2
    &= \frac{2}{(\ell-1) \ell (\ell+1)} \left( (2\ell+1)(\ell+1) \Gamma^2 - 6\Gamma \tilde \Gamma (\ell+1) + 6 \tilde \Gamma^2 \right).\label{eq:a4}
\end{align}

Using Eq.\eqref{eq:a1}, we obtain

\begin{align}
    \langle \bs  A^{-1} \bs g, \bs h \rangle
    &= \sum_{j=a}^b \frac{2 \lamda \gamma_j^-}{(\ell-1) \ell} \Bigg ( \Bigg. \frac{-3j+3\ell j' - 2\ell^2 + 2}{\ell+1} s_{a-2} + (-3j' + 2\ell+1) s_{a-1}\\
    &+ (3j'-\ell-2) s_{b-1} + \frac{-3\ell j + 3j' + \ell^2 - 1}{\ell +1} s_b\Bigg. \Bigg)\label{eq:a5}
\end{align}

\end{proof}

\begin{lemma}(\textbf{bounding $T_1$}) \label{lem:t1}
Consider a bin $[a,b]$. Let $\bs p_t$ be the predictions of FLH-SIONS algorithm with parameters $\epsilon = 2$, $C = 20$ and exp-concavity factor $\sigma$. Suppose $\bs \alpha$ and $\bs \beta$ are as defined in Lemma \ref{lem:t2}. For any $\bs \mu \in \{\bs \alpha, \bs \beta \}$ FLH-SIONS satsifies:
\begin{align}
    \sum_{t=a}^b f_t(p_t) - f_t(\bs \mu^T \bs x_t)
    &\le  256 + \frac{1}{2 \sigma} \log(1+\sigma n / 2) + \frac{4}{\sigma}\log n.
\end{align}

\end{lemma}
\begin{proof}
We will derive the guarantee for $\bs \mu = \bs \alpha$. The guarantee for $\bs \mu = \bs \alpha$ follows similarly.

Let's begin by calculating $\bs v:= A^{-1} \sum_{j=a}^b f_j'(\bs x_j^T \bs \beta) \bs x_j$.

We have,
\begin{align}
    |\bs v[1]| &= \left|\frac{2}{(\ell-1) \ell} \sum_{j=1}^\ell (2\ell + 1 - 3j) f'_{(j+a-1)}(\bs x_{j+a-1}^T \bs \beta) \right|\\
    &_{(a)}\le \frac{2}{(\ell-1) \ell} \cdot 2 \ell (\ell-1)\\
    &= 4, \label{eq:c1}
\end{align}
where line (a) is obtained via Lipschitzness and Holder's inequality $\bs x^T \bs y \le \| \bs x\|_1 \| \bs y\|_\infty$ and the fact that $|2\ell+1-3j| \le 2(\ell-1)$ for all $ j \in [1,\ell]$.

Similarly
\begin{align}
    |\bs v[2]|
    &= \left| \frac{2}{(\ell-1) \ell (\ell+1)} \sum_{j=1}^\ell (-3(\ell+1) + 6j) f'_{(j+a-1)}(\bs x_{j+a-1}^T \bs \beta) \right|\\
    &\le \frac{2}{(\ell-1) \ell (\ell+1)} \cdot 3  \ell (\ell-1)  \\
    &= \frac{6}{(\ell+1)}, \label{eq:c2}
\end{align}
where we used $|-3(\ell+1) + 6j| \le 3(\ell-1)$ for all $ j \in [1,\ell]$.

Combining Eq.\eqref{eq:c1} and \eqref{eq:c2} we conclude that 
\begin{align}
    |\bs v^T \bs x_j |
    &\le 4 + (j-a+1) \frac{6}{(\ell+1)}\\
    &\le 10,
\end{align}
where the last line follows due to the fact $(j-a+1) \le \ell$.

Hence by Triangle inequality we have,
\begin{align}
    |\bs \alpha^T \bs x_j|
    \le | \bs \beta^T \bs x_j | + 10. \label{eq:c3}
\end{align}

Further note that
\begin{align}
    \| \bs v\|_2 \le 8
\end{align}

Notice that $\bs \beta = \bs A^{-1} \sum_{j=a}^\ell u_{j} \bs x_j $ which have similar functional form as $\bs v$. Since $|u_j| \le B$ for all $j \in [n]$, by following similar arguments used in bounding $\bs v$ we obtain $| \bs \beta^T \bs x_j | \le 10$ and

\begin{align}
\| \bs \beta\|_2 \le 8.    
\end{align}

Continuing from \eqref{eq:c3} we get
\begin{align}
    | \bs \alpha^T \bs x_j|
    &\le 20. \label{eq:bounded-pred}
\end{align}

Further,

\begin{align}
    \| \bs \alpha\|_2 
    &\le \| \bs \beta\|_2 + \| \bs v\|_2\\
    &\le 16 \label{eq:alpha-bound}
\end{align}
Since the losses $f_t$ are $\sigma$ exp-concave in $[-1,1]$, b+y Theorem 2 in \citep{Luo2016EfficientSO} and Lemma 3.3 in \citep{hazan2007adaptive}, FLH-SIONS with parameters set as in the statement of the Lemma yields a regret of 

\begin{align}
    \sum_{t=a}^b f_t(p_t) - f_t(\bs \alpha^T \bs x_t)
    &\le 256 + \frac{1}{2 \sigma} \log(1+\sigma n / 2) + \frac{4}{\sigma}\log n.
\end{align}

\end{proof}

\begin{lemma}\label{lem:mono} \textbf{(monotonic slopes)}
Consider a bin $[i_s,i_t] \in \cP$ such that the slopes are monotonic (i.e either non-decreasing or non-increasing). Let $p_j$ be the predictions made by the FLH-SIONS algorithm with parameters as set in Lemma \ref{lem:t1}. Then we have,
\begin{align}
    \sum_{j=i_s}^{i_t} f_j(p_j) - f_j(u_j)
    &\le O  \Bigg( \Bigg. \frac{1}{2 \sigma} \log(1+\sigma n / 2) + \frac{4}{\sigma}\log n + 210408 \Bigg. \Bigg)\\ 
    &= \tilde O(1)
\end{align}
\end{lemma}
\begin{proof}
We will consider the case of non-decreasing slopes. The alternate case can be handled similarly.

Assume that the slope within the bin is not constant, otherwise we trivially get logarithmic regret as we need only to compete with the best fixed linear fit which is handled by the static regret of FLH-SIONS in any interval ($\bs \mu = \bs \beta$ in Lemma \ref{lem:t1}).

The optimal solution within a bin of $\cP$ obtained via Proposition \ref{prop:resid-mono} which doesn't have constant slope may touch either $-1$ or $1$ but not both.  Consider the case where the optimal touches $-1$. Then as the slopes are non-decreasing, once it leaves $-1$, it never touches $-1$ again. So we can split the bin $[i_s,i_t]$ into at-most 3 bins $[a,b]$, $[b+1,c]$ and $[c+1,d]$ such that the optimal touches $-1$ only within $[b+1,c]$. (This bin can be empty if the optimal doesn't touch $-1$ anywhere within $[i_s,i_t]$).

Now we will bound the regret within bin $[a,b]$.

Suppose that $s_{a-1} = 1$ and $s_{b} = 1$. If this condition is not satisfied, we can refine the bin $[a,b]$ into at-most 3 bins $[a_1,b_1],[a_2,b_2],[a_3,b_3]$ such that the optimal has constant slope in the first and last bins and $s_{a_2-1} = s_{b_2} = 1$. This is possible because the slopes in $[a,b]$ are non-decreasing.

Let $\Delta: = \|D^2 u_{a:b}\|_1$ and $\ell: = b-a+1$. Let $p$ and $q$ be two numbers in $[0,2]$. Substituting $s_{a-2} = 1-p$, $s_{a-1} = 1$, $s_{b-1} = 1-q$ and $s_b = 1$ into Lemma \ref{lem:t3} and using the fact that $|jM_j + C_j| \le 20\ell \Delta$ for all $j \in [a,b]$ due to Lemma \ref{lem:residual}, we get

\begin{align}
    T_3
    &\le 40\lamda (p+q) \ell \Delta + 200, \label{eq:b3}\end{align}
where we observed that a term arising from Lemma \ref{lem:t3}: $-M_a + M_{b-1} - \sum_{t=a+1}^b |M_t - M_{t-1}| = 0$  as the slopes are non-decreasing.

By making similar sign substitutions in Lemma \ref{lem:t2} and noting that $\bs h = \bs 0$, we get

\begin{align}
    T_2
    &\le \frac{-2\lamda^2}{2(\ell-1) \ell (\ell+1)} \left( (2\ell^2-3\ell+1)p^2 + (2\ell^2+3\ell+1)q^2 + 2(\ell^2-1)pq \right)\\
    &+ \frac{40 \lamda \ell \Delta}{\ell-1} \left(2p(\ell-1)  +q(\ell+2)\right) + \frac{40\lamda \ell^2 \Delta}{(\ell-1)(\ell+1)} \left( p(\ell-1) + q(\ell+1) \right)\\
    &\le \frac{-2\lamda^2}{2(\ell-1) \ell (\ell+1)} \left( (2\ell^2-3\ell+1)p^2 + (2\ell^2+3\ell+1)q^2 + 2(\ell^2-1)pq \right)\\
    &+ 160 \lamda \ell \Delta (p+q) + 160 \lamda \ell \Delta (p+q), \label{eq:b1}
\end{align}

where in  the last line we used the fact that $\ell -1 \ge \ell /2$ and $\ell + 2 \le 2 \ell$  for all $\ell \ge 2$.

Now consider the case where $p \ge q$. Then,

\begin{align}
    (2\ell^2-3\ell+1)p^2 + (2\ell^2+3\ell+1)q^2 + 2(\ell^2-1)pq
    &\ge (2 \ell^2 - 3\ell +1) p^2\\
    &\ge \ell^2 p^2, \label{eq:p}
\end{align}
where the last line holds for all $\ell \ge 3$. (If $\ell \le 3$, the regret within the bin is trivially $O(1)$ appealing to the Lipschitzness of the losses $f_t$ and the boundedness of the predictions and the comparators (see proof of Lemma \ref{lem:t1})). Thus by using $\ell-1 \le \ell$ and $\ell+1 \le 2\ell$, we get
\begin{align}
    \frac{-2\lamda^2}{2(\ell-1) \ell (\ell+1)} \left( (2\ell^2-3\ell+1)p^2 + (2\ell^2+3\ell+1)q^2 + 2(\ell^2-1)pq \right)
    &\le \frac{-\lamda^2 p^2}{2 \ell}. \label{eq:b2}
\end{align}

Combining Eq. \eqref{eq:b1} and \eqref{eq:b2} and using the fact that $p \ge q$, we have
\begin{align}
    T_2 \le \frac{-\lamda^2 p^2}{2 \ell} + 640\lamda \ell \Delta p . \label{eq:b4}
\end{align}

Similarly from \eqref{eq:b3} using $p \ge q$ we get

\begin{align}
    T_3
    &\le 40\lamda (p+q) \ell \Delta + 200\\
    &\le 80 \lamda p \ell \Delta + 200 \label{eq:b5}
\end{align}

Combining Eq. \eqref{eq:b4} and \eqref{eq:b5} we have
\begin{align}
    T_2 + T_3
    &\le \frac{-\lamda^2 p^2}{2 \ell} + 648 \lamda p \ell \Delta + 200\\
    &= -\left( \frac{\lamda p}{\sqrt{2\ell}}  - 648 \sqrt{2} \ell^{3/2} \Delta \right) + 209952  \ell^3 \Delta^2 + 200\\
    &\le 210152,
\end{align}
where in the last line we dropped the negative term and used the facts that $\Delta \le 1/\ell^{3/2}$.

For the case of $q \ge p$, we have
\begin{align}
    (2\ell^2-3\ell+1)p^2 + (2\ell^2+3\ell+1)q^2 + 2(\ell^2-1)pq
    &\ge (2 \ell^2 - 3\ell +1) q^2\\
    &\ge \ell^2 q^2,
\end{align}
where the last line holds for all $\ell \ge 3$. This is the same expression as in Eq.\eqref{eq:p} with $p$ replaced by $q$. By replacing $p$ with $q$ in the arguments we detailed for the case of $p \ge q$ earlier, we arrive at the same conclusion that $T_2 + T_3 \le 210152$ even when $q \ge p$. (If $\ell \le 3$, the regret within the bin is trivially $O(1)$ appealing to the Lipschitzness of the losses $f_t$ and the boundedness of the predictions and the comparators (see proof of Lemma \ref{lem:t1}))

Similar bound on $T_2 + T_3$ can be shown for bin $[c+1,d]$ by essentially the same arguments. 

Hence through Lemma \ref{lem:t1} we have the dynamic regret in bins $[a,b]$ to be:

\begin{align}
    \sum_{t=a}^b f_t(p_t) - f_t(u_t)
    &\le 256 + \frac{1}{2 \sigma} \log(1+\sigma n / 2) + \frac{4}{\sigma}\log n + 210152\\
    &= \tilde O(1)
\end{align}

Similarly, the regret within bin $[c+1,d]$ is also bounded by the above expression.

As the slope within bin $[b+1,c]$ is constant, the regret incurred within this bin is trivially bounded by $256 + \frac{1}{2 \sigma} \log(1+\sigma n / 2) + \frac{4}{\sigma}\log n$ due to Lemma \ref{lem:t1}.

Adding the regret incurred across the bins $[a,b]$, $[b+1,c]$ and $[c+1,d]$ together yields the lemma.

\end{proof}

Next, we will focus on bounding $T_2+T_3$ for general non-monotonic bins in $\cP$.

\begin{lemma}(\textbf{non-monotonic slopes}) \label{lem:non-mono}
Consider a bin $[i_s,i_t] \in \cP$ such that the slopes are not monotonic. Let $p_j$ be the predictions made by the FLH-SIONS algorithm with parameters as set in Lemma \ref{lem:t1}. Then we have,
\begin{align}
    \sum_{j=i_s}^{i_t} f_j(p_j) - f_j(u_j)
    &\le O \left(\frac{1}{\sigma} \log(1+\sigma n) + \frac{12}{\sigma} \log n  + 1 \right)\\
    &= \tilde O(1)
\end{align}
\end{lemma}
\begin{proof}
Let $[a,b] \in \cP$ be a bin where the slope is not monotonic and not constant.

Assume that $|s_{a-1}| = |s_b| = 1$. Otherwise we can split the original bin into at-most 3 bins $[a,b_1-1],[b_1,b_2],[b_2+1,b]$ such that $|s_{b_1-1}| = |s_{b_2}| = 1$ and slopes are constant in the the other two bins. This is possible because slope in $[a,b]$  is not constant or monotonic.

For a bin $[a,b]$ we define \textbf{boundary signs} to be $s_{a-2},s_{a-1},s_{b-1}$ and $s_b$.

First, we will study the case where the offline optimal touches the boundary $-1$ at two point $r$ and $w$ with $r < w$. The case of arbitrary number of boundary touches will be discussed towards the end. (All arguments can be mirrored appropriately for the case where optimal touches boundary 1).

In what follows we use the notations in the proof of Lemma \ref{lem:t2}.
From Eq.\eqref{eq:st-kkt} we have
\begin{align}
    \bs g + \bs h
    &=\lamda \bs \mu + \gamma_r^- \bs x_r + \gamma_w^- \bs x_w, \label{eq:gnh}
\end{align}
where $\bs \mu \in \mathbb{R}^2$ is a vector depending on the boundary signs and the length $\ell:=b-a+1$. $\bs x_r = [1,r-a+1]^T$ and $\bs x_w$ defined similarly.

Since $\bs g + \bs h$ is an affine map of $[\lamda, \gamma_r^-, \gamma_w^-]^T$ and since $\bs A$ is positive definite for $\ell \ge 2$, we conclude that $\|\bs g  + \bs h\|_{\bs A^{-1}}^2$ is jointly convex in $\lamda, \gamma_r^-, \gamma_w^-$ via appealing to the convexity of squared L2 norm.

First let's focus on the case where boundary signs obey $s_{a-1} = 1$ and $s_{b} = -1$. Let $s_{a-2} = 1-p$ and $s_{b-1} = -1+q$ for some $p,q \in [0,2]$.

Making these sign substitutions in Lemma \ref{lem:t2}, we get:

\begin{align}
    \| \bs g\|_{\bs A^{-1}}^2
    &= \frac{2\lamda^2}{(\ell-1)\ell(\ell+1)} \left( (2\ell^2 - 3\ell+1)p^2 + (2\ell^2+3\ell+1)q^2 - (2\ell^2-2)pq + 12(\ell-1)p - 12(\ell+1)q + 24 \right).
\end{align}

\begin{align}
    \langle \bs g, \bs A^{-1} \bs h \rangle
    &= \frac{\lamda}{(\ell-1)\ell(\ell+1)} \left(-24-6\ell(p-q)+6(p+q)\right)(r'\gamma_r^- + w' \gamma_w^{-})+\\
    &+ \frac{\lamda}{(\ell-1)\ell(\ell+1)} \left( 2\ell^2 (2p-q) - 6\ell q - 4(p+q) + 12(\ell+1) \right) (\gamma_r^- +  \gamma_w^-),
\end{align}
where $r' = r-a+1$ and $w' = w-a+1$.
    
Let $\Delta:=\ell^{-3/2}$. By using equation \eqref{eq:a3} and the facts $\ell-1 \ge \ell/2$, $\ell+1 < \ell$, $p,q \in [0,2]$ and triangle inequality, we bound

\begin{align}
    \langle \bs A^{-1} \bs g, \sum_{j=a}^b \bs x_j (f_j'(\bs x_j^T \bs \beta) - f_j'(u_j)) \rangle
    &\le_{(a)} \frac{40 \lamda \ell \Delta}{\ell-1} \left | 2p(\ell-1) -q(\ell+2) + 6\right|\\
    &+ \frac{40 \lamda \ell^2 \Delta}{\ell-1} \left| 3q(\ell+1) + p(1-\ell) -4 \right|\\
    &_{(b)}\le 640 \lamda \ell \Delta (p+q) + 800 \lamda \Delta,
\end{align}
where the line (a) is obtained by equation \eqref{eq:a3} and making the boundary sign substitutions. Line (b) is obtained using the facts $\ell-1 \ge \ell/2$, $\ell+2 \le 2\ell$ whenever $\ell \ge 2$ and $p,q \in [0,2]$ along with triangle inequality.

From Eq.\eqref{eq:a6}, by using similar triangle inequality based arguments and the fact that $|\tilde \Gamma| \le \ell |\Gamma |$ by Holder's inequality and Corollary \ref{cor:dualprops} in as above we obtain
\begin{align}
    \langle \bs A^{-1} \bs h, \sum_{j=a}^b \bs x_j (f_j'(\bs x_j^T \bs \beta) - f_j'(u_j)) \rangle
    &\le 1200 \ell \Delta (\gamma_r^- + \gamma_w^-).
\end{align}

To bound T3 we observe from Lemma \ref{lem:t3}
\begin{align}
    T_3
    &= 200 + \lamda \Bigg ( \Bigg. (s_{a-1} - s_{a-2})(M_{a}+C_a) - (s_{b} - s_{b-1})(\ell M_b + C_b) \\
   & - s_{a-1} M_a + s_{b-1}M_{b-1} - \sum_{t=2}^{\ell} |M_t - M_{t-1}|  \Bigg. \Bigg ) + 20\ell \Delta (\gamma_r^- + \gamma_w^-)\\
   &\le 200 + \lamda \Bigg ( \Bigg. |(s_{a-1} - s_{a-2})(M_{a}+C_a)| + |(s_{b} - s_{b-1})(\ell M_b + C_b)| \\
   & + |M_a| + |M_{b-1}| + \Delta  \Bigg. \Bigg ) + 20\ell \Delta (\gamma_r^- + \gamma_w^-)\\
   &\le 200 + 80 \lamda \ell \Delta (p+q) +  3 \lamda \Delta+ 20\ell \Delta (\gamma_r^- + \gamma_w^-),
\end{align}where in the last line we used the fact that $|(j-a+1)M_j + C_j| \le 20 \ell \Delta$ from Lemma \ref{lem:residual}.

Recall that $\Delta = \ell^{-3/2}$. Combining all the above equations / inequalities above and Eq. \eqref{eq:a4}, define:

\begin{align}
    T(\lamda,\gamma_r^-,\gamma_w^-)
    &:=\lamda^2 \left( (2\ell^2 - 3\ell+1)p^2 + (2\ell^2+3\ell+1)q^2 - (2\ell^2-2)pq + 12(\ell-1)p - 12(\ell+1)q + 24 \right)\\
    &+\left( (2\ell+1)(\ell+1) (\gamma_r^- + \gamma_w^-)^2 - 6(\gamma_r^- + \gamma_w^-) (r'\gamma_r^- + w'\gamma_w^-) (\ell+1) + 6 (r'\gamma_r^- + w'\gamma_w^-)^2 \right)\\
    &+\lamda \left(-24-6\ell(p-q)+6(p+q)\right)(r'\gamma_r^- + w' \gamma_w^{-})\\
    &+ \lamda \left( 2\ell^2 (2p-q) - 6\ell q - 4(p+q) + 12(\ell+1) \right) (\gamma_r^- +  \gamma_w^-)\\
    &-  ((\ell-1)\ell(\ell+1)) \left(  720 \lamda \ell \ell^{-3/2} (p+q) + 803 \lamda \ell^{-3/2}
    + 1220 \ell \ell^{-3/2} (\gamma_r^- + \gamma_w^-)  \right).\label{eq:tobj}
\end{align}

We have,
\begin{align}
    T_2 + T_3 \le -\frac{T(\lamda,\gamma_r^-,\gamma_w^-)}{(\ell-1)\ell(\ell+1)} + 200. \label{eq:t2t3}
\end{align}

The expression in Eq.\eqref{eq:tobj} can be compactly written as:
\begin{align}
    T(\lamda, \check \gamma_r^-, \check \gamma_w^-)
    &= 0.5 \cdot (\ell-1) \ell (\ell+1) \|\bs g + \bs h \|_{\bs A^{-1}}^2 + \Phi(\lamda,\check \gamma_r^- + \check \gamma_w^-)\\
    &:= Q(\lamda,\gamma_r^- + \gamma_w^-, r\gamma_r^- + w\gamma_w^-), \label{eq:Qobj}
\end{align}
where $\bs g + \bs h$ is as in Eq.\eqref{eq:gnh} (which only depends on the boundary signs and  $\lamda, \gamma_r^- + \gamma_w^-$ and $r\gamma_r^- + w\gamma_w^-$) and $\Phi (\lamda,\check \gamma_r^-, \check \gamma_w^-)$ is a linear function of its arguments namely,
\begin{align}
    \Phi(\lamda,\check \gamma_r^- + \check \gamma_w^-)
    &= -(\ell-1)\ell(\ell+1) \left( 20 \lamda \ell \ell^{-3/2} (p+q) + 803 \lamda \ell^{-3/2}
    + 1220 \ell \ell^{-3/2} (\gamma_r^- + \gamma_w^-)\right)
\end{align}

Since we have established earlier that $\| \bs g\|_{\bs A^{-1}}^2$ is convex in $\lamda,\gamma_r^-,\gamma_w^-$ we will certainly have $T(\lamda,\gamma_r^-,\gamma_w^-)$ as a function jointly convex in its arguments.

The function $B$ referred in Appendix \ref{app:over} is defined to be:
\begin{align}
    B(\lamda, \gamma_r^-, \gamma_w^-;r,w)
    &:=  -\frac{T(\lamda,\gamma_r^-,\gamma_w^-)}{(\ell-1)\ell(\ell+1)} + 200, \label{eq:bobj}
\end{align}
with $r'$ and $w'$ in Eq.\eqref{eq:tobj} to be taken as $r' = r-i_s+1$ and $w' = w-i_s+1$ , $\ell = i_t-i_s+1$ and $T(\lamda,\gamma_r^-,\gamma_w^-)$ is as in Eq.\eqref{eq:tobj}.

So we consider the following convex optimization problem:
\begin{mini!}|s|[2]                   
    {\lamda, \gamma_r^-, \gamma_w^-}                               
    {T(\lamda,\gamma_r^-,\gamma_w^-)}   
    {}{}
    \label{eq:opt}             
    \addConstraint{\lamda}{\ge 0}
\end{mini!}

Note that in the program above we do \emph{unconstrained} minimization over $\gamma_r^-$ and $\gamma_w^-$. Doing so can only further decrease the objective function leading to a valid upper bound on $T_2 + T_3$.

First we will perform a partial minimization wrt the variables $ \gamma_r^-$ and $ \gamma_w^-$. Differentiating the objective wrt $ \gamma_r^-$ and setting to zero yields:

\begin{align}
    &\left(2(2\ell^2 + 3\ell + 1) - 12(\ell+1)r' + 12(r')^2\right) \hat \gamma_r^-\\
    &+ \left(2(2\ell^2+3\ell+1) - 6(\ell+1)(r'+w')+12r'w' \right) \hat \gamma_w^-\\
    &= \lamda r' \left(24+6\ell(p-q)-6(p+q) \right) - \lamda \left( 2\ell^2(2p-q) - 6\ell q - 4(p+q) + 12(\ell+1) \right)\\
    & +1220 \ell^2(\ell^2-1)\ell^{-3/2}.
\end{align}

Similarly differentiating the objective wrt $ \gamma_w^-$ and setting to zero yields:

\begin{align}
    &\left(2(2\ell^2 + 3\ell + 1) - 12(\ell+1)w' + 12(w')^2\right) \hat \gamma_w^-\\
    &+ \left(2(2\ell^2+3\ell+1) - 6(\ell+1)(r'+w')+12r'w' \right) \hat \gamma_r^-\\
    &= \lamda w' \left(24+6\ell(p-q)-6(p+q) \right) - \lamda \left( 2\ell^2(2p-q) - 6\ell q - 4(p+q) + 12(\ell+1) \right)\\
    & +1220 \ell^2(\ell^2-1)\ell^{-3/2}.
\end{align}

Solving the above two equations yields:

\begin{align}
\hat \gamma_r^-
    &=\frac{\begin{aligned}& w^\prime \lamda \ell^{2} p + w^\prime \lamda \ell^{2} q - w^\prime \lamda p - w^\prime \lamda q\\
    &+ 1220 w^\prime \ell^{0.5} - 1220 w^\prime \ell^{2.5} - \lamda \ell^{3} q - \lamda \ell^{2} p\\
    &- \lamda \ell^{2} q + 2 \lamda \ell^{2} + \lamda \ell q + \lamda p \\
    &+ \lamda q - 2 \lamda - 610 \ell^{0.5} - 610 \ell^{1.5} + 610 \ell^{2.5} + 610 \ell^{3.5}\end{aligned}}{r^\prime \ell^{2} - r^\prime - w^\prime \ell^{2} + w^\prime},
\end{align}

and
\begin{align}
    \hat \gamma_w^-
    &=\frac{\begin{aligned}& - r^\prime \lamda \ell^{2} p - r^\prime \lamda \ell^{2} q + r^\prime \lamda p + r^\prime \lamda q\\
    &- 1220 r^\prime \ell^{0.5} + 1220 r^\prime \ell^{2.5} + \lamda \ell^{3} q + \lamda \ell^{2} p\\
    &+ \lamda \ell^{2} q - 2 \lamda \ell^{2} - \lamda \ell q - \lamda p - \lamda q\\
    &+ 2 \lamda + 610 \ell^{0.5} + 610 \ell^{1.5} - 610 \ell^{2.5} - 610 \ell^{3.5}\end{aligned}}{r^\prime \ell^{2} - r^\prime - w^\prime \ell^{2} + w^\prime}.
\end{align}

Substituting the above two expression we get:
\begin{align}
    T(\lamda, \hat \gamma_r^-, \hat \gamma_w^-)
    &= \frac{\begin{aligned}&- 797 \lamda \ell^{2.0} - 1780 \lamda \ell^{3.0} p - 1780 \lamda \ell^{3.0} q\\
    &+ 2391 \lamda \ell^{4.0} + 5340 \lamda \ell^{5.0} p + 5340 \lamda \ell^{5.0} q - 2391 \lamda \ell^{6.0}\\
    &- 5340 \lamda \ell^{7.0} p - 5340 \lamda \ell^{7.0} q + 797 \lamda \ell^{8.0} + 1780 \lamda \ell^{9.0} p + 1780 \lamda \ell^{9.0} q\\
    &+ 744200 \ell^{3.5} - 2232600 \ell^{5.5} + 2232600 \ell^{7.5} - 744200 \ell^{9.5}\end{aligned}}{\ell^{2.5} - 2 \ell^{4.5} + \ell^{6.5}} \label{eq:partopt}
\end{align}

Looking at Eq.\eqref{eq:partopt} we notice that it is a linear fucntion of $\lamda$ which defined the function $\cL(\lamda)$ mentioned in Appendix \ref{app:over}:
\begin{align}
    \cL(\lamda)
    &= \frac{\begin{aligned}&- 797 \lamda \ell^{2.0} - 1780 \lamda \ell^{3.0} p - 1780 \lamda \ell^{3.0} q\\
    &+ 2391 \lamda \ell^{4.0} + 5340 \lamda \ell^{5.0} p + 5340 \lamda \ell^{5.0} q - 2391 \lamda \ell^{6.0}\\
    &- 5340 \lamda \ell^{7.0} p - 5340 \lamda \ell^{7.0} q + 797 \lamda \ell^{8.0} + 1780 \lamda \ell^{9.0} p + 1780 \lamda \ell^{9.0} q\\
    &+ 744200 \ell^{3.5} - 2232600 \ell^{5.5} + 2232600 \ell^{7.5} - 744200 \ell^{9.5}\end{aligned}}{\ell^{2.5} - 2 \ell^{4.5} + \ell^{6.5}} \label{eq:linlamda}
\end{align}

We observe that the leading term (i.e terms whose magnitude is biggest) in the denominator is a positive quantity namely $\ell^{6.5}$. The leading term in the numerator that contains $\lamda$ grows as $1780\lamda \ell^9 (p+q) + 797 \lamda \ell^8$. So the unconstrained minimum of this linear function is attained at $\lamda = -\infty$.

Hence the constrained minimum (with constraint $\lamda \ge 0)$ of the optimization problem \ref{eq:opt} is attained at $\lamda = 0$. We calculate the optimal objective to the constrained problem via Eq.\eqref{eq:partopt} as
\begin{align}
    T(0,\hat \gamma_r^-, \hat \gamma_w^-)
    &= \frac{744200\left(\ell^{1.5} - 3 \ell^{3.5} + 3 \ell^{5.5} - \ell^{7.5}\right)}{ \ell^{0.5} - 2 \ell^{2.5} + \ell^{4.5}},
\end{align}
where we consider bins with length $\ell \ge 14$.

Since $\ell^4 \ge 2\ell^2$ for all $\ell \ge 2$, we continue from the previous display to obtain:
\begin{align}
    T(0,\hat \gamma_r^-, \hat \gamma_w^-)
    &\ge -744200 \cdot (1+3+3+1) \frac{\ell^{7.5}}{ \ell^{\ell.5}}\\
    &= -5953600 \ell^3,
\end{align}

Hence we have
\begin{align}
    \frac{T(0,\hat \gamma_r^-, \hat \gamma_w^-)}{(\ell-1) \ell (\ell+1)}
    &\ge \frac{-5953600 \ell^3}{(\ell-1) \ell (\ell+1)}\\
    &\ge -11907200,
\end{align}
where in the last line we used the fact that $\ell-1 \ge \ell/2$ is satisfied for all $\ell \ge 14$ and $\ell+1 > \ell$.

Hence continuing from Eq.\eqref{eq:t2t3} we conclude that
\begin{align}
    T_2+T_3
    &\le (11907200 + 200)\\
    &= 11907400.
\end{align}

The term $T_1$ can be bound as
\begin{align}
    T_1
    &\le  256 + \frac{1}{2 \sigma} \log(1+\sigma n / 2) + \frac{4}{\sigma}\log n\\
    &= \tilde O(1),
\end{align}
by Lemma  \ref{lem:t1}.

Now suppose that the offline optimal within bin $[a,b]$ touches boundary $-1$ more than two times. In this case we propose a reduction to the previous type of analysis where only $\gamma_r^-$ and $\gamma_w^-$ are potentially non-zero.

The reduction is facilitated by two observations:
\begin{enumerate}
    \item While performing the minimization of function $T(\lamda, \gamma_r^-, \gamma_w^-)$ in Eq.\eqref{eq:tobj} via the optimization problem \ref{eq:opt} we neither used the fact that $r$ and $w$ are integers nor constrained any bounds on them as well
    
    \item The partially minimized objective in Eq.\eqref{eq:partopt} fortunately doesn't depend on neither $r$ nor $w$.
    
\end{enumerate}

Now let's consider the case where arbitrary number of $\gamma_j^-$, $j \in [a,b]$ can be non-zero. We can then write,
\begin{align}
    \Gamma 
    &= \sum_{j=a}^b \gamma_j^-\\
    &= \check \gamma_r^- + \check \gamma_w^-,
\end{align}
where $\check \gamma_r^- := \gamma_1^-$   and $\check \gamma_w^- = \Gamma - \check \gamma_1^-$.

Define $r' := 1$ and $w' := \frac{\sum_{j=a}^b j' \gamma_j^- - \check \gamma_r^-}{\check \gamma_w^-} = \frac{\tilde \Gamma - \check \gamma_r^-}{\check \gamma_w^-}$ where we assume that $\check \gamma_w^- > 0$ (otherwise, we fall back to the earlier analysis).

With these re-definitions we note that 
\begin{align}
    T_2+ T_3
    &\le - \frac{T(\lamda, \check \gamma_r^-, \check \gamma_w^-)}{(\ell-1) \ell (\ell+1)},
\end{align}
still holds. Further, $T(\lamda, \check \gamma_r^-, \check \gamma_w^-)$ is jointly convex in its arguments. This can be seen as follows: Note that $T(\lamda, \check \gamma_r^-, \check \gamma_w^-)$ assumes the form
\begin{align}
    T(\lamda, \check \gamma_r^-, \check \gamma_w^-)
    &= 0.5 \cdot (\ell-1) \ell (\ell+1) \|\bs g + \bs h \|_{\bs A^{-1}}^2 + \Phi (\lamda,\check \gamma_r^- + \check \gamma_w^-),
\end{align}
where $\Phi(\lamda,\check \gamma_r^- + \check \gamma_w^-)$ is an affine function of its arguments and
\begin{align}
    \bs h 
    &= [\Gamma, \tilde \Gamma]^T\\
    &= [\check \gamma_r^- + \gamma_w^-, r' \check \gamma_r^- + w'\check \gamma_w^- ]^T,
\end{align}
where the last line follows due to our re-parametrizations. By following essentially same arguments as earlier for proving  convexity of $T(\lamda,  \gamma_r^-, \gamma_w^-)$ we conclude that $T(\lamda, \check \gamma_r^-, \check \gamma_w^-)$ is also jointly convex in its arguments.

This completes our reduction to the case of two-boundary touches and rest of analysis proceeds by minimizing $T(\lamda, \check \gamma_r^-, \check \gamma_w^-)$ as earlier.

We now consider the case where $s_{a-1} = s_b = 1$. We can split the original bin $[a,b]$ into two sub-bins $[a_1,b_1]$ and $[a_2,b_2]$ with $a_2 = b_1+1$ such that (i) $s_{b_1} = -1$ with $u_{b_1+1} - u_{b_1} > u_{a_2+1} - u_{a_2}$ and (ii) the slopes are non-decreasing within $[a_2,b_2]$. This can be achieved by picking $b_1$ as the last point within $[a,b]$ where $u_{b_1+1} - u_{b_1} > u_{b_1+2} - u_{b_1+1}$.

In the bin $[a_1,b_1]$ we apply the previous analysis to bound regret by $\tilde O(1)$. For the bin $[a_2,b_2]$ we resort to Lemma \ref{lem:mono} to bound regret by $\tilde O(1)$.

The analysis for the case of boundary signs assignments $s_{a-1} = -1$ and $s_b = 1$ as well as $s_{a-1} = -1$ and $s_b = -1$ can be done similarly.

Adding the regret bounds across all newly formed bins due to potential splitting yields the lemma.

\end{proof}

Next, we provide the full regret guarantee in a uni-variate setting.
\begin{theorem} \label{thm:main}
Let $ p_t$ be the predictions of FLH-SIONS algorithm with parameters $\epsilon = 2$, $C = 20$ and exp-concavity factor $\sigma$. Under Assumptions A1-A4, we have that,
\begin{align}
    \sum_{t=1}^n f_t( p_t) - f_t( w_t)
    &= \tilde O (n^{1/5} C_n^{2/5}  \vee 1 ),
\end{align}
for any comparator sequence $w_{1:n} \in \mathcal{TV}^{(1)}(C_n)$. Here $\tilde O$ hides poly-logarithmic factors of $n$ and $a \vee b = \max\{ a,b\}$.
\end{theorem}
\begin{proof}
The proof is complete by adding the $\tilde O(1)$ dynamic regret bound from Lemmas \ref{lem:mono} and \ref{lem:non-mono} across $O(n^{1/5}C_n^{2/5} \vee 1)$ bins in the partition $\cP$.

\end{proof}

The proof of Lemma \ref{lem:nd} stated in Appendix \ref{app:over} is similar to the arguments used to derive Eq.\eqref{eq:nd1}. We record it for the sake of completeness.

\lemND*
\begin{proof}
We follow the same notations used in defining Lemma \ref{lem:nd}.

Let's begin by calculating $\bs v:= \bs A^{-1} \sum_{j=a}^b f_j'(\bs x_j^T \bs \beta) \bs x_j$.

We have,
\begin{align}
    |\bs v[1]| &= \left|\frac{2}{(\ell-1) \ell} \sum_{j=1}^\ell (2\ell + 1 - 3j) f'_{(j+a-1)}(\bs x_{j+a-1}^T \bs \beta) \right|\\
    &_{(a)}\le \frac{2}{(\ell-1) \ell} \cdot 2 \ell (\ell-1)\\
    &= 4, \label{eq:c11}
\end{align}
where line (a) is obtained via Lipschitzness and Holder's inequality $\bs x^T \bs y \le \| \bs x\|_1 \| \bs y\|_\infty$ and the fact that $|2\ell+1-3j| \le 2(\ell-1)$ for all $ j \in [1,\ell]$.

Similarly
\begin{align}
    |\bs v[2]|
    &= \left| \frac{2}{(\ell-1) \ell (\ell+1)} \sum_{j=1}^\ell (-3(\ell+1) + 6j) f'_{(j+a-1)}(\bs x_{j+a-1}^T \bs \beta) \right|\\
    &\le \frac{2}{(\ell-1) \ell (\ell+1)} \cdot 3  \ell (\ell-1)  \\
    &= \frac{6}{(\ell+1)}, \label{eq:c22}
\end{align}
where we used $|-3(\ell+1) + 6j| \le 3(\ell-1)$ for all $ j \in [1,\ell]$.

Combining Eq.\eqref{eq:c11} and \eqref{eq:c22} we conclude that 
\begin{align}
    |\bs v^T \bs x_j|
    &= 4 + (j-a+1) \frac{6}{(\ell+1)}\\
    &\le 10, \label{eq:c44}
\end{align}
where the last line follows due to the fact $(j-a+1) \le \ell$.

Hence by Triangle inequality we have 
\begin{align}
    |\bs \alpha^T \bs x_j|
    \le | \bs \beta^T \bs x_j | + 10. \label{eq:c33}
\end{align}

Now we bound $|\bs \beta^T \bs x_j|$ using similar arguments. We have $\bs v':= \bs A^{-1} \sum_{j=a}^b u_j \bs x_j$. Now noting that $|u_j| \le 1$ by Assumption A1 and using similar arguments used to obtain Eq.\eqref{eq:c44} we conclude that

\begin{align}
    |\bs \beta^T \bs x_j| \le 10. \label{eq:boundedlstsq}
\end{align}

So continuing from Eq.\eqref{eq:c33} we have $|\bs \alpha^T \bs x_j| \le 20$.

For some $\bs z = t \bs \alpha + (1-t) \bs \beta$, $t \in [0,1]$ we have by Taylor's theorem that
\begin{align}
    F(\bs \alpha) - F(\bs \beta)
    &= - \langle \grad F(\bs \beta), \bs A^{-1} \grad F(\bs \beta) \rangle + \frac{1}{2} \|  \bs A^{-1} \grad F(\bs \beta)\|_{\grad^2 F(\bs z)}^2\\
    &\le - \langle \grad F(\bs \beta), \bs A^{-1} \grad F(\bs \beta) \rangle +  \frac{1}{2} \|  \bs A^{-1} \grad F(\bs  \beta)\|_{\bs A}^2\\
    &= -\frac{1}{2} \|\grad F(\bs \beta)\|_{\bs A^{-1}}^2,
\end{align}
where in the first inequality we used that fact that $\grad^2 F(\bs z) \preccurlyeq \bs A$ due to the fact that the functions $f_j$ are $1$ gradient Lipschitz in $[-20,20]^d$ via Assumption A3.
\end{proof}

\subsection{Multi-dimensional setting} \label{app:hd}

\begin{figure}[h!]
	\centering
	\fbox{
		\begin{minipage}{14 cm}
		    \code{splitMonotonic}: Inputs- (1) offline optimal sequence (2) A bin $[i_s,i_t]$ (3) A coordinate $k \in [d]$
            \begin{enumerate}

                \item Compute $\bs z_j[k] = \bs u_{j+1}[k] - \bs u_j[k]$
                
                \item If $\bs z[k]$ is constant in $[i_s,i_t]$ return $\{ i_s,i_t\}$.

                \item If $\bs z[k]$ is non-decreasing (non-increasing) across $[i_s,i_t]$: //\texttt{ensure equal boundary signs (see caption) for bin } $[b+1,c]$ \texttt{ below}.
                    
                    \begin{enumerate}
                        \item Split $[i_s,i_t]$ into at-most three bins $[i_s,b],[b+1,c],[c+1,i_t]$ such that $\bs z_j[k]$ remains constant in the first and last bins. Further $\bs z_{b+1}[k] > (<) \bs z_b[k]$ and $\bs z_{c+1}[k] > (<) \bs z_c[k]$.
                        
                        \item Return $\{i_s,b,b+1,c,c+1,i_t \}$
                    \end{enumerate}
            \end{enumerate}
		\end{minipage}
	}
	\caption{\emph{\code{splitMonotonic} procedure. If line 3 is replaced by ``If $\bs z[k]$ is non-increasing ...'', then we propagate that change by replacing the symbols $>/<$ in the lines below 3 by the bracketed statements next to it. For a bin $[a,b]$, we refer to $s_{a-1}$ and $s_b$ as the boundary signs.}}
	\label{fig:splitmono}
\end{figure}

\begin{figure}[h!]
	\centering
	\fbox{
		\begin{minipage}{14 cm}
		    \code{generateBins}: Input- (1) offline optimal sequence
            \begin{enumerate}
                \item Form consecutive bins  $[i_s,i_t]$ such that: \texttt{// coarse partition based on TV1 distance}
                \begin{enumerate}
                    \item $\|D^2 \bs u_{i_s:i_t}\|_1 \le 1/\ell_{i_s \rightarrow i_t}^{3/2}$
                    \item $\|D^2 \bs u_{i_s:i_t+1}\|_1 > 1/\ell_{i_s \rightarrow i_t+1}^{3/2}$,
                \end{enumerate}
                where $\ell_{a \rightarrow b} := b-a+1$.

            \item Let the partition of the time horizon be represented as $\c P' := \{[1_s,1_t], \ldots, [i_s,i_t], \ldots [M_s,M_t] \}$  where $M = |\cP'|$.
            
            \item Initialize $\cR \leftarrow \Phi$.
            \item For each bin $[i_s,i_t] \in \cP'$: \texttt{// ensuring} $\bs \gamma_j^+[k] \gamma_j^-[k] = 0$ \texttt{for all} $k \in [d]$
            \begin{enumerate}
                \item $\cR = \cR \cup \{i_s,i_t \}$.
                \item For each coordinate $k \in [d]$:
                \begin{enumerate}
                    \item If $\bs u_{i_s}[k] = 1 (-1)$ and there exists a point $p \in [i_s,i_t]$ such that $\bs u_{p} = -1 (1)$ then $\cR \leftarrow \cR \cup \{p-1,p \}$
                    
                    \item If $\bs u_{i_t}[k] = 1 (-1)$ and there exists a point $p \in [i_s,i_t]$ such that $\bs u_{p} = -1 (1)$ then $\cR \leftarrow \cR \cup \{p-1,p\}$                    
                \end{enumerate}
            \end{enumerate}

            \item Remove duplicates from $\cR$ and form a partition $\cP$ by splitting at each point in $\cR$

                \item Return $\cP$
            \end{enumerate}
		\end{minipage}
	}
	\caption{\emph{\code{generateBins} procedure. If line 7(d) is replaced by ``If $\bs z_p[k] < \bs z_{p-1}[k]$'', then we propagate that change by replacing the symbols $>/<$ in the lines below 7(d) by the bracketed statements next to it. For a bin $[a,b]$, we refer to $s_{a-1}$ and $s_b$ as the boundary signs.}}
	\label{fig:split}
\end{figure}

\begin{lemma} \label{lem:kkt-high}
Consider the following convex optimization problem.
\begin{mini!}|s|[2]                   
    {\tilde {\bs u}_1,\ldots,\tilde {\bs u}_n, \tilde {\bs z}_1,\ldots,\tilde {\bs z}_{n-1}}                               
    {\sum_{t=1}^n f_t(\tilde { \bs u}_t)}   
    {\label{eq:Example1}}             
    {}                                
    \addConstraint{\tilde {\bs z}_t}{=\tilde {\bs u}_{t+2} - 2\tilde {\bs u}_{t+1} + \tilde {\bs u}_t \: \forall t \in [n-2],}    
    \addConstraint{\sum_{t=1}^{n-2} \|\tilde {\bs z}_t\|_1}{\le C_n/n, \label{eq:constr-ec-1d}}  
    \addConstraint{-1}{\le \tilde {\bs u}_t[k] \: \forall t \in [n], \: \forall k \in [d]\label{eq:constr-ec-2d}}
    \addConstraint{\tilde {\bs u}_t[k]}{\le 1 \: \forall t \in [n], \: \forall k \in [d]\label{eq:constr-ec-3d}}
\end{mini!}

Let $\bs u_1,\ldots,\bs u_n,\bs z_1,\ldots,\bs z_{n-2}$ be the optimal primal variables and let $\lambda \ge 0$ be the optimal dual variable corresponding to the constraint \eqref{eq:constr-ec-1d}. Further, let $\bs \gamma_t^- \ge 0 , \bs \gamma_t^+ \ge 0$ (coordinate-wise) be the optimal dual variables that correspond to constraints \eqref{eq:constr-ec-2d} and \eqref{eq:constr-ec-3d} respectively for all $t \in [n]$. Note that $\bs \gamma_t^-,\bs \gamma_t^+ \in \mathbb{R}^d$. By the KKT conditions, we have

\begin{itemize}
    \item \textbf{stationarity: } $\grad f_t({ \bs u}_t) = \lambda \left ( (\bs s_{t-1} - \bs s_{t}) - (\bs s_{t-2} - \bs s_{t-1}) \right) +  \bs \gamma^-_t -  \bs \gamma^+_t$, where $\bs s_t[k] \in \partial|\bs z_t[k]|$ (a subgradient) for $k \in [d]$. Specifically, $\bs s_t[k]=\sign((\bs u_{t+2}[k] - \bs u_{t+1}[k])-(\bs u_{t+1}[k] - \bs u_t[k]))$ if $|(\bs u_{t+2}[k] - \bs u_{t+1}[k])-(\bs u_{t+1}[k] - \bs u_t[k])|>0$ and $\bs s_t[k]$ is some value in $[-1,1]$ otherwise. For convenience of notations, we also define 
    $\bs s_{-1} = \bs s_0 = \bs 0$.
    \item \textbf{complementary slackness: } (a) $\lamda \left( \sum_{t=1}^{n-2} \| {\bs z}_t\|_1 - C_n/n \right) = 0$; (b)  $ \bs \gamma^-_t[k] ( \bs u_t[k] + 1) = 0$ and $ \bs \gamma^+_t[k] ( \bs u_t[k] - 1) = 0$ for all $t \in [n]$.
\end{itemize}

\end{lemma}

The proof of above Lemma is similar to that of Lemma \ref{lem:kkt} and hence omitted.

\begin{lemma} \citep{Luo2016EfficientSO} \label{lem:SON-stat}
Consider an online learning setting where at each round $t$, we are given a feature vector $\bs x_t \in \mathbb{R}^{2}$. Define $\tilde f_t(\bs v) = f_t(\bs x_t^T\bs v[1:2], \ldots, \bs x_t^T \bs v[2d-1:2d])$ for some vector $\bs v \in \mathbb{R}^{2d}$. Let the function $f(\bs r)$ be $\sigma$ exp-concave and $G$ Lipschitz for $\bs r \in \mathbb{R}^d$ with $\| \bs r\|_\infty \le C$. Define $\cK_t := \{\bs w \in \mathbb{R}^{2d} : |\bs x_t^T \bs w[2k-1:2k]| \le C \: \forall k \in [d] \}$. Let $\cK := \cap_{t=1}^T \cK_t$ and $\bs g_t := \grad \tilde f_t(\bs p_t)$. Consider a variant of the algorithm proposed by \citep{Luo2016EfficientSO} where the algorithm makes a prediction $\hat{\bs p}_{t+1} \in \mathbb{R}^{d}$ at round $t+1$ as:
\begin{align}
    &\bs w_{t+1} = \bs p_t - \bs A_t^{-1} \bs g_t\\
    &\bs p_{t+1} = \argmin_{\bs w \in \cK_{t+1}} \|\bs w - \bs w_{t+1} \|_{\bs A_t}\\
    &\hat{\bs p}_{t+1} = \left[\bs x_{t+1}^T\bs p_{t+1}[1:2], \ldots, \bs x_{t+1}^T \bs p_{t+1}[2d-1:2d] \right]^T
\end{align}
where $\bs A_t = \epsilon \bs I + \sum_{s=1}^t \sigma \bs g_s \bs g_s^T$ with $\bs I$ is the identity matrix and $\epsilon$ is an input parameter.

Then for any $\bs w \in \cK$ we have the regret controlled as
\begin{align}
    \sum_{t=1}^T f_t( \hat{\bs p}_t) - \tilde f_t(\bs w)
    &= \sum_{t=1}^T \tilde f_t(\bs p_t) - \tilde f_t(\bs w)\\
    &\le \frac{ \epsilon \| \bs w\|_2^2}{2}  + \frac{2d}{\sigma} \log \left(1 + \frac{\sigma TG^2}{d\epsilon} \right).
\end{align}

We will call this algorithm as SIONS (Scale Invariant Online Newton Step).

\end{lemma}
\begin{proof}
First we show that exp-concavity is invariant to affine transforms. Since $f_t$ is $\sigma$ exp-concave, we have
\begin{align}
    \tilde f_t(\bs w) 
    &\ge \tilde f_t(\bs v) + \Bigg \langle \Bigg. \grad f_t(\bs x_t^T\bs v[1:2], \ldots,\bs x_t^T \bs v[2d-1:2d]),\\
    &[\bs x_t^T(\bs w[1:2] - \bs v[1:2]),\ldots,\bs x_t^T(\bs w[2d-1:2d] - \bs v[2d-1:2d])]^T] \Bigg. \Bigg \rangle \\
    &+ \frac{\sigma}{2} \Bigg ( \Bigg. \Bigg \langle \Bigg. \grad f_t(\bs x_t^T\bs v[1:2], \ldots,\bs x_t^T \bs v[2d-1:2d]),\\
    &[\bs x_t^T(\bs w[1:2] - \bs v[1:2]),\ldots,\bs x_t^T(\bs w[2d-1:2d] - \bs v[2d-1:2d])]^T] \Bigg. \Bigg \rangle \Bigg. \Bigg)^2.
\end{align}

For the sake of brevity let's denote $f_t^{(k)} := \grad f_t(\bs x_t^T\bs v[1:2], \ldots,\bs x_t^T \bs v[2d-1:2d])[k]$ for $k \in [d]$. Then we have
\begin{align}
    \grad \tilde f_t(\bs v)
    &= \left[ f_t^{(1)} \bs x_t^T, \ldots, f_t^{(d)} \bs x_t^T \right]^T.
\end{align}

Let 
\begin{align}
    A
    &= \Bigg \langle \Bigg. \grad f_t(\bs x_t^T\bs v[1:2], \ldots,\bs x_t^T \bs v[2d-1:2d]),\\
    &[\bs x_t^T(\bs w[1:2] - \bs v[1:2]),\ldots,\bs x_t^T(\bs w[2d-1:2d] - \bs v[2d-1:2d])]^T] \Bigg. \Bigg \rangle.
\end{align}

With this, we observe that,
\begin{align}
    A
    &= (\bs w - \bs v)^T \grad \tilde f_t(\bs v).
\end{align}

Thus, we obtain the affine invariance of exp-concavity as:
\begin{align}
    \tilde f_t(\bs w) 
    &\ge \tilde f_t(\bs v) +  (\bs w - \bs v)^T \grad \tilde f_t(\bs v) + \frac{\sigma}{2} \left( (\bs w - \bs v)^T \grad \tilde f_t(\bs v) \right)^2. \label{eq:affine-exp}
\end{align}

Note that the set $\mathcal K_t$ is convex. This can be seen as follows: if $\bs v, \bs w \in \c K_t$, then we have $|\bs x_t^T\bs v[2k-1:2k]| \le C$ and $|\bs x_t^T\bs w[2k-1:2k]| \le C$ for all $k \in [d]$. Now for any $t \in [0,1]$ let $\bs z = t \bs v + (1-t) \bs w$. Then we have for any $k \in [d]$ that
\begin{align}
    |\bs x_t^T\bs z[2k-1:2k]|
    &\le t |\bs x_t^T\bs v[2k-1:2k]| + (1-t) |\bs x_t^T\bs 2[2k-1:2k]|\\
    &\le C,
\end{align}
where the first inequality is via triangle inequality. Thus $\bs z \in \cK_t$ so the set $\cK_t$ is convex.

So by the properties of projection to convex sets (see for example, Lemma 16 in \citep{hazan2007logregret}) and the definition of the algorithm, we have that
\begin{align}
    \|\bs p_{t+1} - \bs w \|_{\bs A_t}^2
    &\le \|\bs w_{t+1} - \bs w \|_{\bs A_t}^2\\
    &= \|\bs p_t - \bs w \|_{\bs A_t}^2 + \bs g_t^T \bs A_t^{-1} \bs g_t - 2 \bs g_t^T (\bs p_t - \bs w).
\end{align}

Let $R_T(\bs w) := \sum_{t=1}^T \tilde f_t(\bs p_t) - \tilde f_t(\bs w)$. Since each $f_t$ is exp-concave, we have by Eq.\eqref{eq:affine-exp} and the previous inequality that
\begin{align}
    2R_T(\bs w)
    &\le \sum_{t=1}^T 2 \bs g_t^T (\bs p_t - \bs w) - \sigma (\bs g_t^T (\bs p_t - \bs w))^2\\
    &\le \sum_{t=1}^T \bs g_t^T \bs A_t^{-1} \bs g_t + \|\bs p_t - \bs w \|_{\bs A_t}^2 - \|\bs p_{t+1} - \bs w \|_{\bs A_t}^2 - \sigma (\bs g_t^T (\bs p_t - \bs w))^2\\
    &\le \| \bs w\|_{\bs A_0}^2 + \sum_{t=1}^T \bs g_t^T \bs A_t^{-1} \bs g_t + (\bs p_t - \bs w)^T (\bs A_t - \bs A_{t-1} - \sigma \bs g_t \bs g_t^T ) (\bs p_t - \bs w)\\
    &= \| \bs w\|_{\bs A_0}^2 + \sum_{t=1}^T \bs g_t^T \bs A_t^{-1} \bs g_t,
\end{align}
where the last line is by the definition of $\bs A_t$.

By using the arguments of Lemma 12 of \citep{hazan2007logregret} we have
\begin{align}
    \sum_{t=1}^T \bs g_t^T \bs A_t^{-1} \bs g_t
    &\le \frac{2d}{\sigma} \log \left(1 + \frac{\sigma TG^2}{d\epsilon} \right).
\end{align}

Thus overall we have,
\begin{align}
    R_T(\bs w)
    &\le  \frac{ \epsilon \| \bs w\|_2^2}{2}  + \frac{2d}{\sigma} \log \left(1 + \frac{\sigma TG^2}{d\epsilon} \right)
\end{align}

\end{proof}

\begin{corollary} \label{cor:t1-high} \citep{hazan2007adaptive}
Consider the FLH algorithm from \citep{hazan2007adaptive} with SIONS from Lemma \ref{lem:SON-stat} as the base experts with parameter $\epsilon = 2$ as described in Fig.\ref{fig:flh}. Consider an arbitrary interval $[a,b] \subseteq [n]$. 
Then the regret of FLH-SIONS within this interval is controlled as:
\begin{align}
    \sum_{j=a}^b f_j(\bs y_j) - \tilde f_j(\bs w)
    &\le  \| \bs w\|_2^2  + \frac{2d}{\sigma} \log \left(1 + \frac{\sigma n^3 G^2}{d\epsilon} \right) + \frac{4\log n}{\sigma},
\end{align}
where $\bs w \in \cap_{j=a}^b \cK_j$ and $\tilde f$ is as defined in Lemma \ref{lem:SON-stat}.
\end{corollary}
\begin{proof}
Since the loss functions $f_j$ are $\sigma$ exp-concave, by Lemma 3.3 in \citep{hazan2007adaptive} we have that
\begin{align}
    \sum_{j=a}^b f_j(\bs y_j)  \le \frac{4 \log n}{\sigma} + \sum_{j=a}^b f_j(E_a(j)).
\end{align}
Subtracting $\tilde f_j(\bs w)$ from both sides and using Lemma \ref{lem:SON-stat} now yields the result.
\end{proof}

\begin{corollary} \label{cor:binsh}
The number of bins $M:= | \cP|$ formed via a call to $\texttt{generateBins}(\bs u_{1:n})$ is at-most $O(n^{1/5}C_n^{2/5} \vee 1)$.
\end{corollary}
\begin{proof}
The proof is similar to that of Lemma \ref{lem:keypart}.
\end{proof}

\begin{lemma} \label{lem:bin-bound}
Let $[i_s,i_t] \in \cP$ where $\cP$ is the partition produced via the \code{generateBins} procedure. We have that the dynamic rgeret of FLH-SIONS within this bin controlled as
\begin{align}
    \sum_{j=i_s}^{i_t} f_j(\hat{\bs p}_j) - f_j(\bs u_j)
    &= \tilde O(d^2),
\end{align}
where $\hat{\bs p}_j \in \mathbb{R}^d$ are the predictions of the algorithm.
\end{lemma}
\begin{proof}
Consider a bin $[i_s,i_t]$. Let $\cQ = \texttt{refineSplit}([i_s,i_t])$. Define $\tilde f_j(\bs v) := \tilde f_j(\bs y_j^T \bs v)$ for $v \in \mathbb{R}^{2d}$.

Next, we proceed to construct the details of a regret decomposition within a bin $[i_s,i_t]$:
\begin{align}
    \sum_{j=i_s}^{i_t} f_j(\hat{\bs p}_j) - f_j(\bs u_j)
    &= \underbrace{\sum_{j=i_s}^{i_t} f_j(\hat{\bs p}_j) -  f_j(\bs X_j \bs \alpha_j)}_{T_1}
    + \underbrace{\sum_{j=i_s}^{i_t}  f_j(\bs X_j \bs \alpha_j) -  f_j(\bs X_j \bs \beta_j)}_{T_2} + \underbrace{\sum_{j=i_s}^{i_t}  f_j(\bs X_j \bs \beta_j) - f_j(\bs u_j)}_{T_3},\quad \label{eq:regdecomp-multi}
\end{align}
where we will construct appropriate $\bs y_j, \bs \alpha_j, \bs \beta_j \in \mathbb{R}^{2d}$ and $\bs X_j \in \mathbb{R}^{d \times 2d}$ in what follows. 

\begin{figure}[h!]
	\centering
	\fbox{
		\begin{minipage}{14 cm}
		    \code{AssignCo-variatesAndSlopes1}: Inputs- (1) offline optimal sequence (2) A bin $[a,b]$ (3) A coordinate $k \in [d]$
            \begin{enumerate}

                \item Let $\bs \beta_k$ be the least square fit coefficient computed with labels being $\bs u_a[k], \ldots, \bs u_b[k]$ and co-variates $\bs x_j := [1,j-a+1]^T$ so that the fitted value at time $j$ is given by $\hat {\bs u_j}[k] = \bs \beta ^T \bs x_j$.
                
                \item Set $\bs \beta_j[2k-1:2k] \leftarrow \beta_k$ for all $j \in [a,b]$.
                
                \item Set $\bs \alpha_k \leftarrow \beta_k$
                
                \item Set $\bs \alpha_j[2k-1:2k] \leftarrow \bs \alpha_k$ for all $j \in [a,b]$.
                
                \item Set $\bs y_j[2k-1:2k] = \bs x_j$ for all $j \in [a,b]$.
                
            \end{enumerate}
		\end{minipage}
	}
	\caption{\emph{\code{AssignCo-variatesAndSlopes1} used to set the parameters in the regret decomposition of Eq.\eqref{eq:regdecomp-multi} whenever the offline optimal is constant across the specified coordinate $k$ within the interval $[a,b]$. We use a 1-based indexing. i.e $\bs v[1]$ refers the first element of a vector $\bs v$.}}
	\label{fig:cov1}
\end{figure}

\begin{figure}[h!]
	\centering
	\fbox{
		\begin{minipage}{14 cm}
		    \code{AssignCo-variatesAndSlopes2}: Inputs- (1) offline optimal sequence (2) A bin $[a,b]$ (3) A coordinate $k \in [d]$
            \begin{enumerate}

                \item Let $\bs \beta_k$ be the least square fit coefficient computed with labels being $\bs u_a[k], \ldots, \bs u_b[k]$  and co-variates $\bs x_j := [1,j-a+1]^T$ so that the fitted value at time $j$ is given by $\hat {\bs u_j}[k] = \bs \beta ^T \bs x_j$.
                
                \item Set $\bs \beta_j[2k-1:2k] \leftarrow \beta_k$ for all $j \in [a,b]$.
                
                \item Set $\bs y_j[2k-1:2k] \leftarrow \bs x_j$ for all $j \in [a,b]$.
                
                \item Define $\bs A_k := \sum_{j=a}^b \bs x_j \bs x_j^T $, $\tilde f_j(\bs v) := \tilde f_j(\bs y_j^T \bs v)$ for some $v \in \mathbb{R}^{2d}$.
                
                \item Set $\bs \alpha_k \leftarrow \bs \beta_k - \bs A_k^{-1} \sum_{j=a}^b \grad \tilde f(\bs \beta_j)[2k-1:2k]$.
                
                \item Set $\bs \alpha_j[2k-1:2k] \leftarrow \bs \alpha_k$ for all $j \in [a,b]$.                
                
            \end{enumerate}
		\end{minipage}
	}
	\caption{\emph{\code{AssignCo-variatesAndSlopes2} used to set the parameters in the regret decomposition of Eq.\eqref{eq:regdecomp-multi} whenever the offline optimal may not be constant across the specified coordinate $k$ within the interval $[a,b]$. We use a 1-based indexing. i.e $\bs v[1]$ refers the first element of a vector $\bs v$.}}
	\label{fig:cov2}
\end{figure}

\noindent\textbf{\setword{(A1)}{Word:A1}:} Consider a coordinate $k \in [d]$ such that $\bs u[k]$ is not monotonic in $[i_s,i_t]$ and do not touch boundary 1. Let $[i_s,i_t] = [i_s,a-1] \cup [a,b] \cup [b+1,c] \cup [c+1,i_t]$ such that $\bs u[k]$ is constant in bins $[i_s,a-1]$ and $[c+1,i_t]$. Further we consider the case where $s_{a-1} = 1$ and $s_b = -1$ with $\bs u[k]$ non-decreasing within $[b+1,c]$. (Note that this can be guaranteed by picking $b$ as the last point with $\bs u_{b+1}[k] - \bs u_b[k] > \bs u_{b+2}[k] - \bs u_{b+1}[k]$.) The alternate case where $s_{a-1} = -1$ and $s_b = 1$ with $\bs u[k]$ non-increasing within $[b+1,c]$ can be handled similarly. All the arguments we explain for the case of offline optimal touching the boundary $-1$ can be mirrored to handle the case where the offline optimal touches the boundary 1. (The offline optimal can't touch both boundaries simultaneously along a coordinate, see Lemma \ref{lem:dual-bounded})

We will use 1-based indexing. (i.e $\bs v[1]$ denotes the first element of a vector). For each $k \in [d]$:
\begin{itemize}

    \item Call $\texttt{AssignCo-variatesAndSlopes1}(\bs u_{1:n},[i_s,a-1],k)$.
    
    \item Call $\texttt{AssignCo-variatesAndSlopes2}(\bs u_{1:n},[a,b],k)$.
    
    \item Let $[b+1,t_1-1],[t_1,t_2],[t_2+1,c]$ be the bins returned by a call to $\texttt{splitMonotonic}(\bs u_{1:n},[b+1,c],k)$.
    
    \item Call $\texttt{AssignCo-variatesAndSlopes1}(\bs u_{1:n},[b+1,t_1-1],k)$.
    
    \item Call $\texttt{AssignCo-variatesAndSlopes2}(\bs u_{1:n},[t_1,t_2],k)$.
    
    \item Call $\texttt{AssignCo-variatesAndSlopes1}(\bs u_{1:n},[t_2+1,c],k)$.    
    \item Call $\texttt{AssignCo-variatesAndSlopes1}(\bs u_{1:n},[c+1,i_t],k)$.
    
\end{itemize}

For a vector $\bs y$ we treat $\bs y[m:n] = [\bs y[m],\ldots,\bs y[n]]^T$.
Define $\bs X_j \in \mathbb{R}^{d \times 2d}$ as

\begin{gather}
    \bs X_j^T
    =
    \begin{bmatrix}
    \bs y_j[1:2] & \bs 0 & \dots & \bs 0\\
    \bs 0 & \bs y_j[3:4] & \dots & \bs 0\\
    \vdots & \ddots & & \vdots\\
    \bs 0 & \dots & & \bs y_j[2d-1:2d]
    \end{bmatrix}, \label{eq:block-cov}
\end{gather} 
where $\bs 0 = [0,0]^T$ and $\bs y_j$ is set according to various calls of $\texttt{AssignCo-variatesAndSlopes1}$ and\\ $\texttt{AssignCo-variatesAndSlopes2}$ as done previously.

We proceed to bound $T_2 + T_3$ in Eq.\eqref{eq:regdecomp-multi}. First notice that due to Taylor's theorem,
\begin{align}
    \tilde f_j(\bs \alpha_j) - \tilde f_j(\bs \beta_j)
    &= \langle \grad \tilde f_j(\bs \beta_j), \bs \alpha_j - \bs \beta_j \rangle + \frac{1}{2} \| \bs \alpha_j - \bs \beta_j \|_{\grad^2 \tilde f_j(\bs v)}^2, 
\end{align}
where $\bs v = t \bs \alpha_j + (1-t) \bs \beta_j$ for some $t \in [0,1]$. Now we use Lemma \ref{lem:diagonal} to obtain,
\begin{align}
    \tilde f_j(\bs \alpha_j) - \tilde f_j(\bs \beta_j)
    &\le \langle \grad \tilde f_j(\bs \beta_j), \bs \alpha_j - \bs \beta_j \rangle + \frac{1}{2} \sum_{k'=1}^d \|\bs \alpha_j[2k'-1:2k'] -  \bs \beta_j[2k'-1:2k']\|_{\bs y_j[2k'-1:2k'] \bs y_j[2k'-1:2k']^T}^2\\
    &= \sum_{k'=1}^d \langle \grad f_j(\bs X_j \bs \beta_j)[k'] \bs y_j[2k'-1:2k'], \bs \alpha_j[2k'-1:2k'] -  \bs \beta_j[2k'-1:2k'] \rangle\\
    &\quad+ \frac{1}{2} \sum_{k'=1}^d \|\bs \alpha_j[2k'-1:2k'] -  \bs \beta_j[2k'-1:2k']\|_{\bs y_j[2k'-1:2k'] \bs y_j[2k'-1:2k']^T}^2 
    \label{eq:t2diag}
\end{align}

Further, due to gradient Lipschitzness,
\begin{align}
    \tilde f_j(\bs \beta_j) - f_j(\bs u_j)
    &= f_j(\bs X_j \bs \beta_j) - f_j(\bs u_j)\\
    &\le \langle \grad f_j(\bs u_j), \bs X_j \bs \beta_j - \bs u_j \rangle + \frac{1}{2} \| \bs X_j \bs \beta_j - \bs u_j \|_2^2\\
    &= \sum_{k'=1}^d \grad f_j(\bs u_j)[k'] \cdot \left( \bs \beta_j[2k'-1:2k']^T \bs y_j[2k'-1:2k'] - \bs u_j[k'] \right)\\
    &\quad + \sum_{k'=1}^d \frac{1}{2} \| \bs \beta_j[2k'-1:2k']^T \bs y_j[2k'-1:2k'] - \bs u_j[k'] \|_2^2 \label{eq:t3diag}
\end{align}

Looking at Eq.\eqref{eq:t2diag} and \eqref{eq:t3diag}, we see that they decompose across each coordinate $k'$. So we can bound $T_2 + T_3$ in any bin $[i_s,i_t]$ coordinate wise:

\begin{align}
    T_2 + T_3
    &= \sum_{k'=1}^d \sum_{j=i_s}^{i_t} \langle \grad f_j(\bs X_j \bs \beta_j)[k'] \bs y_j[2k'-1:2k'], \bs \alpha_j[2k'-1:2k'] -  \bs \beta_j[2k'-1:2k'] \rangle\\
    &\quad + \frac{1}{2} \sum_{k'=1}^d \|\bs \alpha_j[2k'-1:2k'] -  \bs \beta_j[2k'-1:2k']\|_{\bs y_j[2k'-1:2k'] \bs y_j[2k'-1:2k']^T}^2\\
    &\quad + \grad f_j(\bs u_j)[k'] \cdot \left( \bs \beta_j[2k'-1:2k']^T \bs y_j[2k'-1:2k'] - \bs u_j[k'] \right)\\
    &\quad + \frac{1}{2} \| \bs \beta_j[2k'-1:2k']^T \bs y_j[2k'-1:2k'] - \bs u_j[k'] \|_2^2\\
    &:= \sum_{k'=1}^d \sum_{j=i_s}^{i_t} t_{2,j,k'} + t_{3,j,k'}, \label{eq:t2}
\end{align}
where in the last line we define:
\begin{align}
    t_{2,j,k'}
    &:= \langle \grad f_j(\bs X_j \bs \beta_j)[k'] \bs y_j[2k'-1:2k'], \bs \alpha_j[2k'-1:2k'] -  \bs \beta_j[2k'-1:2k'] \rangle\\
    &\quad + \frac{1}{2}  \|\bs \alpha_j[2k'-1:2k'] -  \bs \beta_j[2k'-1:2k']\|_{\bs y_j[2k'-1:2k'] \bs y_j[2k'-1:2k']^T}^2,\label{eq:t3}
\end{align}

and
\begin{align}
    t_{3,j,k'}
    &:= \grad f_j(\bs u_j)[k'] \cdot \left( \bs \beta_j[2k'-1:2k']^T \bs y_j[2k'-1:2k'] - \bs u_j[k'] \right)\\
    &\quad + \frac{1}{2} \| \bs \beta_j[2k'-1:2k']^T \bs y_j[2k'-1:2k'] - \bs u_j[k'] \|_2^2.
\end{align}

Next, we proceed to bound $\sum_{j=i_s}^{i_t} t_{2,j,k} +  t_{3,j,k}$ for the coordinate $k$ with a structure as mentioned in Paragraph \ref{Word:A1}.

Recall that $[i_s,i_t] = [i_s,a-1] \cup [a,b] \cup [b+1,t_1-1] \cup [t_1,t_2] \cup [t_2+1,c] \cup [c+1,i_t]$. So we will consider each of these sub-bins separately.

For bin $[i_s,a-1]$ we have $\bs \alpha_j[2k-1:2k] = \bs \beta_j[2k-1:2k]$ and $\bs \beta_j[2k-1:2k]^T \bs y_j[2k-1:2k] = \bs u_j[k]$. So we trivially have
\begin{align}
 \sum_{j=i_s}^{a-1} t_{2,j,k} +  t_{3,j,k} = 0   . \label{eq:bin1}
\end{align}

Next, we focus on the bin $[a,b]$. We note that by construction, $\bs \alpha_j[2k-1:2k]$ and $\bs \beta_j[2k-1:2k]$ are fixed for all $j \in [a,b]$. Let's denote these fixed values by $\bs \alpha_k$ and $\bs \beta_k$ respectively. For the sake of brevity let's denote $\bs x_j := \bs y_j[2k-1:2k]$ and $\bs A_k = \sum_{j=a}^b \bs x_j \bs x_j^T$. We have the relation,
\begin{align}
    \bs \alpha_k 
    &= \bs \beta_k - \bs A_k^{-1} \sum_{j=a}^b \grad \tilde f(\bs \beta_j)[2k-1:2k]\\
    &= \bs \beta_k - \bs A_k^{-1} \sum_{j=a}^b \grad f_j(\bs X_j \bs \beta_j)[k] \bs x_j. \label{eq:update}
\end{align}

By the new compact notations, we have
\begin{align}
    t_{2,j,k}
    &= \langle \grad f_j(\bs X_j \bs \beta_j)[k] \bs x_j, \bs \alpha_k -  \bs \beta_k \rangle
     + \frac{1}{2} \|\bs \alpha_k -  \bs \beta_k\|_{\bs x_j \bs x_j^T}^2,
\end{align}

and
\begin{align}
    t_{3,j,k}
    &= \grad f_j(\bs u_j)[k] \cdot \left( \bs \beta_k^T \bs x_j - \bs u_j[k] \right)
     + \frac{1}{2} \| \bs \beta_k^T \bs x_j - \bs u_j[k] \|_2^2.     
\end{align}

From Eq.\eqref{eq:update} we have,
\begin{align}
    \sum_{j=a}^b t_{2,j,k}
    &= - \left \|    \sum_{j=a}^b \grad f_j(\bs X_j \bs \beta_j)[k] \bs x_j \right \|_{\bs A_k^{-1}}^2 + \frac{1}{2} \left \| \bs A_k^{-1} \sum_{j=a}^b \grad f_j(\bs X_j \bs \beta_j)[k] \bs x_j \right \|_{\bs A_k}^2\\
    &= -\frac{1}{2} \left \|   \sum_{j=a}^b \grad f_j(\bs X_j \bs \beta_j)[k] \bs x_j  \right\|_{\bs A_k^{-1}}^2\\
    &\le -\frac{1}{2} \left \|   \sum_{j=a}^b \grad f_j(\bs u_j)[k] \bs x_j  \right\|_{\bs A_k^{-1}}^2 + 2 \langle \bs A_k^{-1} \sum_{j=a}^b \grad f_j(\bs u_j)[k] \bs x_j, \sum_{j=a}^b \bs x_j \left (\grad f_j(\bs X_j \bs \beta_j)[k] - \grad f_j(\bs u_j)[k] \right)\rangle.
\end{align}

Now we define $\bs g_k = \lamda [-\bs s_{a-2}[k] + \bs s_{a-1}[k] + \bs s_{b-1}[k] - \bs s_b[k], -\bs s_{a-2}[k] + (\ell+1)\bs s_{b-1}[k] - \ell \bs s_b[k]]^T$ and $\bs h_k = [\bs \Gamma[k], \tilde {\bs \Gamma}[k]]^T$ where $\bs \Gamma = \sum_{j=a}^b \bs \gamma_j^- - \bs \gamma_j^+$ and $\tilde {\bs \Gamma} = \sum_{j=a}^b  j'(\bs \gamma_j^- - \bs \gamma_j^+)$ where $j' = j-a+1$ so that $\sum_{j=a}^b \grad f_j(\bs u_j)[k] \bs x_j = \bs g_k + \bs h_k$ via the KKT conditions in Lemma \ref{lem:kkt-high}. 

With these, we can bound:
\begin{align}
    \sum_{j=a}^b 2\cdot t_{2,j,k}
    &\le -\| \bs g_k\|_{\bs A_k^{-1}}^2 - \| \bs h_k \|_{\bs A_k^{-1}}^2 - 2 < \bs g_k, \bs A_k^{-1} \bs h_k \rangle \\
    &+ 2 \langle \bs A_k^{-1} (\bs g_k + \bs h_k), \sum_{j=a}^b \bs x_j \left (\grad f_j(\bs X_j \bs \beta_j)[k] - \grad f_j(\bs u_j)[k] \right) \rangle \label{eq:h1}
\end{align}

Proceeding similarly to Eq.\eqref{eq:a9} and \eqref{eq:a10} by gradient Lipschitzness we obtain,
\begin{align}
    \sum_{j=a}^b \grad f_j(\bs X_j \bs \beta_j)[k] - \grad f_j(\bs u_j)[k]
    &\le \sum_{j=a}^b \|\bs X_j \bs \beta_j -  \bs u_j\|_1\\
    &\le 20\ell^2 \ell^{-3/2},
\end{align}
where in the last line we used  Lemma \ref{lem:residual} coordinate-wise and the fact that $\| D^2 \bs u_{a:b}\|_1 \le \ell^{-3/2}$.

Similarly,
\begin{align}
    \sum_{j=a}^b j'( \grad f_j(\bs X_j \bs \beta_j)[k]) - \grad f_j(\bs u_j)[k]
    &\le \sum_{j=a}^b j' \|\bs X_j \bs \beta_j -  \bs u_j\|_1\\
    &\le 20\ell^3 \ell^{-3/2}.
\end{align}

Hence by KKT conditions in Lemma \ref{lem:kkt-high}, we can further bound
\begin{align}
    \langle \bs A_k^{-1} \bs g_k, \sum_{j=a}^b \bs x_j \left (\grad f_j(\bs X_j \bs \beta_j)[k] - \grad f_j(\bs u_j)[k] \right) \rangle
    &\le \frac{40\lamda \ell^{-1/2} }{(\ell-1)} \Bigg | \Bigg. (2-2\ell)\bs s_{a-2}[k] + (2\ell+1)\bs s_{a-1}[k]\\
    &\quad- (\ell+2)\bs s_{b-1}[k] + (\ell-1)\bs s_b[k] \Bigg. \Bigg | \\
    &\quad+ \frac{40\lamda\ell^{1/2}}{(\ell-1)} \Bigg | \Bigg. \frac{3(\ell-1)}{\ell+1}\bs s_{a-2}[k] - 3\bs s_{a-1}[k]\\
    &\quad+ 3 \bs s_{b-1}[k] - \frac{3(\ell-1)}{\ell+1}\bs s_b[k]  \Bigg. \label{eq:h2} \Bigg|,
\end{align}

and
\begin{align}
    \langle \bs A_k^{-1} \bs h_k, \sum_{j=a}^b \bs x_j \left (\grad f_j(\bs X_j \bs \beta_j)[k] - \grad f_j(\bs u_j)[k] \right) \rangle
    &\le \frac{40\ell^{-1/2}}{(\ell-1)} |(2\ell+1) {\bs \Gamma}[k] - 3 \tilde {\bs \Gamma}[k]|\\
    &\quad+ \frac{40 \ell^{1/2}}{(\ell-1)} \left| \frac{6 \tilde {\bs \Gamma}[k]}{\ell+1} - 3{\bs \Gamma}[k] \right|. \label{eq:h3}
\end{align}

We observe that Eq.\eqref{eq:h1},\eqref{eq:h2},\eqref{eq:h3} are semantically same as Eq.\eqref{eq:t2}, \eqref{eq:a9} and \eqref{eq:a10} respectively in the 1D case.

Next, we proceed to setup a similar observation for bounding $\sum_{j=a}^b t_{3,j,k}$. From KKT conditions in Lemma \ref{lem:kkt-high} and proceeding similar to the arguments in Lemma \ref{lem:t3} we get,
\begin{align}
    \sum_{j=a}^b \grad f_j(\bs u_j)[k] \cdot \left( \bs \beta_j[2k-1:2k]^T \bs y_j[2k-1:2k] - \bs u_j[k] \right)
    &= \sum_{j=a}^{b} \lamda \Bigg( \Bigg. \left((\bs s_{j-1}[k] - \bs s_{j-2}[k]) - (\bs s_{j}[k] - \bs s_{j-1}[k]) \right) \times\\
    &\qquad  ((j-a+1) \bs M_j[k] + \bs C_j[k]) \Bigg. \Bigg)\\
    &\quad + \sum_{j=a}^{b} (\bs \gamma_j^-[k] - \bs \gamma_j^+[k])\times\\
    &\qquad(\bs \beta_j[2k-1:2k]^T \bs y_j[2k-1:2k] - \bs u_j[k])\\
    &\le \sum_{j=a}^{b} \lamda \Bigg( \Bigg. \left((\bs s_{j-1}[k] - \bs s_{j-2}[k]) - (\bs s_{j}[k] - \bs s_{j-1}[k]) \right) \times\\
    &\qquad  ((j-a+1) \bs M_j[k] + \bs C_j[k]) \Bigg. \Bigg)\\
    &\qquad + 20 \ell^{-1/2} \sum_{j=a}^{b} \left|\bs \gamma_j^-[k] - \bs \gamma_j^+[k] \right|,\label{eq:h4}
\end{align}
where similar to Lemma \ref{lem:t3}, we represent $\bs \beta_j[2k-1:2k]^T \bs y_j[2k-1:2k] - \bs u_j[k] = (j-a+1) \bs M_j[k]+ \bs C_j[k]$ with $\bs M_a[k] = \bs M_{a+1}[k]$, $\bs C_{a}[k] = \bs C_{a+1}[k]$, $\bs M_{b}[k] = \bs M_{b-1}[k]$ and $\bs C_{b}[k] = \bs C_{b-1}[k]$. The last line is obtained due to Lemma \ref{lem:residual}.

Further, by using Lemma \ref{lem:residual} we obtain,
\begin{align}
    \sum_{k'=1}^d \sum_{j=a}^b  \frac{1}{2} \| \bs \beta_j[2k'-1:2k']^T \bs y_j[2k'-1:2k'] - \bs u_j[k'] \|_2^2
    &\le 200.
\end{align}

Combining the last two inequalities yields,
\begin{align}
    \sum_{j=a}^b t_{3,j,k}
    &\le 200 
    + \sum_{j=a}^{b} \lamda \Bigg( \Bigg. \left((\bs s_{j-1}[k] - \bs s_{j-2}[k]) - (\bs s_{j}[k] - \bs s_{j-1}[k]) \right) ((j-a+1) \bs M_j[k] + \bs C_j[k]) \Bigg. \Bigg)\\
    &\quad + 20 \ell^{-1/2} \sum_{j=a}^{b} \left |\bs \gamma_j^-[k] - \bs \gamma_j^+[k] \right|. \label{eq:h5}
\end{align}

We observe that the last inequality is semantically similar to Eq.\eqref{eq:t3lemma} for 1D case. Recall that Eq.\eqref{eq:h1},\eqref{eq:h2},\eqref{eq:h3} are also semantically same as Eq.\eqref{eq:t2}, \eqref{eq:a3} and \eqref{eq:a6} respectively in the 1D case.

Hence we can proceed to bound 
\begin{align}
  \sum_{j=a}^b t_{2,j,k} + t_{3,j,k} = O(1),  \label{eq:bin2}
\end{align}
using the same arguments as in Lemma \ref{lem:non-mono}.

Observe that by construction, the slopes across coordinate $k$ are constant in the bins $[b+1,t_1-1],[t_2+1,c]$ and $[c+1,i_t]$. So by using similar arguments used for handling the bin $[i_s,a-1]$ we obtain,
\begin{align}
    \sum_{j \in \mathcal I} t_{2,j,k} + t_{3,j,k} = 0,\label{eq:bin3}
\end{align}
where $\mathcal I \in \{[b+1,t_1-1],[t_2+1,c], [c+1,i_t]\}$.

By appealing to our reduction to 1D case facilitated by Eq.\eqref{eq:h1} and \eqref{eq:h5} and using similar arguments used to handle the monotonic slopes case as in Lemma \ref{lem:mono} we obtain,
\begin{align}
    \sum_{j=t_1}^{t_2} t_{2,j,k} + t_{3,j,k} = O(1).\label{eq:bin4}
\end{align}

So far we have discussed bounding $\sum_{j=i_s}^{i_t} t_{2,j,k} + t_{3,j,k}$ for a bin with structure across coordinate $k$ as described in Paragraph \ref{Word:A1}. We remark that if the slopes across a coordinate $k$ assumes a monotonic structure across $[i_s,i_t]$, we can handle it in the same way as we handled the sub-bin $[t_1,t_2]$ above.

We pause to remark that Eq.\eqref{eq:bin1},\eqref{eq:bin2},\eqref{eq:bin3} and \eqref{eq:bin4} together gives a way to bound to $\sum_{j=i_s}^{i_t} t_{2,j,k'} + t_{3,j,k'}$ across any coordinate $k'$ as we comprehensively considered all the  possible structures across a coordinate $k'$. (The alternate cases where where $s_{a-1} = -1$ and $s_b = 1$ with $\bs u[k']$ non-increasing within $[b+1,c]$ can be handled similarly to the case described in Paragraph \ref{Word:A1}. Finally the case where the offline optimal touches boundary 1 instead of $-1$ can be handled using similar arguments.)

Thus overall we obtain that for any bin $[i_s,i_t] \in \cP$ we have:
\begin{align}
    T_2 + T_3
    &\le \sum_{k'=1}^d \sum_{j=i_s}^{i_t} {t_2} t_{2,j,k'} + t_{3,j,k'}\\
    &= O(d),
\end{align}
where $T_2$ and $T_3$ are as defined in Eq.\eqref{eq:regdecomp-multi}.

Next, we proceed to control $T_1$. Recall that
\begin{align}
    T_1
    &=
    \sum_{j=i_s}^{i_t} f_j(\bs p_j) - f_j(\bs X_j \bs \alpha_j).
\end{align}

Let's revisit bin $[i_s,i_t]$ with structure as described in Paragraph \ref{Word:A1} across coordinate $k$. First we consider the bin $[a,b]$. Through the call to $\texttt{AssignCo-variatesAndSlopes2}(\bs u_{1:n}, [a,b],k)$ we set $\bs \alpha_k$. Further $\bs \alpha_j[2k-1:2k] = \bs \alpha_k$ for all $j \in [a,b]$. By using similar arguments as in the proof of Lemma \ref{lem:t1} which lead to Eq.\eqref{eq:bounded-pred}, we have that $|\bs y_j[2k-1:2k]^T \bs \alpha_j| \le 20$. For other bins such as $[i_s,a-1],[b+1,t_1-1],[t_2+1,c],[c+1,i_t]$ where the slope of the offline optimal across coordinate $k$ remains constant, we set $\alpha_j[2k-1:2k]$ for $j \in \cI$ with $\cI \in \{ [i_s,a-1],[b+1,t_1-1],[t_2+1,c],[c+1,i_t]\}$ to be a constant value obtained as the least square fit coefficients with co-variates $\bs y_j[2k-1:2k]$ and labels set appropriately via the call to \code{AssignCo-variatesAndSlopes1}. Hence in this case also we have $|\bs y_j^T[2k-1:2k] \bs \alpha_j[2k-1:2k]| \le 10$ via the arguments in Lemma \ref{lem:t1}.

For the alternate cases (i) where $s_{a-1} = -1$ and $s_b = 1$ with $\bs u[k']$ non-increasing within $[b+1,c]$ as described in Paragraph \ref{Word:A1} (ii) case where the offline optimal touches boundary 1 instead of -1
(iii) The offline optimal across coordinate $k$ is non-decreasing within $[i_s,i_t]$ and (iv) The offline optimal across coordinate $k$ is non-increasing within $[i_s,i_t]$. In all these cases we can set the quantities $\bs \alpha_j[2k-1:2k],\bs y_j[2k-1:2k]$ by similar calls to \code{AssignCo-variatesAndSlopes1} or \code{AssignCo-variatesAndSlopes2} such that $\bs y_j[2k-1:2k]^T \bs \alpha_j[2k-1:2k] \le 20$ for all $j \in [i_s,i_t]$. For example, for case (iii) we can resort to similar arguments used for handling sub-bin $[t_1,t_2]$ which is again similar to how we handled the bin $[a,b]$. (see Paragraph \ref{Word:A1}).

Further, even-though we create at-most 6 sub-bins across each coordinate for an interval $[i_s,i_t] \in \cP$ (see Paragraph \ref{Word:A1} and the sequence of calls beneath), doing so for each coordinate can result in at-most $6d$ partitions of $\bs u_{i_s:i_t}$ overall. However, if we consider any sub-bin $[p,q]$ of this partition, we have that $\bs \alpha_j[2k-1:2k]$ is fixed and $\bs \beta_j[2k-1:2k]$ is fixed for all $j \in [p,q]$ across any coordinate $k \in [d]$ and $y_j[2k-1:2k][2]$ is monotonically increasing wrt $j \in [p,q]$ for all coordinates $k \in [d]$. Now suppose that $k' \in [d]$ is such that $\bs y_j[2k'-1:2k'][1] \le \bs y_j[2k-1:2k][1]$ for all $k \neq k'$ and for all $j \in [p,q]$. With a change of variables we have that $\tilde {\bs \alpha}_j[2k-1:2k]^T \bs y_j[2k'-1:2k']  = \bs \alpha_j[2k-1:2k]^T \bs y_j[2k-1:2k]$ by setting $\tilde {\bs \alpha}_j[2k-1:2k][2] = {\bs \alpha}_j[2k-1:2k][2]$ and $\tilde {\bs \alpha}_j[1] = {\bs \alpha}_j[1] + (\bs y_j[2k-1:2k][2] - \bs y_j[2k'-1:2k'][2]) \bs \alpha_j[2k-1:2k][2]$ for $k \neq k'$ within the bin $[p,q]$. Since $(\bs y_j[2k-1:2k][2] - \bs y_j[2k'-1:2k'][2]) \le \bs y_j[2k-1:2k][2]$ by Eq.\eqref{eq:c2} we have that
\begin{align}
    |(\bs y_j[2k-1:2k][2] - \bs y_j[2k'-1:2k'][2]) \bs \alpha_j[2k-1:2k][2]| \le 6. \label{eq:z1}
\end{align}

Further we have from Eq.\eqref{eq:bounded-pred} that $|\bs \alpha_j[2k-1:2k][2] \bs y_j[2k'-1:2k'][2] + \bs \alpha_j[2k-1:2k][1]  | \le 20$ due to the fact that $\bs \alpha_j[2k-1:2k]$ remains fixed from a time point $j^* \le p$ such that $\bs y_{j^*}[2k-1:2k] = [1,1]^T$. Further we have that
\begin{align}
    \|\tilde{\bs \alpha}_j[2k-1:2k]  \|_2^2
    &=  (\bs \alpha_j[2k-1:2k][2])^2\\
    &+ (\bs \alpha_j[2k-1:2k][1] + (\bs y_j[2k-1:2k][2] - \bs y_j[2k'-1:2k'][2]) \bs \alpha_j[2k-1:2k][2])^2\\
    &\le 2 \left( \bs \alpha_j[2k-1:2k][2] + \bs \alpha_j[2k-1:2k][1] \right)^2\\
    &+ 2 \left( (\bs y_j[2k-1:2k][2] - \bs y_j[2k'-1:2k'][2])\bs \alpha_j[2k-1:2k][2] \right)^2\\
    &\le  584, \label{eq:alphanorm}
\end{align}
where the last line is due to Eq.\eqref{eq:alpha-bound} and \eqref{eq:z1}.

Let's represent $\bs \mu \in \mathbb{R}^{2d}$ such that $\bs \mu[2k'-1:2k'] = \bs \alpha [2k'-1:2k']$ and $\bs \mu[2k-1:2k] = \bs \alpha [2k-1:2k]$ for all other $k \in [d]$.

Thus within the sub-bin $[p,q]$, we have that $|\bs \mu^T[2k-1:2k] \bs y_j[2k'-1:2k'] | \le 20$ for all $k \in [d]$. Further, due to Eq.\eqref{eq:alphanorm} we have that $\| \bs \mu\|_2^2 \le 584d$. Hence we can use a base expert that starts at time $p$ which gives the co-variate $\bs y_j[2k'-1:2k']$ to all coordinates where $j \in [p,q]$. Note that the sub-bin $[p,q]$ must have been resulted via a splitting across coordinate $k'$ at time $p$. So by the calls to \code{AssignCo-variatesAndSlopes1} or \code{AssignCo-variatesAndSlopes2} we set $\bs y_p[2k'-1:2k'] = [1,1]^T$. Thus there exists a base expert in FLH-SIONS (Fig.\ref{fig:flh}) that provides  the co-variate $\bs y_j[2k'-1:2k']$ to all coordinates where $j \in [p,q]$.

This expert will have a regret of $\tilde O(d)$ against $\bs \mu$ via Lemma \ref{lem:SON-stat}. By using Strong Adaptivity from Corollary \ref{cor:t1-high} (set $\bs w = \bs \mu$ there and recall that $\| \bs \mu\|_2^2 \le 584d$) and adding the regret across all $6d$ sub-bins of $[i_s,i_t]$ lead to an $\tilde O(d^2)$ on $T_1$ in Eq.\eqref{eq:regdecomp-multi}. Thus for any bin in $\cP$ produced by generate bins procedure, we have its dynamic regret bounded by $\tilde O(d^2)$. 

\end{proof}

\begin{proof}[\textbf{Proof of Theorem} \ref{thm:main-d}] The proof is now complete by adding the $\tilde O(d^2)$ dynamic regret bound across all $O(n^{1/5}C_n^{2/5} \vee 1)$ bins in $\cP$ from Corollary \ref{cor:binsh}.

\end{proof}

The proof of Lemma \ref{lem:diagonal-main} is same as that of the lemma below, albeit with slightly different notations for $\bs X_j$.
\begin{lemma} \label{lem:diagonal}
Let $\bs X_j$ be as defined in Eq.\eqref{eq:block-cov}. Let $\tilde f_j(\bs v) = f_j(\bs X_j \bs v)$ for some $\bs v \in \mathbb{R}^{2d}$ and let $\bs \Sigma:= \bs X_j^T \bs X_j \in \mathbb{R}^{2d \times 2d}$. We have,
\begin{align}
    \grad^2 \tilde f_j(\bs v) \preccurlyeq \bs \Sigma
\end{align}
\end{lemma}
\begin{proof}

We have,
\begin{align}
    \tilde f_j(\bs v)
    &= f_j \left(\langle \bs y_j[1:2], \bs v[1:2] \rangle, \ldots, \langle \bs y_j[2d-1:2d], \bs v[2d-1:2d] \rangle \right).
\end{align}

Let
\begin{align}
    f''_{jk}
    &:= \grad^2 f_j \left(\langle \bs y_j[1:2], \bs v[1:2] \rangle, \ldots, \langle \bs y_j[2d-1:2d], \bs y_j[2d-1:2d] \rangle \right)[j][k],
\end{align}
be the Hessian of $f$ evaluated at the vector $\left[\langle \bs y[1:2], \bs v[1:2] \rangle, \ldots, \langle \bs y[2d-1:2d], \bs v[2d-1:2d] \rangle \right]^T \in \mathbb{R}^d$.

By straightforward calculations, we obtain
\begin{gather}
    \grad^2 \tilde f_j(\bs v)
    =
    \begin{bmatrix}
    f''_{11} \bs y_j[1:2] \bs y_j[1:2]^T & \dots & f''_{1d} \bs y_j[1:2] \bs y_j[2d-1:2d]^T\\
    \vdots & \ddots & \vdots\\
    f''_{d1} \bs y_j[2d-1:2d] \bs y_j[1:2]^T  & \dots & \bs f''_{dd} \bs y_j[2d-1:2d] \bs y_j[2d-1:2d]^T
    \end{bmatrix}, 
\end{gather} \label{eq:block-hess}

Let $\bs I \in \mathbb{R}^{d \times d}$ be the identity matrix and $\bs 1 \in \mathbb{R}^{2 \times 2}$ be the matrix of all ones. Further let's denote $\bs b := \left[\langle \bs y[1:2], \bs v[1:2] \rangle, \ldots, \langle \bs y[2d-1:2d], \bs v[2d-1:2d] \rangle \right]^T$ We can succinctly write:
\begin{align}
    \bs \Sigma-\grad^2 \tilde f_j(\bs v)
    &=  \left( \left(\bs I - \grad^2 f(\bs b) \right) \otimes \bs 1 \right) \circ \bs y_j \bs y_j^T,
\end{align}
where $\bs \otimes$ denotes the Kronecker product and $\circ$ denotes the Hadamard product.

Recall that the loss functions $f_j$ are 1-gradient Lipschitz. So we have $\left(\bs I - \grad^2 f(\bs b) \right)$ is Positive Semi Definite (PSD). The matrices $\bs 1$ and $\bs y_j \bs y_j^T$ are also PSD. Since both Kronecker and Hadamard products preserves positive semidefiniteness, we have $\grad^2 \tilde f_j(\bs v) \preccurlyeq \bs \Sigma$ which proves the lemma.

\end{proof}


\begin{proposition} \label{prop:embed}
Consider the sequence class $\mathcal{TV}^{1}(C_n)$ as per Eq.\eqref{eq:tvk}. Under Assumption A1 (see Section \ref{sec:main}) we have that $\mathcal{TV}^{(1)}(C_n) \subseteq \mathcal{TV}^{(0)}(2C_n + 20d)$.
\end{proposition}
\begin{proof}
We start by considering a 1D setting. Consider a sequence $w_{1:n} \in \mathcal{TV}^{(1)}(C_n)$. We can represent it as sum (point-wise) of two sequences as
\begin{align}
    w_{1:n} = p_{1:n} + q_{1:n}, \label{eq:seqsum}
\end{align}
where $q_{1:n} = \bs \beta^T \bs x_t$ where $\bs x_t = [1,t]^T$ and $\bs \beta$ is the least square fit coefficnts computed by using covariates $\bs x_t$ and labels $w_t$, $t \in [n]$. Here the $p_{1:n}$ is the residual sequence obtained by subtracting the least square fit sequence from the true sequence.

Following the terminology in Lemma \ref{lem:t3}, we can represent $p_t = tM_t+C_t$. Further, due to Eq.\eqref{eq:incpt} (with $a=1$) we have that $p_{t+1} - p_t = M_{t+1}$.

Applying triangle inequality to Eq.\eqref{eq:seqsum} we have
\begin{align}
    \|D w_{1:n}\|_1
    &\le  \|D p_{1:n}\|_1 +  \|D q_{1:n}\|_1.
\end{align}

Further,
\begin{align}
     \|D p_{1:n}\|_1
     &= \sum_{t=2}^n |M_t|\\
     &= \sum_{t=2}^n \left |M_1 + \sum_{j=1}^{t-1} M_{j+1} - M_{j} \right|\\
     &\le \sum_{t=2}^n |M_1| + D^2 \| p_{1:n}\|_1\\
     &=_{(a)} n |M_1| + n D^2 \| w_{1:n}\|_1\\
     &\le_{(b)} 2n D^2 \| w_{1:n}\|_1,
\end{align}
where in line (a) we used the fact that $\|D^2 p_{1:n} \|_1 = \|D^2 w_{1:n} \|_1$ as subtracting a linear sequence doesn't affect the TV1 distance. In line (b) we applied $|M_1| \le \|D^2 w_{1:n} \|_1$ as shown in Lemma \ref{lem:t3}.

It remains to bound $\|Dq_{1:n} \|_1$. For this we note that $\| q_t\| \le 10$ for all $t \in [n]$ due to Eq.\eqref{eq:boundedlstsq}. Since $q_{1:n}$ is a monotonic sequence we have that its variation $\|Dq_{1:n} \|_1 \le 20$.

Thus overall we obtain that
\begin{align}
    \|D w_{1:n}\|_1
    &\le   2n D^2 \| w_{1:n}\|_1 +  20\\
    &\le 2C_n + 20.
\end{align}

For multiple dimensions we apply the same argument across each dimension and add them up to yield the lemma.

\end{proof}

\section{Proof of Proposition \ref{prop:low-order}} \label{app:tv0}
In this section, we first prove the following result.

\begin{theorem} \label{thm:tv0}
Let $\bs p_t$ be the predictions of FLH-SIONS algorithm with parameters $\epsilon = 2$, $C = 20$ and exp-concavity factor $\sigma$. Under Assumptions A1-A4, we have that,
$$
    \sum_{t=1}^n f_t(\bs p_t) - f_t(\bs w_t)
    = \tilde O (d^2 n^{1/3} C_n^{2/3}  \vee d^2 ),\nonumber
$$
for any $C_n>0$ and any comparator sequence $\bs w_{1:n} \in \mathcal{TV}^{(0)}(C_n)$. Here $\tilde O$ hides poly-logarithmic factors of $n$ and $a \vee b = \max\{ a,b\}$.
\end{theorem}
\begin{proof}
The proof follows almost directly from the arguments in \citet{Baby2021OptimalDR}. First, we use the partition $\cP$ mentioned in Lemma 30 in \citet{Baby2021OptimalDR}. Let the partition be $cP = \{[1_s,1_t], \ldots, [M_s,M_t] \}$, with $|\cP| = M$.

Consider the following convex optimization problem.
\begin{mini!}|s|[2]                   
    {\tilde{ \bs u}_1,\ldots,\tilde{ \bs u}_n,\tilde{ \bs z}_1,\ldots,\tilde{ \bs z}_{n-1}}                               
    {\sum_{t=1}^n f_t(\tilde {\bs u}_t)}   
    {\label{eq:Example1}}             
    {}                                
    \addConstraint{\tilde{ \bs z}_t}{=\tilde{ \bs u}_{t+1} - \tilde{ \bs u}_{t} \: \forall t \in [n-1],}    
    \addConstraint{\sum_{t=1}^{n-1} \|\tilde{ \bs z}_t\|_1}{\le C_n, \label{eq:constr-ec-1d}}  
    \addConstraint{\| \tilde {\bs u}_t\|_\infty}{\le B \: \forall t \in [n],\label{eq:constr-ec-2d}}
\end{mini!}

Let $\bs u_1,\ldots,\bs u_n$ be the optimal solution to the above problem. Let $\bs w_j$ be the prediction of the FLH-SIONS algorithm at time $j$. Define:

\begin{align}
    R_n(C_n)
    &= \sum_{t=1}^n f_j(\bs w_t) - f_t(\bs u_t).
\end{align}

Define $\bar {\bs u}_i = \frac{1}{n_i} \sum_{j=i_s}^{i_t} \bs u_j$ and $\dot {\bs u}_i = \bar {\bs u}_i - \frac{1}{n_i} \sum_{j=i_s}^{i_t} \grad f_j(\bar {\bs u}_i)$. We can use the regret decomposition of \citet{Baby2021OptimalDR}.

\begin{align}
    R_n(C_n)
    &\le \sum_{i=1}^{M} \underbrace{\sum_{j=i_s}^{i_t} f_j(\bs w_j) - f_j(\dot {\bs u}_i)}_{T_{1,i}} + \sum_{i=1}^{M} \underbrace{\sum_{j=i_s}^{i_t} f_j(\dot {\bs u}_i) - f_j(\bar {\bs u}_i)}_{T_{2,i}} + \sum_{i=1}^{M} \underbrace{\sum_{j=i_s}^{i_t} f_j(\bar {\bs u}_i) - f_j(\bs u_j)}_{T_{3,i}}.
\end{align}

For any bin $[i_s,i_t] \in \cP$, we can bound $T_{2,i}+T_{3,i} = O(1)$ by using the arguments in the proof of Theorem 14 of \citet{Baby2021OptimalDR} since the losses in our case are also gradient-Lipschitz as per Assumption A3. So we only need to consider the term $T_{1,i}$. Observe that

\begin{align}
    \| \dot {\bs u}_i\|_\infty
    &\le \|  \bar {\bs u}_i \|_\infty + \frac{1}{n_i} \sum_{j=i_s}^{i_t} \|\grad f_j(\bar {\bs u}_i)\|_\infty\\
    &\le 2,
\end{align}
as per Assumptions A1-A2. Further we can view the comparator $\dot {\bs u}_i$ as a linear predictor with slope zero. The output of this linear predictor is bounded in magnitude by 2 which is less that 20. Hence FLH-SIONS under the setting of the current theorem leads to $T_{1,i} = \tilde O(d)$. Since $M = O(dn^{1/3}C_n^{2/3} \vee d)$ for the partition in Lemma 30 of adding the regret across all bins results in the theorem.
\end{proof}

Theorem\ref{thm:tv0} when combined with Theorem \ref{thm:main-d} now directly leads to Proposition \ref{prop:low-order}.

\section{Proof of Proposition \ref{prop:lb}} \label{app:lb}
The result proven in this section is mainly due to the geometric arguments in \citet{donoho1990minimax,donoho1998minimax} (or see \citet{DJBook} for a comprehensive monograph) with an extra technicality of handling boundedness constraint as per Assumption A1 (in Section \ref{sec:main}).

In the proof we make extensive use of wavelet theory and refer readers to \citet{DJBook} for necessary preliminaries.

\proplb*
\begin{proof}
We consider a uni-variate setting with the losses $f_t(w) = (d_t - w)^2$ where $d_t = u_t + \cN(0,1)$ with $u_{1:n} \in \mathcal{TV}^{(1)}(C_n)$. At each step, $d_t$ is revealed to the learner as doing so can only make learning easier. 

Let $\mathbb{W}$ be the set of whole numbers. For the purposes of analysis, we start with an abstract observation model:
\begin{align}
    y_j = \theta_j + \epsilon \cN(0,1), \: j \in \mathbb{W} \label{eq:abs}
\end{align}
where $\theta_j$ are the wavelet coefficients in a regularity-three CDJV multi-resolution basis \citep{cdjv} of a function in $\cF_1(C_n)$ from which the discrete samples $u_{1:n}$ are generated.

In what follows we will show that for any procedure estimating the wavelet coefficients (let the estimate be $\hat \theta_j$, $j \in \mathbb{W}$) we have that
\begin{align}
    \sum_{j \in \mathbb{W}} (\hat \theta_j - \theta_j)^2
    &= \Omega(C^{2/5} \epsilon^{8/5}).
\end{align}

Due to Section 15.5 of \citep{DJBook}, by taking $\epsilon = 1/\sqrt{n}$, such a guarantee will then imply a lower bound of $\Omega(n^{-4/5}C^{2/5})$ for $\frac{1}{n} \sum_{t=1}^n (u_t - \hat u_t)^2$, where $\hat u_t$ is the estimate produced by observing the data $d_t$ (assume $C = \Omega(1/\sqrt{n})$ for now). This rate will finally imply a dynamic regret lower bound in the  following manner:
\begin{align}
    E \left[ \sup_{r_{1:n} \in \mathcal{TV}^{(1)}(C)} \sum_{t=1}^n f_t(\hat u_t) - f_t(r_t)\right]
    &\ge \sup_{r_{1:n} \in \mathcal{TV}^{(1)}(C)} E \left[\sum_{t=1}^n f_t(\hat u_t) - f_t(r_t)\right]\\
    &=_{(a)} \sup_{r_{1:n} \in \mathcal{TV}^{(1)}(C)} \sum_{t=1}^n E[(\hat u_t - u_t)^2] - (r_t - u_t)^2\\
    &= \sum_{t=1}^n E[(\hat u_t - u_t)^2], \label{eq:dyn-sq}
\end{align}
where in line (a) we used the bias variance decomposition and the fact that $\hat u_t$ is independent of $d_t$ for online algorithms.

In what follows we use a dyadic indexing scheme for referring to wavelet coefficients in Eq.\eqref{eq:abs} as $\theta_{jk}$ which means the $k^{th}$ wavelet coefficient in resolution $j \ge 0$. There are $2^j$ wavelet coefficients in resolution $j$. We will also use $\theta_{j\cdot}$ to denote a sequence of $2^j$ wavelet coefficients at resolution $j$.

Let $\beta$ be the subset of wavelet coefficients at resolutions less than or equal to 2. i.e, $\beta = [\theta_{0\cdot}, \theta_{1\cdot},\theta_{2\cdot}]$ which has a length of 7.

Define a Besov norm as follows:
\begin{align}
    \| \theta \|_{b_{1,1}^{3/2}} := \| \beta \|_1 + \sum_{j \ge 3} 2^{3j/2} \|\theta_{j\cdot} \|_1.
\end{align}

Define a Besov space as:
\begin{align}
    \cA(B) := \{\theta :  \| \theta \|_{b_{1,1}^{3/2}} \le B\}.
\end{align}

It is known that $A(\kappa C) \subseteq \cF_1(C)$ for some constant $ 0 < \kappa \le 1$. (see for eg. Eq.(33) in \citep{trendfilter} along with Theorem 1 in \citep{donoho1998minimax}). 

Since the space $\cA(B)$ is solid and orthosymmetric (see Section 4.8 in \citet{DJBook}) we have that the risk of estimating coefficients from $\cA$ is lower bounded by the risk (i.e $\sum_{j \ge 0} (\hat \theta_j - \theta_j)^2$) of the hardest rectangular sub-problem as shown by \citet{donoho1990minimax}.

A hyper-rectangle is defined as follows:
\begin{align}
    \Theta(\tau) = \{\theta: |\theta_j| \le \tau_j, \: j \ge 0 \}.
\end{align}

From \citet{donoho1990minimax}, the minimax risk over a hyper-rectangle under the observation model Eq.\eqref{eq:abs} is known to be:
\begin{align}
    R^*(\tau)
    &:= \min_{\hat \theta} \max_{\theta \in \Theta(\tau)}\sum_{j \ge 0} (\hat \theta_j - \theta)^2\\
    &\ge \sum_{j \ge 0} \min\{\tau_j^2, \epsilon^2 \}.
\end{align}

So all we need to show is an appropriate hyper-rectangle (which is identified by $\tau$) within $\cA(B)$ whose minimax risk is sufficiently large. 

We next proceed to give such a hyper-rectangle. Let $j_* \in \mathbb{W}$ be the smallest number such that 
\begin{align}
2^{j_*}
&\ge \frac{C^{2/5}}{\epsilon^{2/5}}.
\end{align}

For simplicity, from now on-wards, let's assume that $j_*$ is an integer that satisfy $2^{j_*} = \frac{C^{2/5}}{\epsilon^{2/5}}$.

Define the hyper-rectangle coordinates by 
\begin{align}
    \tau_{j_*k} = \frac{\kappa C}{2^{5j_*/2}}, \label{eq:hyper}
\end{align}
for all $k = 0,1,\ldots,2^{j_*}-1$ and $\tau_{j\cdot} = 0$ for all other resolutions.

Note that $\frac{\kappa C}{2^{5j_*/2}} = \epsilon$. The minimax risk over such a hyper-rectangle then becomes
\begin{align}
    R^*(\tau)
    &= 2^{j_*} \epsilon\\
    &= (\kappa C)^{2/5} \epsilon^{8/5}.
\end{align}

Now it remains to verify that 
\begin{enumerate}
    \item The hyper-rectangle in Eq.\eqref{eq:hyper} is indeed in $\cA(\kappa C)$.
    
    \item The function produced by the coefficients in that hyper rectangle is bounded by 1 point-wise in magnitude.
\end{enumerate}

First we notice that by taking $\epsilon = 1/\sqrt{n}$ as mentioned earlier, we have
\begin{align}
    2^{j_*} > 4, 
\end{align}
whenever $C > 4^{5/2}/\sqrt{n}$. We first consider the case where $C$ is within this regime.

For the first item, we have that
\begin{align}
    \| \tau \|_{b_{1,1}^{3/2}}
    &= 2^{3j_*/2} \cdot 2^{j_*} \frac{\kappa C}{2^{5j_*/2}}\\
    &= \kappa C,
\end{align}
where we used the fact that $j_* > 2$ in the regime $C > 4^{5/2}/\sqrt{n}$.

Hence $\Theta(\tau) \subseteq \cA(\kappa C)$.

For the second item, we notice that due to Lemma B.18 in \citet{DJBook}, it is sufficient to show that $2^{j_*/2} \|\theta_{j_* \cdot} \|_\infty = O(1)$. Taking $\epsilon = 1/\sqrt{n}$ as mentioned earlier, we have that
\begin{align}
    2^{j_*/2} \|\theta_{j_* \cdot} \|_\infty
    &= \frac{\kappa C}{2^{2j_*}}\\
    &= \kappa^{1/5} C^{1/5} \epsilon^{4/5}\\
    &= \frac{\kappa^{1/5} C^{1/5}}{n^{2/5}}\\
    &\le 1,
\end{align}
in the non-trivial regime of $C \le n^2$ where we recall that $\kappa \le 1$.

For the regime where $C \le 1/\sqrt{n}$, the trivial lower bound of $ \Omega(1)$ estimation error kicks in. Thus overall we have shown that for any online algorithm producing estimates $\hat u_t$ we have that
\begin{align}
    \sum_{t=1}^n E[(\hat u_t - u_t)^2]
    &= \Omega(n^{1/5}C^{2/5} \vee 1),
\end{align}
thus obtaining a lower bound on the dynamic regret as per Eq.\eqref{eq:dyn-sq}. 

In multiple-dimensions we can consider a similar setup as before with losses $f_t(\bs w) = \|\bs d_t - \bs w \|_2^2$ with $\bs d_t[k] = \bs u_t[k] + \cN(0,1)$ where $\bs u_{1:n} \in \mathcal{TV}^{(1)}(C)$. We can consider a sequence $\bs u_{1:n}$ such that $\|nD^2 \bs u_{1:n}[k] \|_1 = C/d $ across each coordinate $k \in [d]$.
\begin{align}
    \min_{{\bs p}_{1:n}} \max_{\bs w_{1:n} \in \mathcal{TV}^{(1)}(C)} \sum_{t=1}^n f_t(\bs p_t) - f_t(\bs w_t)
    &=\sum_{k=1}^d \sum_{t=1}^n \Omega(n^{1/5}(C/d)^{2/5} \vee 1)\\
    &= \Omega(d^{3/5} n^{1/5}C^{2/5} \vee d).
\end{align}

This completes the proof of the proposition.

Next, we verify the fact in Remark \ref{rmk:inst} that the rate of $n^{1/3}[\TV_0]^{1/3}$ is of the same order as $n^{1/5}[\TV_1]^{2/5}$ for the comparator sequence constructed above.

Define the following norm:
\begin{align}
    \| \theta \|_{b_{1,1}^{1/2}} := \| \beta \|_1 + \sum_{j \ge 2} 2^{j/2} \|\theta_{j\cdot} \|_1.
\end{align}

Define another Besov space as:
\begin{align}
    \cG(B) := \{\theta :  \| \theta \|_{b_{1,1}^{1/2}} \le B\}.
\end{align}

Let $w_{1:n}$ be the sequence of reals and let $\theta_{1:n}$ be its wavelet coefficients. It is known from \citet{DeVore1993ConstructiveA} that

\begin{align}
     \| \theta \|_{b_{1,1}^{1/2}} \asymp \TV_0(w_{1:n}) ,
\end{align}
where $\asymp$ means that the quantities have similar scaling. 

So we only need to compute the norm $\| \theta \|_{b_{1,1}^{1/2}}$ for the hard instance created above. We have that:

\begin{align}
    \| \theta \|_{b_{1,1}^{1/2}}
    &= 2^{j_*/2} \cdot 2^{j_*} \kappa C/2^{5j_*/2}\\
    &= \kappa C/2^{j_*}\\
    &= \kappa  C^{3/5}/n^{1/5}.
\end{align}

Thus the $\TV_0$ of the sequence scales as $C^{3/5}/n^{1/5}$. Hence the rate:
\begin{align}
    n^{1/3}[\TV_0(w_{1:n})]^{2/3}
    \asymp n^{1/5} C^{2/5}.
\end{align}

Thus for the hard instance constructed in the proof, both the rates grow with similar scale.

\end{proof}

\section{Why the analysis of \citet{Baby2021OptimalDR} leads to sub-optimal regret?} \label{app:compare}

For simplicity, we consider a uni-variate setting. First we derive a tighter regret guarantee (than one implied by Proposition \ref{prop:embed}) of   $O(n^{1/3}C_n^{2/3} \vee 1)$ for the results of \citet{Baby2021OptimalDR} when applied to our setting. Then we explain the source of sub-optimality in their analysis. Throughout this section, we assume that the condition of low TV1 regime defined in Section \ref{sec:intro} is satisfied.

First, let's define the comparator classes:
\begin{align}
    \TV^{(1)}(C)
    &:= \{\theta_{1:n} : \TV_1(\theta_{1:n}) \le C \},
\end{align}

and

\begin{align}
    \TV^{(0)}(C)
    &:= \{\theta_{1:n} : \TV_0(\theta_{1:n}) \le C \}.
\end{align}


Let $u_{1:n}$ be the offline optimal sequence as per Lemma \ref{lem:kkt}.

In accordance to the details in Section \ref{sec:alg}, we can interpret a comparator sequence $u_{1:n} \in \mathcal{TV}^{(1)}(C_n)$ as a continuous piece-wise linear sequence. Then the dynamic regret can be expressed as:
\begin{align}
    R_n(u_{1:n})
    &= \sum_{t=1}^n f_t(p_t) - f_t(u_t)\\
    &=_{(a)} \sum_{t=1}^n f_t(\bs \alpha_t^T \bs x_t) - f_t(\bs \beta_t^T \bs x_t)\\
    &:=_{(b)} \sum_{t=1}^n \tilde f_t(\bs \alpha_t) - \tilde f_t(\bs \beta_t),
\end{align}
where in Line (a) we define $\bs x_t = [1,t/n]^T$ and $\bs \alpha$ and $\bs \beta$ are chosen such that $p_t = \bs \alpha_t^T \bs x_t$ and $u_t = \bs \beta_t^T \bs x_t$. Further the predictors $\bs \beta_t$ are chosen to satisfy $\bs \beta_t^T \bs x_t = \bs \beta_{t+1}^T \bs x_t$ so that the sequence $u_{1:n}$ can be interpreted as a piece-wise linear signal that is also continuous at every transition point where the slope changes (see Definition \ref{def:piecelinear}). 

In Line (b) we define $\tilde f_t(\bs v) = f_t(\bs v^T \bs x_t)$. We chose the co-variates as $\bs x_t = [1,t/n]^T$ instead of $\bs x_t = [1,t]^T$ so that the losses $\tilde f_t(\bs v)$ remains Lipschitz and gradient Lipschitz whenever $|\bs v^T \bs x_t| = O(1)$ which is a requirement for the results in \citep{Baby2021OptimalDR}.

By using similar line of arguments used to derive \eqref{eq:incpt}, we obtain
\begin{align}
    \bs \beta_{t+1}[1] - \bs \beta_{t}[1]
    &= \frac{t}{n} \left( \bs \beta_{t}[2] - \bs \beta_{t+1}[2] \right).
\end{align}

Hence we have that
\begin{align}
    \sum_{t=1}^{n-1} |\bs \beta_{t+1}[1] - \bs \beta_{t}[1]|
    &\le \sum_{t=1}^{n-1} | \bs \beta_{t}[2] - \bs \beta_{t+1}[2] |\\
    &= n\|D^2 u_{1:n} \|_1,
\end{align}
where we used the fact that the sum of difference of the slopes (see Definition \ref{def:piecelinear}) in the linear representation of $u_{1:n}$ with co-variates $\bs x_t = [1,t/n]^T$ is exactly equal to $n\|D^2 u_{1:n} \|_1$.

Thus overall, we obtain that
\begin{align}
    \sum_{t=1}^{n-1} \|\bs \beta_t - \bs \beta_{t+1} \|_1
    &\le 2n\|D^2 u_{1:n} \|_1\\
    &\le 2C_n, \label{eq:beta-tv}
\end{align}
as $u_{1:n} \in \mathcal{TV}^{(1)}(C_n)$.

Hence by the results of \citet{Baby2021OptimalDR} we have that
\begin{align}
    R_n(u_{1:n})
    &= \tilde O(n^{1/3}C_n^{2/3} \vee 1).
\end{align}

Next, we proceed to explain source of this sub-optimality in the analysis of \citet{Baby2021OptimalDR}.

In \citet{Baby2021OptimalDR} (Lemma 5)  they form a partition $\cP'$ of $\bs \beta_{1:n}$ so that in the $i^{th}$ bin (represented by $[i_s,i_t]$) we have that:
\begin{itemize}
    \item $\|D \bs \beta_{i_s:i_t}\| \le 1/\sqrt{\ell_{i_s \rightarrow i_t}}$
    \item $\|D \bs \beta_{i_s:i_t+1}\| > 1/\sqrt{\ell_{i_s \rightarrow i_t+1}}$
\end{itemize}
where we recall that $\ell_{a \rightarrow b} = b-a+1$.

So we have that within bin $[i_s,i_t] \in \cP'$, $\|D \bs \beta_{i_s:i_t}[2]\|_1 \le 1/\sqrt{\ell_{i_s \rightarrow i_t}}$. This amounts to saying that
\begin{align}
    \|D^2 u_{i_s:i_t} \|_1
    &\le 1/(n\sqrt{\ell_{i_s \rightarrow i_t}}).
\end{align}

While in the partition $\cP$ that we construct in Lemma \ref{lem:keypart} we have that
\begin{align}
        \|D^2 u_{i_s:i_t} \|_1
    &\le 1/{\ell_{i_s \rightarrow i_t}}^{3/2}.
\end{align}

Comparing the previous two inequalities, we conclude that the sequence within each bin of $\cP'$ is much smoother than that of $\cP$.

This will result in the formation of $| \cP'| = O(n^{1/3}C_n^{2/3} \vee 1)$ bins overall as per Eq.\eqref{eq:beta-tv} (see Lemma 5 in \citet{Baby2021OptimalDR}) which is larger than the $O(n^{1/5}C_n^{2/5} \vee 1)$ bins in $\cP$ under the low TV1 regime.

Within each bin $ [i_s,i_t] \in \cP'$ \citet{Baby2021OptimalDR} uses a three term regret decomposition as follows:

\begin{align}
    T_{[i_s,i_t]}
    &:=\sum_{j=i_s}^{i_t} \tilde f_j(\bs \alpha_j) - \tilde f_j(\bs \beta_j)\\
    &= \underbrace{\sum_{j=i_s}^{i_t} \tilde f_j(\bs \alpha_j) - \tilde f_j(\dot{\bs \beta})}_{T'_1}
    + \underbrace{\sum_{j=i_s}^{i_t} \tilde f_j(\dot{\bs \beta}) - \tilde f_j(\bar{\bs \beta})}_{T'_2} + \underbrace{\sum_{j=i_s}^{i_t} \tilde f_j(\bar{\bs \beta}) - \tilde f_j({\bs \beta_j})}_{T'_3}, \label{eq:regdcomp}
\end{align}
where $\bar{\bs \beta} = \frac{1}{n} \sum_{j=i_s}^{i_t} \bs \beta_j$ and $\dot{\bs \beta} = \bar{\bs \beta} - \frac{1}{\ell_{i_s \rightarrow i_t}} \sum_{j=i_s}^{i_t} \grad \tilde f_j(\bar{ \bs \beta})$.

Then \citet{Baby2021OptimalDR} proceed to show that this one step gradient descent based decomposition is sufficient to keep $T_{[i_s,i_t]} = O(1)$ leading to an overall regret of $O(n^{1/3}C^{2/3} \vee 1)$ when summed across all bins.

In our case the main challenge is to keep $T_{[i_s,i_t]} = \tilde O(1)$ for $[i_s,i_t] \in \cP$ while dealing with the fact that sequence within each bin of $\cP$ is much less smooth than that in $\cP'$. We accomplish this via a newton step based decomposition with a careful analysis as detailed in Section \ref{sec:1d} (It was found that the one-step gradient descent as in Eq.\eqref{eq:regdcomp} doesn't keep $T'_2$ negative enough to make $T'_2 + T'_3 = O(1)$ for bins in $\cP$). Eventhough the sequence in bins $\cP$ is wigglier than that of bins in $\cP'$, overall the sequence, $u_{1:n}$ from a $\mathcal{TV}^{(1)}$ class is much smoother than the sequences from $\mathcal{TV}^{(0)}$ class in the low TV1 regime due to sufficiently slowly changing piecewise linear structure. This extra smoothness property is what allowed us to consider \emph{larger} (in terms of mean bin width) bins and hence smaller partition size (when compared to $\cP'$) and still bound the regret within each bin to be $\tilde O(1)$. Adding this bound across all bins in $\cP$ then lead to the optimal rate of $ \tilde O(n^{1/5}C_n^{2/5} \vee 1)$.

\section{More examples from the low TV1 regime} \label{app:example}

We list some examples where the low TV1 regime defined in Section \ref{sec:intro} is satisfied. Under this regime, the rate of $\tilde O(n^{1/5}[\TV_1(\bs w_{1:n})]^{2/5} \vee 1)$ attained by FLH-SIONS via Theorem \ref{thm:main-d} is faster than the rate of $\tilde O(n^{1/3} [\TV_0(\bs w_{1:n})]^{2/3} \vee 1)$ attained by \citet{Baby2021OptimalDR}. This is a non-exhaustive list of examples and one can construct many other examples as well. All the examples we consider here are for uni-variate setting, through the extension to multi-dimensions is a straight-forward replication of the sequence generating process across each coordinate. 

We begin by a minimalist example yielding logarithmic dynamic regret rate.
\begin{example}
Consider a sequence $\theta_{1:n}$ such that $\theta_t = t/n$ for $t \in \{0,1,\ldots,n \}$. This is a sequence obtained via descretizing a linear signal. We have that $\TV_1(\theta_{1:n}) = 0$ and $\TV_0(\theta_{1:n}) = 1$. So by Theorem \ref{thm:main-d} we have that the rate attained by FLH-SIONS is $O(\log n)$ while the rate attained by \citet{Baby2021OptimalDR} is $O(n^{1/3})$.
\end{example}

Next, we give an example where both $\TV_1$ and $\TV_0$ distance of a sequence is growing with $n$.
\begin{example}
For an integer $s < n$, define $a_{1:s} = 0,s/n,2s/n,\ldots, \frac{s(n/s-1)}{n}$. Let $b_{1:s}$ be the mirror image of $a_{1:s}$.i.e $b_{1:s} =  \frac{s(n/s-1)}{n},\ldots,s/n,0$. For simplicity lets' assume that $n/s$ is an integer. Form a sequence $\theta_{1:n} := a_{1:s},b_{1:s},a_{1:s},b_{1:s},\ldots,a_{1:s},b_{1:s}$ by concatenating the sequences $a_{1:s},b_{1:s}$ for $s/2$ times. This sequence transitions between 0 and 1 through linear sections. For this sequence, we have that $\TV_0(\theta_{1:n}) = s$ and $\TV_1(\theta_{1:n}) = 2s^2$. Let $s = n^{\alpha}$ for some $0 < \alpha < 1$. Thus Theorem \ref{thm:main-d} yields a rate of $\tilde O(n^{\frac{4\alpha+1}{5}})$ while the results in \citet{Baby2021OptimalDR} yields only a rate of $\tilde O(n^{\frac{2\alpha+1}{3}})$  which is a slower rate for all $\alpha < 1$.
\end{example}

\end{document}